\documentclass[twoside]{article}

\usepackage[accepted]{aistats2023}
\usepackage[utf8]{inputenc} 
\usepackage[T1]{fontenc}    
\usepackage{hyperref}       
\usepackage{xurl}           
\usepackage{booktabs}       
\usepackage{amsfonts}       
\usepackage{nicefrac}       
\usepackage{microtype}      
\usepackage{xcolor}         

\usepackage{graphicx}
\usepackage{subcaption}
\usepackage{epsf}
\usepackage{fancyhdr}
\usepackage{graphics}
\usepackage{psfrag}

\usepackage{wrapfig}
\usepackage{float}

\usepackage[round]{natbib}

\usepackage{algorithm}
\usepackage{algorithmic}

\usepackage{color}

\usepackage{amsmath}
\usepackage{amssymb}
\usepackage{mathtools}
\usepackage{amsthm,amssymb}
\usepackage{bm,bbm}
\usepackage{multirow}
\usepackage{balance}

\DeclareMathOperator*{\argmax}{arg\,max}
\DeclareMathOperator*{\argmin}{arg\,min}

\usepackage{verbatim}

\newcommand\blfootnote[1]{%
  \begingroup
  \renewcommand\thefootnote{}\footnote{#1}%
  \addtocounter{footnote}{-1}%
  \endgroup
}

\usepackage{hyperref}

\let\emptyset\varnothing

\makeatletter
\def\munderbar#1{\underline{\sbox\tw@{$#1$}\dp\tw@\z@\box\tw@}}
\makeatother

\newtheorem{theorem}{Theorem}[section]
\newtheorem{definition}[theorem]{Definition}
\newtheorem{lemma}[theorem]{Lemma}
\newtheorem{remark}[theorem]{Remark}
\newtheorem{corollary}[theorem]{Corollary}
\newtheorem{proposition}[theorem]{Proposition}

\newcommand{\be}{\begin{equation}}
\newcommand{\ee}{\end{equation}}
\newcommand{\bea}{\begin{equation*}\begin{aligned}}
\newcommand{\eea}{\end{aligned}\end{equation*}}

\newcommand{\R}{\mathbb{R}}
\newcommand{\N}{\mathbb{N}}

\newcommand{\G}{{\mathbb G}}
\newcommand{\K}{{\mathbb K}}

\newcommand{\calS}{{\mathcal S}}
\newcommand{\calM}{{\mathcal M}}
\newcommand{\calT}{{\mathcal T}}
\newcommand{\calW}{{\mathcal W}}

\newcommand{\calF}{{\mathcal F}}

\newcommand{\dd}{\mathrm{d}}
\newcommand{\e}{{\epsilon}}

\newcommand{\Let}{\triangleq}

\begin{document}

%

%

\runningauthor{Tam Le, Truyen Nguyen, Kenji Fukumizu}

\twocolumn[

\aistatstitle{Scalable Unbalanced Sobolev Transport for Measures on a Graph}

\aistatsauthor{Tam Le $^{\ast, \dagger, \ddagger}$ \And Truyen Nguyen $^{\ast, \diamond}$ \And  Kenji Fukumizu $^{\dagger}$ }


\vspace{0.5mm}
\aistatsaddress{
The Institute of Statistical Mathematics $^{\dagger}$ \\
The University of Akron $^{\diamond}$ \\ RIKEN AIP $^{\ddagger}$} 

]

\begin{abstract}
Optimal transport (OT) is a popular and powerful tool for comparing probability measures. However, OT suffers a few drawbacks: (i) input measures required to have the same mass, (ii) a high computational complexity, and (iii) indefiniteness which limits its applications on kernel-dependent algorithmic approaches. To tackle issues (ii)--(iii),~\cite{le2022st} recently proposed Sobolev transport for measures on a graph  having the \emph{same total mass} by leveraging the graph structure over supports. In this work, we consider measures that may have \emph{different total mass} and are supported on a graph metric space. To alleviate the disadvantages (i)--(iii) of OT,  we propose a novel and scalable approach to extend Sobolev transport for this \emph{unbalanced} setting where measures may have different total mass. We show that the proposed \emph{unbalanced Sobolev transport} (UST) admits a closed-form formula for fast computation, and it is also negative definite. Additionally, we derive geometric structures for the UST and establish relations between our UST and other transport distances. We further exploit the negative definiteness to design positive definite kernels and evaluate them on various simulations to illustrate their fast computation and comparable performances against other transport baselines for unbalanced measures on a graph.
\end{abstract}

\section{INTRODUCTION}

Optimal transport (OT) has become a popular approach and its theory lays out a compelling toolkit for data analysis on probability distributions. OT has been leveraged in several research areas such as machine learning~\citep{peyre2019computational, nadjahi2019asymptotic, titouan2019optimal, bunne2019, bunne2022proximal, janati2020entropic, muzellec2020missing,  paty2020regularity, mukherjee2021outlier, altschuler2021averaging, fatras2021unbalanced, le2021flow, le2021adversarial, liu2021lsmi, nguyen2021optimal, scetbon2021low, si2021testing, pmlr-v151-takezawa22a, fan2022complexity}, computer vision~\citep{nguyen2021point, Saleh_2022_CVPR, wang2022optimal}, and statistics~\citep{mena2019statistical, pmlr-v99-weed19a, liu2022entropy, nguyen2022many, nietert2022outlier, pmlr-v151-wang22f} to name a few. Nevertheless, it has some fundamental disadvantages.\blfootnote{$^*$: Two authors contributed equally.}

One drawback of OT is that it requires input measures having the \emph{same mass} for the transportation. To address this problem, several proposals have been developed in the recent literature. For examples, the \emph{partial optimal transport} (POT)~\citep{CM, figalli2010optimal} constraints a fixed amount of mass for transportation; the \emph{optimal entropy transport} (OET)~\citep{Liero2018, chizat2018unbalanced, kondratyev2016new} optimizes a sum of a transport functional and two convex entropy functionals. Additionally, there are various other approaches, e.g., the Kantorovich-Rubinstein discrepancy~\citep{hanin1992kantorovich, guittet2002extended, lellmann2014imaging, sato2020fast}, the unbalanced mass transport~\citep{benamou2003numerical}, the generalized Wasserstein distance~\citep{P1, P2}, the unnormalized optimal transport~\citep{gangbo2019unnormalized}, and the entropy partial transport~\citep{le2021ept}. These approaches are either special cases of the OET (e.g., by using some specific instances of entropy functional such as the total variation distance, $\ell^2$ distance), or a variant of OET (e.g., by using the $\ell^p$ distance, partial transport in place of the entropy functional, transport functional respectively). It is worth pointing  out that the unbalanced setting for measures with unequal mass has been applied in several application domains and learning problems, e.g., color transfer and shape matching~\citep{bonneel2015sliced}; multi-label learning~\citep{frogner2015learning}; positive-unlabeled learning~\citep{chapel2020partial}; natural language processing and topological data analysis~\citep{le2021ept}. In particular, the unbalanced approach becomes essential when supports of input measures are subject to noise or have outliers since such supports are not desirably aligned in the matching problem~\citep{frogner2015learning, balaji2020robust, mukherjee2021outlier}.  

Another drawback of standard OT is that it has a high computational complexity. This disadvantage also exists in the unbalanced optimal transport (UOT), which hinders its applications, especially for large-scale settings. For examples, let us consider the OET with Kullback-Leibler divergence for the entropy functional which is widely used in applications. For this, one can leverage the entropic regularization to derive efficient Sinkhorn-based algorithmic approach~\citep{frogner2015learning, chizat2018scaling, sejourne2019sinkhorn} which has a quadratic complexity \citep{pham2020unbalanced}. Another popular approach to scale up UOT is to exploit geometric structures of supports, e.g., one-dimensional structure~\citep{bonneel2019spot, pmlr-v151-sejourne22a}, tree structure~\citep{le2021ept, sato2020fast}. More concretely, \cite{bonneel2019spot} proposed the sliced partial optimal transport (SPOT) by projecting supports into a random one-dimensional space. By assuming a unit mass on each support, they developed an efficient algorithmic approach with a quadratic complexity for the worst case. Nonetheless, SPOT suffers a curse of dimensionality since using one-dimensional projections for supports limits its ability to capture topological structures of distributions, especially in a high-dimensional space. \cite{le2021ept} proposed the entropy partial transport (EPT) by exploiting a tree structure to remedy the curse of dimensionality for SPOT. Moreover, EPT yields the first closed-form solution among various variants of UOT (i.e., its complexity is linear to the number of edges in a tree) for fast computation which is applicable for large-scale settings. However, tree structure may be a restricted condition which narrows down its practical usage in applications. 

The aforementioned circumstances motivate us to consider measures with \emph{unequal mass} and supported on a \emph{graph metric} space which has more degrees of freedom (i.e., graph structure rather than tree structure) and appears more popularly in applications. Inspired by the Sobolev transport~\citep{le2022st} for probability measures on a graph, we propose a \emph{novel and scalable} approach to leverage graph structure and extend Sobolev transport for the \emph{unbalanced} setting. At a high level, our contributions are three-fold as follow: 

\begin{itemize}
\vspace{-4pt}
\item we propose a novel \emph{$p$-order unbalanced Sobolev transport} (UST) ($p \ge 1$) for measures with unequal mass and supported on a graph metric space. We prove that UST admits a \emph{closed-form formula} for a fast computation and it is \emph{negative definite};
\vspace{-4pt}
\item we derive geometric structures for the UST and propose \emph{positive definite kernels} built upon the UST. Additionally, we establish relations between UST and the EPT on a \emph{graph};
\vspace{-4pt}
\item we empirically illustrate that UST is fast for computation (i.e., \emph{closed-form solution} of UST). Also various simulations demonstrate that  the performances of the proposed kernels for UST compare favorably with other unbalanced transport baselines for measures with unequal mass on a graph.

\end{itemize}

The paper is organized as follows: we introduce notations and the problem setup in \S\ref{sec:pre}. In \S\ref{sec:entropy_partialOT}, we extend and derive the EPT for unbalanced measures \emph{on a graph}. We then present our main contribution: the UST for measures with unequal mass on a graph in \S\ref{sec:regularization} and derive its properties in \S\ref{sec:Prop_US}. In \S\ref{sec:experiments}, we evaluate the proposed kernel for UST against other unbalanced transport baselines for measures with unequal mass on a graph on various simulations. We conclude our work in \S\ref{sec:conclusion}. The detailed proofs for our theoretical results are placed in Appendix \S\ref{app:subsec:detailedproofs}. Furthermore, we have released code for our proposals.\footnote{\url{https://github.com/lttam/UnbalancedSobolevTransport}}

\section{PRELIMINARIES}\label{sec:pre}

In this section, we introduce our problem setting, notations, and review relevant definitions. 

We consider the same graph setting $\G =(V,E)$ where $V, E$ are sets of nodes and edges respectively as in~\citep{le2022st} for Sobolev transport. More precisely, $\G$ is an undirected, connected and physical graph in the sense that $V \subset \R^n$ and each edge $e \in E$ is the standard line segment in $\R^n$ connecting the two corresponding end-points of $e$. Graph $\G$ has positive edge lengths $\{w_e\}_{e\in E}$ and is imposed a graph metric $d_{\G}(\cdot, \cdot)$ which equals to the length of the shortest path on $\G$. Following a convention in~\citep{le2022st}, by graph $\G$, we mean the set of all nodes in $V$ and all points forming the edges in $E$, i.e., the continuous setting for graph $\G$. We also assume that there exists a fixed root node $z_0\in V$ such that for every $x \in \G$,  $d_\G(x,z_0)$ is attained by the unique shortest path connecting $x$ and $z_0$, i.e., the uniqueness property of the shortest paths~\citep{le2022st}.

Given a point $x\in \G$ (resp. an edge $e \in E$ in $\G$), we denote $\Lambda(x)$ (resp. $\gamma_e$) as the collection of all points $y \in \G$ such that the unique shortest path in $\G$ connecting the root node $z_0$ and $y$ contains the point $x$ (resp. the edge $e$). That is, 
\begin{equation}\label{sub-graph}
\vspace{-1pt}
  \Lambda(x)  \Let \big\{y\in \G: \, x\in [z_0,y]\big\},
\end{equation}
\begin{equation}
  \gamma_e  \Let \big\{y\in \G: \, e\subset  [z_0,y]\big\}, 
\end{equation}
where  we write $[z_0, y]$ for  the shortest path in $\G$ connecting the root node $z_0$ and $y$. 

We denote $\calM(\G)$ (resp.~$\calM(\G\times \G)$) as the set of all nonnegative Borel measures on $\G$ (resp.~$\G\times \G$) with a finite mass. By continuous function $f$ on $\G$, we mean that $f: \G\to \R$ is continuous w.r.t.~the topology on $\G$ induced by the Euclidean distance. Similar adoption is also applied for continuous functions on $\G\times\G$. We denote $C(\G)$ as the collection of all continuous functions on $\G$.
 
Given a scalar $b>0$, a function $w:\G\to\R$ is called $b$-Lipschitz w.r.t.~the graph metric $d_\G$ if 
\[
|w(x) - w(y)|\leq b \, d_\G(x,y), \forall x, y \in \G.
\]
For $1\leq p \leq  \infty$, we denote $p'$ as its conjugate, i.e., $p' \in [1,\infty]$ s.t., $\frac{1}{p} +\frac{1}{p'}=1$. For a nonnegative Borel measure $\omega$ on $\G$, let   $L^p( \G, \omega)$ denote the space of all Borel measurable functions  $ f:\G\to \R$ satisfying $\int_\G |f(y)|^p \omega(\mathrm{d}y) <\infty$. When $p=\infty$, we assume that $f$ is bounded $\omega$-a.e. instead. Functions $f_1, f_2 \in L^p( \G, \omega)$ are considered to be the same  if $f_1(x) =f_2(x)$ for $\omega$-a.e. $x\in\G$. Then, $L^p( \G, \omega)$ is a normed space with the norm  defined by
\[
\|f\|_{L^p(\G, \omega)} \Let \left(\int_\G |f(y)|^p \omega(\dd y)\right)^\frac{1}{p} \text{ for } 1\leq p < \infty, \text{ and}
\]
\[
\|f\|_{L^{\infty}(\G, \omega)} \Let \inf\left\{t \in \R:\, |f(x)|\leq t \mbox{ for $\omega$-a.e. } x\in\G\right\}.
\]

Recall that Sobolev transport for probability measures on a graph is an instance of integral probability metrics (IPM)~\citep{muller1997integral}. Intuitively, the definition of Sobolev transport is based on the dual form of the $1$-order Wasserstein distance, but its Lipschitz constraint for the critic function is considered in the graph-based Sobolev space (see \citep[\S3]{le2022st} for the detail). As a consequence, it may \emph{not} possible to directly leverage approaches for standard OT (e.g., partial OT, entropy (partial) transport) to extend Sobolev transport for \emph{unbalanced} measures on a \emph{graph}. 

In this paper, we propose a \emph{detour} to develop unbalanced Sobolev transport for measures with unequal mass on a graph. We first take a step back to leverage the EPT (for unbalanced measures on a \emph{tree})~\citep{le2021ept} and extend it for unbalanced measures on a \emph{graph} (\S \ref{sec:entropy_partialOT}). Although it is still a great challenge to efficiently compute the EPT for unbalanced measures on a \emph{graph}, this novel extension (especially its dual form) plays a cornerstone in deriving a scalable approach for the proposed unbalanced Sobolev transport (UST) (\S \ref{sec:regularization}).

\section{ENTROPY PARTIAL TRANSPORT ON A GRAPH}\label{sec:entropy_partialOT}

The entropy partial transport (EPT)~\citep{le2021ept} is  developed for unbalanced measures on a \emph{tree}. 
In this section, we  propose an extension of EPT  for unbalanced measures on a \emph{graph}. Intuitively, EPT optimizes a sum of a transport function and two convex entropy functions in a similar spirit to the OET~\citep{Liero2018, chizat2018unbalanced}. We first consider the primal formulation of EPT on a \emph{graph}. We then derive its dual formulation which is the main result of this section. This novel dual formulation paves the way for our development of the UST (\S\ref{sec:regularization}).

Given two measures $\mu, \nu \in \calM(\G)$ which may have different total mass, consider the set
\[
\Pi_{\leq}(\mu,\nu) \Let \left\{ \gamma \in \calM(\G \times \G):  \, \gamma_1\leq \mu, \, \gamma_2\leq \nu \right\}
\]
where $\gamma_1$ and $\gamma_2$  respectively denote the first and second marginals of $\gamma$; by $\gamma_1\leq \mu$, we mean that $\gamma_1(B)\leq \mu(B)$ for every Borel set $B\subset \G$. Similar convention is used  when we write $\gamma_2\leq \nu$. 

For $\gamma \in \Pi_{\leq}(\mu,\nu)$, let $f_1$ and $f_2$ respectively be  the Radon-Nikodym derivatives of $\gamma_1$ w.r.t. $\mu$ and of $\gamma_2$ w.r.t. $\nu$, i.e., $\gamma_1=f_1 \mu$ and $\gamma_2 = f_2 \nu$. Then, we have $0\leq f_1\leq 1$ $\mu$-a.e., and $0\leq f_2\leq 1$ $\nu$-a.e. The weighted relative entropies  of $\gamma_1$ w.r.t. $\mu$ and of $\gamma_2$ w.r.t. $\nu$ are defined by
\[
\calF_1(\gamma_1| \mu) \Let \int_\G w_1(x) F_1(f_1(x) ) \mu(\dd x),
\]
\[
\calF_2(\gamma_2| \nu) \Let \int_\G w_2(x) F_2(f_2(x) ) \nu(\dd x),
\]
where $F_1, \, F_2: [0,1]\to (0,\infty)$ are convex and lower semicontinuous entropy functions; and $w_1, w_2:\G \to [0,\infty)$ are given nonnegative weight functions.

Given a continuous cost function $c:\G\times\G \to \R$ with $c(x,x)=0$, a constant $b\geq 0$ and a fixed scalar $m\in [0,\bar m]$ where $\bar m \Let \min\{\mu(\G), \nu(\G) \}$, we consider the primal formulation of EPT problem on a \emph{graph}:
\begin{eqnarray}\label{original}
\hspace{-0.1em} \mathrm{W}_{c,m}(\mu,\nu) 
 \Let \hspace{-1.5em} \inf_{\gamma \in \Pi_{\leq}(\mu,\nu), \, \gamma(\G\times \G)=m}
\hspace{-0.2em} \Big[ \calF_1(\gamma_1| \mu )  + \calF_2(\gamma_2| \nu )  \nonumber \\
+ \, b \int_{\G \times \G} c(x,y) \gamma(\dd x, \dd y) \Big]. \hspace{0.2em}
\end{eqnarray}
Following \citep{le2021ept}, we consider \[
F_1(s)=F_2(s)=|s-1|
\]
for the entropy functions in \eqref{original} and form a Lagrange multiplier $\lambda\in\R$ conjugate to the constraint $\gamma(\G\times \G)=m$. As a result, we instead study the problem
\begin{eqnarray}\label{P1}
\mathrm{ET}_{c,\lambda}(\mu,\nu) 
= \inf_{\gamma \in \Pi_{\leq}(\mu,\nu)} \mathcal{C}_\lambda(\gamma),
\end{eqnarray}
where $\mathcal{C}_\lambda(\gamma)$  is defined as
\begin{eqnarray}\label{easier-form}
& \hspace{-2.5em} \mathcal{C}_\lambda(\gamma) \Let \int_\G  \hspace{-0.1 em} w_1 \mu(\dd x) + \int_\G  \hspace{-0.1 em} w_2  \nu(\dd x) -\int_\G  \hspace{-0.1 em} w_1 \gamma_1(\dd x) \nonumber \\
& \hspace{1.4em} - \int_\G  \hspace{-0.1 em} w_2\gamma_2(\dd x)+ b  \int_{\G \times \G} [c(x,y)-\lambda]\gamma(\dd x, \dd y). \hspace{0.2em}
\end{eqnarray}
The connection between problem  \eqref{original} with mass constraint $m$  and problem \eqref{P1} with Lagrange multiplier $\lambda$ is given in Theorem~\ref{thm:m-via-lambda} (Appendix \S\ref{app:subsec:furthertheoreticalresults}). Also, from Theorem~\ref{thm:m-via-lambda}, we see that solving the auxiliary problem \eqref{P1} gives us a solution  to  the original problem  \eqref{original}. We now derive a novel dual formulation for  problem \eqref{P1} which paves the way for our proposed UST (\S\ref{sec:regularization}).

\begin{theorem}[Dual formula for general cost]\label{thm:duality} 
For $\lambda \geq 0$, nonnegative weights $w_1, w_2$, and two input measures $\mu, \nu \in \calM(\G)$, 
we have
\[
\mathrm{ET}_{c,\lambda}(\mu,\nu) = \sup_{(u,v)\in \K} \Big[\int_{\G}  u(x) \mu(\dd x) +  \int_{\G}  v(x)  \nu(\dd x)\Big],
\]
where $\K \Let\Big\{(u,v):\,  u\leq w_1,\, -b \lambda + \inf_{x\in\G} [b\, c(x,y) -w_1(x)]\leq v(y)\leq w_2(y),\,  u(x) + v(y)\leq  b[c(x,y) - \lambda]\Big\}$.
\end{theorem}

The main idea of proving this result is to attach to the graph $\G$ a new point $\hat s$, and then suitably and carefully extend the cost $c$ and the input distributions $\mu, \nu$ to the set $\hat \G \Let \G \cup \{\hat s\}$ inspired by an observation in \citep{CM}. The key point  of this extension is to ensure that the extended input distributions on $\hat \G$ have the same total mass and the value of the new balanced OT between extended input distributions on $\hat \G$ is \emph{equal} to that of the original EPT on graph $\G$ (i.e., the unbalanced setting). We then exploit the dual theory for the new balanced OT problem on $\hat \G$ to establish the dual formulation for our EPT problem on graph $\G$ (see Appendix \S\ref{app:subsec:detailedproofs} for detailed proof). When the ground cost $c$ is the graph metric $d_\G$, the dual formula in Theorem~\ref{thm:duality}  can be rewritten in a simpler and more symmetric form as follows.
\begin{corollary}[Dual formula for graph metric]\label{cor:duality} Assume that  $\lambda \geq 0$ and   the nonnegative weight functions $w_1, w_2$ are $b$-Lipschitz w.r.t. $d_\G$. For simplicity, let $\mathrm{ET}_\lambda \Let\mathrm{ET}_{d_\G,\lambda}$.
Then, we have
\begin{eqnarray}\label{equ:ETlambda}
\mathrm{ET}_\lambda(\mu,\nu) = \sup_{f\in \mathbb{U}} \int_\G f (\mu - \nu) - \frac{b\lambda}{2}\big[ \mu(\G) +  \nu(\G)\big],
\end{eqnarray}
where $ \mathbb{U} \Let \big\{f\in C(\G) :
 -w_2 - \frac{b\lambda}{2}\leq f \leq w_1  + \frac{b\lambda}{2}, \, |f(x)-f(y)|\leq b \, d_\G(x,y)\big\}$.
\end{corollary}

\begin{remark}
We remark that one cannot directly use the dual formulation in \citep{le2021ept}, or that of \citep{P1, P2} for unbalanced measures on a \emph{graph} since the considered problem does not satisfy the conditions imposed in these approaches for duality.
\end{remark}

In principal, for input unbalanced measures on a graph, it is simpler to learn the optimal $f^*$ in dual form \eqref{equ:ETlambda} than to learn the optimal $\gamma^*$ in  primal form \eqref{P1}. This is due to the fact that the critic $f^*$ is a function on the lower dimensional space compared to $\gamma^*$. Moreover, the Lipschitz constraint for $f^*$ is easier to handle than the constraint $\Pi_{\leq}(\mu,\nu)$ for $\gamma^*$. Nevertheless, it is still a challenge to effectively compute $\mathrm{ET}_\lambda$ using \eqref{equ:ETlambda}. 

As illustrated in \citep{LYFC, le2021ept} for transport problems on a \emph{tree}, the Lipschitz constraint for the critic $f$ can be effectively optimized by leveraging the \emph{tree structure} supports. Furthermore, the Lipschitz constraint is linked with the  $1$-order Wasserstein distance via the Kantorovich duality formulation. Due to the different nature of duality for $p$-order Wasserstein distance when $p>1$, it is however unknown that one can extend the  fast computational results in \citep{LYFC, le2021ept} to $p$-order Wasserstein distance with $p > 1$, even for measures on a \emph{tree}. 

To alleviate this, we propose in the next section an efficient $p$-order unbalanced Sobolev transport for measures with unequal mass on a \emph{graph} for any $p \ge 1$. 

\section{UNBALANED SOBOLEV TRANSPORT}\label{sec:regularization}

As pointed out in \S\ref{sec:entropy_partialOT}, it is a great challenge to efficiently compute $\mathrm{ET}_\lambda$ (i.e., the EPT problem) for unbalanced measures on a \emph{graph} using either the primal form \eqref{P1} or the dual form \eqref{equ:ETlambda}.
To overcome this issue, we propose in this section an efficient variant called unbalanced Sobolev transport (UST) distance. We further derive a novel closed-form formula which allows a fast computation for the proposed transport distance, especially for large-scale settings.

Our strategy in defining the UST is based on the dual formulation \eqref{equ:ETlambda} (in Corollary~\ref{cor:duality}) but by simultaneously relaxing the two constraints for critic function $f$ in the set $\mathbb{U}$. This approach is partially adopted in \citep{le2021ept} 
for the EPT problem for measures on a \emph{tree}, but they only relax the first corresponding constraint for $f$ in the set $\mathbb U$ (i.e., the \emph{bounded constraint} for the critic function $f$). However, keeping the \emph{Lipschitz  constraint} for $f$ limits the approach in \citep{le2021ept} to be extended to more general structures rather than tree structure (e.g., graph structure). We note that the Lipschitz  constraint is about bounding the derivative of $f$ and hence it is more fundamental and relevant than the first constraint. In this paper, we propose to also relax the Lipschitz constraint by leveraging a  notion of Sobolev functions. This approach relies on the following concept
of derivatives for functions on graphs introduced by~\cite{le2022st}, which can be viewed as a generalized version of the fundamental theorem of calculus for a \emph{graph}.

\begin{definition}[Graph-based Sobolev space~\citep{le2022st}] \label{def:Sobolev}
Let $\omega$ be a nonnegative Borel measure on $\G$, and let  $1\leq p\leq \infty$. A continuous function $f: \G \to \R$ is in the Sobolev space $W^{1,p}(\G, \omega)$ if there exists a function $h\in L^p( \G, \omega) $ satisfying
\[
f(x) -f(z_0) =\int_{[z_0,x]} h(y) \omega(\mathrm{d}y), \forall x\in \G.
\]
Such function  $h$ is unique in $L^p(\G, \omega) $ and is called the graph derivative of $f$ w.r.t.~the measure $\omega$. Hereafter, this graph derivative of $f$ is denoted by $f'$.
\end{definition}

From  Definition~\ref{def:Sobolev} and the property of $L^p(\G, \omega) $ space, we have 
\[
W^{1,p_2}(\G, \omega)\subset W^{1,p_1}(\G, \omega),
\]
whenever $1 \leq p_1 \leq p_2\leq \infty$. In particular, $W^{1,\infty}(\G, \omega)$  is the smallest space and $W^{1,1}(\G, \omega)$ is the largest space. Additionally, we prove that  $W^{1,\infty}(\G, \omega^*)$ contains the space of all Lipschitz continuous functions, and both spaces coincide when $\G$ is a tree (see Lemma~\ref{lm:lipschitz-vs-sobolev} in Appendix~\S\ref{app:subsec:furthertheoreticalresults} for the detail). Hereafter, let $\omega^*$ denote the length measure on $\G$ as defined in \cite[\S4.1]{le2022st} (see Appendix~\S \ref{app:subsec:review} for a review). We propose to regularize the transport $\mathrm{ET}_\lambda$ in \eqref{equ:ETlambda} by relaxing the constraint set $\mathbb{U}$ for critic function $f$ in two ways:

\textbf{$\bullet$} Firstly, we replace the \emph{Lipschitz condition} for the critic function $f$ in the set $\mathbb U$ (in Corollary~\ref{cor:duality}) by instead considering this constraint in the graph-based Sobolev space, i.e., $f \in W^{1,p'}(\G, \omega)$ with $\|f'\|_{L^{p'}(\G, \omega)}\leq b$. This has the following advantages: (i) we can enlarge the constraint set on the Sobolev space $W^{1,p'}(\G, \omega)$ by decreasing the value of parameter $p'$; (ii) we can vary the constraint set by choosing a suitable measure $\omega$ on $\G$. The measure $\omega$ can be interpreted as a cost of moving a unit mass from one location to another, and this cost is the same as the graph metric $d_\G$ when $\omega$ is chosen as the length measure $\omega^*$  of $\G$. Even when $p=1$ and $\omega=\omega^*$, this relaxation viewpoint still has the fundamental benefit: it allows us to extend most of the main results in \citep{le2021ept} for \emph{tree} structure to \emph{graph} structure. 

We emphasize that extending the approach in \citep{le2021ept} (i.e., EPT problem for measures on a \emph{tree}) to EPT problem for measures on a \emph{graph} $\G$ is problematic. In this special case, we know from Lemma \ref{lm:lipschitz-vs-sobolev} (Appendix~\S\ref{app:subsec:furthertheoreticalresults}) that our corresponding \emph{Sobolev constraint} is equivalent the \emph{Lipschitz constraint} when $\G$ is a tree. However, Lemma~\ref{lm:lipschitz-vs-sobolev} also implies that the Sobolev constraint set is possibly \emph{larger} for a general graph $\G$. This flexibility of Sobolev functions enables us to overcome the limitation of the approach in \citep{le2021ept} (i.e., for a \emph{tree} structure) and gives us an effective way to  exploit the \emph{graph} structure by working with critic function $f$ of a specific form in Sobolev space (see Definition~\ref{def:Sobolev}). Our obtained results in this section reveal that \emph{critic of Sobolev type in the sense of Definition~\ref{def:Sobolev} is more suitable for EPT problem for measures on a \emph{graph} than critic of the Lipschitz type}. 

\textbf{$\bullet$} Secondly, we relax the first condition for $f$ in the set $\mathbb{U}$ (i.e., the \emph{bounded constraint} for the critic function $f$) by using the following observation. According to Definition~\ref{def:Sobolev},  any function $f\in W^{1,p'}(\G, \omega)$ can be represented as
\[
f(x)= f(z_0) +  \int_{[z_0,x]} f'(y) \omega(dy).
\]
If in addition $\|f'\|_{L^{p'}(\G, \omega)}\leq b$, then by H\"older inequality, the second term on the right hand side  is controlled 
by $b \, \omega\big([z_0, x]\big)^\frac1p$.
Thus, instead of requiring
\[
-w_2(x) - \frac{b\lambda}{2}\leq f(x) \leq  w_1 (x)+ \frac{b\lambda}{2}, \quad \forall x\in\G
\]
 as in the definition of $\mathbb{U}$, we suggest to constrain only the first term $f(z_0)$. 

Putting these two ways of regularization together, we propose to consider the following constraint set $\mathbb U_{p'}^\alpha$ as a relaxation of the constraint set $\mathbb U$ for the critic function $f$ in Corollary~\ref{cor:duality}. Note that the choice of  $\alpha\!=\!0$ corresponds to our above discussion. Here, we generalize our theoretical development for a more general $\alpha$ to allow an extra degree of freedom which might be potentially useful in practical applications, e.g., by tuning $\alpha$ for further improvement.
\begin{definition}[The regularized set $\mathbb U_{p'}^\alpha$ for critic function]\label{def:regSetU}
For $1\leq p\leq \infty$ and $0\leq \alpha\leq \frac12 [b\lambda + w_1(z_0) + w_2(z_0)]$,  let  $\mathbb U_{p'}^{\alpha}$ be  the collection of all functions $f\in W^{1,p'}(\G, \omega)$ satisfying 
\[
f(z_0) \in I_\alpha\Let \Big[  -w_2(z_0)- \frac{b\lambda}{2}+\alpha, w_1(z_0) + \frac{b\lambda}{2} -\alpha\Big]
\]
and 
\[
\|f'\|_{L^{p'}(\G, \omega)}\leq b.
\]
Equivalently,  $\mathbb U_{p'}^{\alpha}$ is  the collection of all functions $f$ of the form
 \begin{equation}\label{alternative_representation}
  f(x)= s +  \int_{[z_0,x]} h(y) \omega(\dd y)
 \end{equation}
with $s\in I_\alpha$ and with $h:\G \to \R$ being some function satisfying 
\[
\|h\|_{L^{p'}(\G, \omega)}\leq b.
\]
\end{definition}
It is clear from Definition~\ref{def:regSetU} that $\mathbb U \subset \mathbb U_{p'}^0$ (see Corollary~\ref{cor:duality} for set $\mathbb U$). The requirement  $ \alpha\leq \frac12 [b\lambda + w_1(z_0) + w_2(z_0)]$ is  to ensure that the interval $I_\alpha$ is nonempty. By constraining critic  $f$ to the relaxed set $\mathbb U_{p'}^\alpha$ and noting that the last term in \eqref{equ:ETlambda} is simply a constant depending on the total masses of $\mu$ and $\nu$, we propose the following regularization of the transport $\mathrm{ET}_\lambda$ in Corollary~\ref{cor:duality}, namely \emph{unbalanced Sobolev transport} (UST).
\begin{definition}[Unbalanced Sobolev transport]\label{def:discrepancy}
Let $\omega$ be a nonnegative Borel measure on graph $\G$. Given $1\leq p\leq \infty$ and $0\leq \alpha\leq \frac12 [b\lambda + w_1(z_0) + w_2(z_0)]$. For $\mu, \nu\in \calM(\G)$, the unbalanced Sobolev transport is defined as follow 
\[
\mathrm{US}_p^\alpha(\mu,\nu ) \Let
 \sup_{f\in \mathbb U_{p'}^{\alpha}} \Big[\int_\G f(x) \mu(\mathrm{d}x) - \int_\G f(x) \nu(\mathrm{d}x)\Big]. 
\]
\end{definition}
The measure $\omega$ used for representing critic  $f$ in $U_{p'}^{\alpha}$ (see \eqref{alternative_representation}) acts as the ground cost of moving masses on graph $\G$ from one location to another. Especially, when $\omega$ is chosen as the length measure $\omega^*$ of graph $\G$, we have $\omega([x,y]) = d_\G(x,y)$ (see Lemma~\ref{lem:length-measure} in Appendix~\S\ref{app:subsec:review}). 

We then show the connection between $1$-order UST and the dual formulation of EPT on graph $\G$ with the \emph{Lipschitz constraint}, but the \emph{bounded constraint} only applied on the critic function at root node $z_0$. Precisely, we obtain:
\begin{lemma}\label{lm:lower-bound-part1}
Recall that $\omega^*$ be the length measure of graph $\G$. For $\omega = \omega^*$,  we have 
\begin{eqnarray}\label{sobolev-vs-lipschitz}
\mathrm{US}_1^0(\mu,\nu ) \geq  \sup \Big[ \int_\G f (\mu - \nu):\, f\in \mathbb{U}_0  \Big]
\end{eqnarray}
where $ \mathbb{U}_0 \Let \Big\{f\in C(\G):\, 
  -w_2(z_0) - \frac{b\lambda}{2}\leq f(z_0) \leq    w_1(z_0) + \frac{b\lambda}{2}, \, |f(x)-f(y) |\leq b \, d_\G(x,y)\Big\}$.
  Moreover, the inequality in \eqref{sobolev-vs-lipschitz} becomes the equality if  $\G$ is a tree.
\end{lemma}

We next state our fundamental result, which demonstrates that the proposed UST (Definition~\ref{def:discrepancy}) for measures with unequal mass on a \emph{graph} is computationally effective. We in fact obtain a closed-form formula for UST in terms of an integral explicitly depending on the input measures. This yields a substantial  computational advantage in comparison with the EPT approach for unbalanced measures on a graph (i.e., $\mathrm{ET}_\lambda$) which requires to solve sophisticated optimization problems either in the primal \eqref{P1} or its dual \eqref{equ:ETlambda}. To our knowledge, \emph{the proposed UST is the \emph{first} approach which yields a closed-form solution among available variants of unbalanced OT for measures with unequal mass on a \emph{graph}}.

\begin{proposition}\label{prop:closed-form}
Let $\omega$ be a nonnegative measure on graph $\G$. Let  $1\leq p\leq \infty$  and  $0\leq \alpha\leq \frac12 [b\lambda + w_1(z_0) + w_2(z_0)]$.  Then, for two input measures $\mu,\nu\in \calM(\G)$, we have 
\begin{eqnarray*}
& \hspace{-3.5em}\mathrm{US}_p^\alpha(\mu,\nu ) = b\, \Big[\int_{\G} | \mu(\Lambda(x)) -  \nu(\Lambda(x))|^p \, \omega(\dd x)\Big]^\frac{1}{p} \\
& \hspace{14.5em} + \, \Theta |\mu(\G)-\nu(\G)|,
\end{eqnarray*}
where $\Lambda(x)$ is defined by \eqref{sub-graph} and 
\begin{align}\label{def:M}
\hspace{-1em} \Theta \Let  \left\{\begin{array}{lr}
 w_1(z_0) +\frac{b\lambda}{2} -\alpha \hspace{1.5em} &\mbox{if}\quad\mu(\G)\geq \nu(\G),\\
w_2(z_0) +\frac{b\lambda}{2} -\alpha \hspace{1.5em} &\mbox{if}\quad\mu(\G)< \nu(\G).
\end{array}\right.
\end{align}
\end{proposition}
The constant $\Theta$ depends on $\mu$ and $\nu$ unless $\mu(\G) =\nu(\G)$ or $w_1(z_0) = w_2(z_0)$. The integral in the above expression can be computed explicitly and efficiently as in the following corollary  when the two input distributions are supported on nodes of the graph (i.e., the node set $V$ of graph $\G$). 

\begin{corollary}\label{cor:closed-form}
Under the same assumptions as in Proposition~\ref{prop:closed-form}
and assume in addition that $\omega(\{x\}) = 0$ for every $x\in \G$. Suppose that $\mu,\nu\in \calM(\G)$ are supported on nodes in $V$ of graph $\G$.\footnote{We discuss an extension for measures supported in $\G$ in Appendix \S\ref{lem:length-measure}.} Then, we have
\begin{eqnarray}\label{equ:closed_form_US}
& \hspace{-4.1em} \mathrm{US}_p^\alpha(\mu,\nu ) = b\, \Big(\sum_{e \in E} w_e \left|\mu(\gamma_e) - \nu(\gamma_e) \right|^{p}\Big)^{\frac{1}{p}} \nonumber \\
& \hspace{11.5em} + \,  \Theta |\mu(\G)-\nu(\G)|.
\end{eqnarray}
\end{corollary}
\begin{remark}[UST for non-physical graph]
We have assumed that $\G$ is a physical graph as in \S\ref{sec:pre}. However, Corollary~\ref{cor:closed-form} shows that the $p$-order unbalanced Sobolev transport $\mathrm{US}_p^\alpha$ does not depend on this physical assumption when input measures are supported on nodes. Precisely, it only depends on the graph structure $(V, E)$ and edge weights $w_e$. Thus, $\mathrm{US}_p^\alpha$ can be applied for non-physical graph $\G$.
\end{remark}

We next describe a preprocessing step on graph $\G$ and analyze the time complexity in computing  $\mathrm{US}_p^\alpha$ .

\textbf{Preprocessing step.} To compute $\mathrm{US}_p^\alpha$, we apply a preprocessing step to form the set $\gamma_e$ for each edge $e\in E$ in graph $\G$ by identifying shortest paths from the root node $z_0$ to other nodes (e.g., by Dijkstra algorithm with a complexity $\mathcal{O}(|E| + |V| \log{|V|})$ where $|E|, |V|$ are the numbers of egdes and nodes of graph $\G$ respectively). Especially, observe that any edge $e$ with $\gamma_e = \emptyset$ does not contribute to the computation of $\mathrm{US}_p^\alpha$. Therefore, one can remove such edge $e$ in the
summation in \eqref{equ:closed_form_US}. We emphasize that this preprocessing step only involves the graph structure itself and is independent of input measures.

\textbf{Computational complexity.} Let $E_{\mu, \nu} \Let \left\{e \in E \mid e \subset [z_0, z] \mbox{ for some $z \in$ }\text{supp}(\mu) \cup \text{supp}(\nu) \right\}$, where $\text{supp}(\mu), \text{supp}(\nu)$ are respectively the support of measures $\mu, \nu$. Then, the computational complexity of $\mathrm{US}_p^\alpha(\mu, \nu)$ is linear to the number of edges in $E_{\mu, \nu}$. 

\textbf{Related work.} 
Beyond the pure graph of supports, the metric structure inherited from the graph metric space plays an important role in our work. More precisely, an edge weight $w_e$ is considered as a cost to move a unit mass from one node to the other node of edge $e$ (i.e., graph metric distance between two edge nodes). Therefore, one should distinguish our approach with the unbalanced diffusion earth mover's distance~\citep{tong2022embedding} which uses an affinity between two edge nodes in their graph.

\textbf{$\bullet$ Relation with Sobolev transport (ST)~\citep{le2022st}.} We emphasize that ST is only valid for measures with \emph{equal} mass on a graph. It  \emph{cannot} be applied for our considered problem where input measures may have \emph{different} total mass. Even though both ST and the proposed UST are  instances of integral probability metrics (IPM), it is nontrivial to effectively extend ST for unbalanced measures on a \emph{graph} by defining a function set for the critic. The theoretical results of EPT on a \emph{graph} in \S\ref{sec:entropy_partialOT} play the fundamental role in developing our proposed UST.

\begin{remark}[The special case of balanced mass]
When input measures have the same mass, from Lemma~\ref{lm:sam-sobolev} of \S \ref{app:subsec:balancedmass}, the proposed unbalanced Sobolev transport (with $b = 1$) coincides with the balanced Sobolev transport~\citep[Definition 3.2]{le2022st}.
\end{remark}

\textbf{$\bullet$ Relation with EPT on a tree~\citep{le2021ept}.} As we discussed previously, extending the approach in~\citep{le2021ept} for EPT on a \emph{tree} to our considered problem (i.e., EPT on a \emph{graph}) is problematic. We see from Lemma~\ref{lm:lipschitz-vs-sobolev} (Appendix~\S\ref{app:subsec:furthertheoreticalresults}) and Lemma~\ref{lm:lower-bound-part1}
that the \emph{Sobolev constraint} set in our approach is possibly \emph{larger} than the \emph{Lipschitz constraint} set for a general graph $\G$, but these two constraint sets coincide  when $\G$ is a tree. 
Our results illustrate that it is more efficient to exploit \emph{graph structure} for \emph{critic of Sobolev type} (as in our approach) than \emph{critic of the Lipschitz type} (as in EPT on a tree).

\section{PROPERTIES OF UNBALANCED SOBOLEV TRANSPORT}\label{sec:Prop_US}

In this section, we derive geometric structures together with bounds for UST 
and prove its negative definiteness. Consequently, we develop positive definite kernels upon UST, required in many kernel-dependent frameworks.

We first show that $\mathrm{US}_p^\alpha$ possess the metric property. Moreover, it makes the space of measures $\calM(\G)$ a geodesic space. Thus, $(\calM(\G), \mathrm{US}_p^\alpha)$ inherits all geometric properties of the geodesic space.

\begin{proposition}[Geometric structures of $\mathrm{US}_p^\alpha$]\label{geodesic-space} 
Let $\omega$ be a nonnegative Borel measure on $\G$.
Assume that  $\lambda, w_1(z_0), w_2(z_0) \ge 0$. For $1\leq p\leq \infty$ and $0\leq \alpha< \frac{b\lambda}{2} +\min\{w_1(z_0), w_2(z_0)\}$, then we have
\begin{enumerate}
\item[i)] $\mathrm{US}_p^\alpha(\mu +\sigma,\nu +\sigma) = \mathrm{US}_p^\alpha(\mu,\nu)$, $\forall \mu, \nu, \sigma \in\calM(\G)$.

\item[ii)]  $\mathrm{US}_p^\alpha$ is a divergence\footnote{I.e., $\mathrm{US}_p^\alpha\geq 0$, and $\mathrm{US}_p^\alpha(\mu,\nu) = 0$ if and only if $\mu=\nu$.}
and satisfies the triangle inequality:
\[
\hspace{-0.8em} \mathrm{US}_p^\alpha(\mu,\nu)\leq \mathrm{US}_p^\alpha(\mu,\sigma) + \mathrm{US}_p^\alpha(\sigma, \nu), \forall \mu, \nu, \sigma \in\calM(\G).
\]

\item[iii)] If in addition $w_1(z_0)=w_2(z_0)$, then $\mathrm{US}_p^\alpha$ is a metric and  $(\calM(\G), \mathrm{US}_p^\alpha)$ is a complete metric space. 
Moreover, it is a geodesic space in the sense that for every two points $\mu$ and $\nu$ in $\calM(\G)$ there exists a  path $\varphi: [0,a ]\to \calM(\G)$ with $a\Let\mathrm{US}_p^\alpha(\mu,\nu)$ such that $\varphi(0)=\mu$, $\varphi(a)=\nu$, and
\[
\mathrm{US}_p^\alpha(\varphi(t), \varphi(s)) = |t-s|, \text{ for all } t,s\in [0,a].
\]
\end{enumerate}
\end{proposition}
In Proposition~\ref{prop:upper_for_US} (Appendix~\S\ref{app:subsec:furthertheoreticalresults}), we also establish a comparison between $\mathrm{US}_p^\alpha$
for different exponent~$p$. We next derive a lower bound for $\mathrm{US}_1^0$ in terms of $\mathrm{ET}_\lambda$. In fact, a more general estimate holds true for every $p\geq 1$ and is given in Proposition~\ref{prop:lower-more-general} (Appendix~\S\ref{app:subsec:furthertheoreticalresults}). As a consequence of Corollary~\ref{cor:duality} and  Lemma~\ref{lm:lower-bound-part1} and since $\mathbb{U} \subset \mathbb{U}_0$, we obtain:
\begin{proposition}[Lower bound for $\mathrm{US}_1^0$] \label{prop:lower}
Recall that $\omega^*$ is the length measure on $\G$. Assume that $w_1, w_2$ are $b$-Lipschitz w.r.t. $d_\G$.
For $\omega=\omega^*$, $\mu, \nu\in \calM(\G)$, we have
 \begin{equation*}\label{lower-bound-us1}
 \mathrm{US}_1^0(\mu,\nu )
 \geq   \mathrm{ET}_\lambda(\mu,\nu) + \frac{b\lambda}{2}\big[ \mu(\G) +  \nu(\G)\big].
 \end{equation*}
\end{proposition}

We emphasize that when $\G$ is a tree, our EPT on a \emph{graph} (i.e., $\mathrm{ET}_{c,\lambda}$ and $\mathrm{ET}_\lambda$) coincide with the ones defined in \citep{le2021ept}. Furthermore, we have:

\begin{proposition}[Lower bounds]\label{prop:D-via-ET}
Assume that  $\G$ is a tree  and $\omega = \omega^*$. The followings hold true:
\begin{enumerate}

\item[i)] $\mathrm{US}_1^\alpha(\mu,\nu )= d_\alpha(\mu,\nu)$. Also for $1\leq p \leq \infty$, we have
\begin{eqnarray*}
& \hspace{-8.5em}
\mathrm{US}_p^\alpha(\mu,\nu ) \geq \omega^*(\G)^{-\frac{1}{p'}  } d_\alpha(\mu,\nu ) \\
& \hspace{4.5em} + \, \Theta \left[1- \omega^*(\G)^{-\frac{1}{p'}  }\right] \left|\mu(\G)-\nu(\G)\right|,
\end{eqnarray*}
where $d_\alpha$ is defined in \cite[Eq.~(9)]{le2021ept}.
\item[ii)] If $\mu(\G) =\nu(\G)$, then for $1\leq p \leq \infty$, we have 
\[
\hspace{-1em} \mathrm{US}_p^\alpha(\mu,\nu ) \geq b \, \omega^*(\G)^{-\frac{1}{p'}} \hspace{-0.2em}\left[\sup_{x,y\in\G} d_\G(x,y)\right]^{1-p} \hspace{-0.8em} \calW_p^p(\mu,\nu),
\]
where $\calW_p$ is the $p$-order Wasserstein distance\footnote{The definition of  $\calW_p$ is recalled in Appendix~\S\ref{app:subsec:review}.} with cost $d_\G^p$. Moreover, the equality is attained when $p=1$.
\end{enumerate}
\end{proposition}

We next prove the negative definiteness for UST. This important property allows us to build positive definite kernels upon UST, required for kernel-dependent machine learning algorithmic approaches.

\begin{proposition}\label{prop:negative_definite}
Under the same assumptions as in Corollary~\ref{cor:closed-form} and $w_1(z_0)=w_2(z_0)$. Then, $\mathrm{US}_p^\alpha$ is negative definite on $\calM(\G)$ for any  $1 \le p \le 2$.
\end{proposition}

From Proposition~\ref{prop:negative_definite} and by using~\cite[Theorem 3.2.2]{Berg84}, we obtain that the kernel
\[
k_{\mathrm{US}_p^\alpha}(\mu,\nu) \Let \exp(-t \mathrm{US}_p^\alpha(\mu,\nu))
\]
is positive definite on $\calM(\G)$ for any given $t > 0$ and $1 \le p \le 2$.

\begin{figure*}
  \begin{center}
    \includegraphics[width=0.9\textwidth]{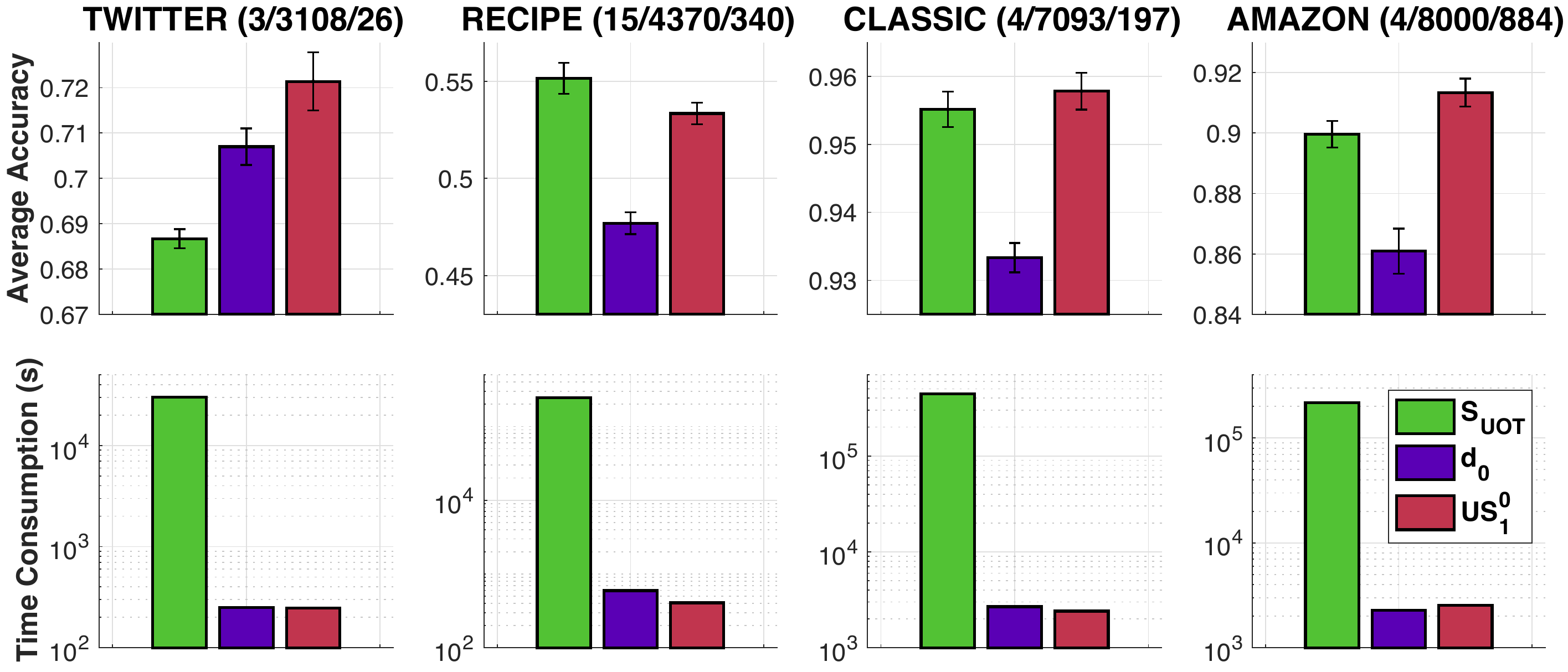}
  \end{center}
  \caption{SVM results and time consumption for kernel matrices in document classification with graph $\G_{\text{Sqrt}}$. For each dataset, the numbers in the parenthesis are the number of classes; the number of documents; and the maximum number of unique words for each document respectively.}
  \label{fg:DOC_10KSqrt_main}
\end{figure*}

\section{EXPERIMENTS}\label{sec:experiments}

In this section, we illustrate the fast computation (i.e., closed-form solution) of the proposed UST and comparable performances of the proposed positive definite kernel associated to UST against other popular unbalanced transport baselines and their corresponding kernels. More concretely, we evaluate for \emph{measures with unequal mass on a given graph} under two simulations: document classification and topological data analysis (TDA).

\textbf{Document classification.} We consider four traditional document datasets: \texttt{TWITTER}, \texttt{RECIPE}, \texttt{CLASSIC}, and \texttt{AMAZON}. Their characteristics are summarized in Figure~\ref{fg:DOC_10KSqrt_main}. We represent each document as a measure by considering each word in the document as its support with a unit mass. Therefore, \emph{documents with different lengths have different total mass}. We employ the same word embedding procedure as in~\citep{le2021ept} to embed words into vectors in $\R^{300}$.

\begin{figure}[h]
  \begin{center}
    \includegraphics[width=0.45\textwidth]{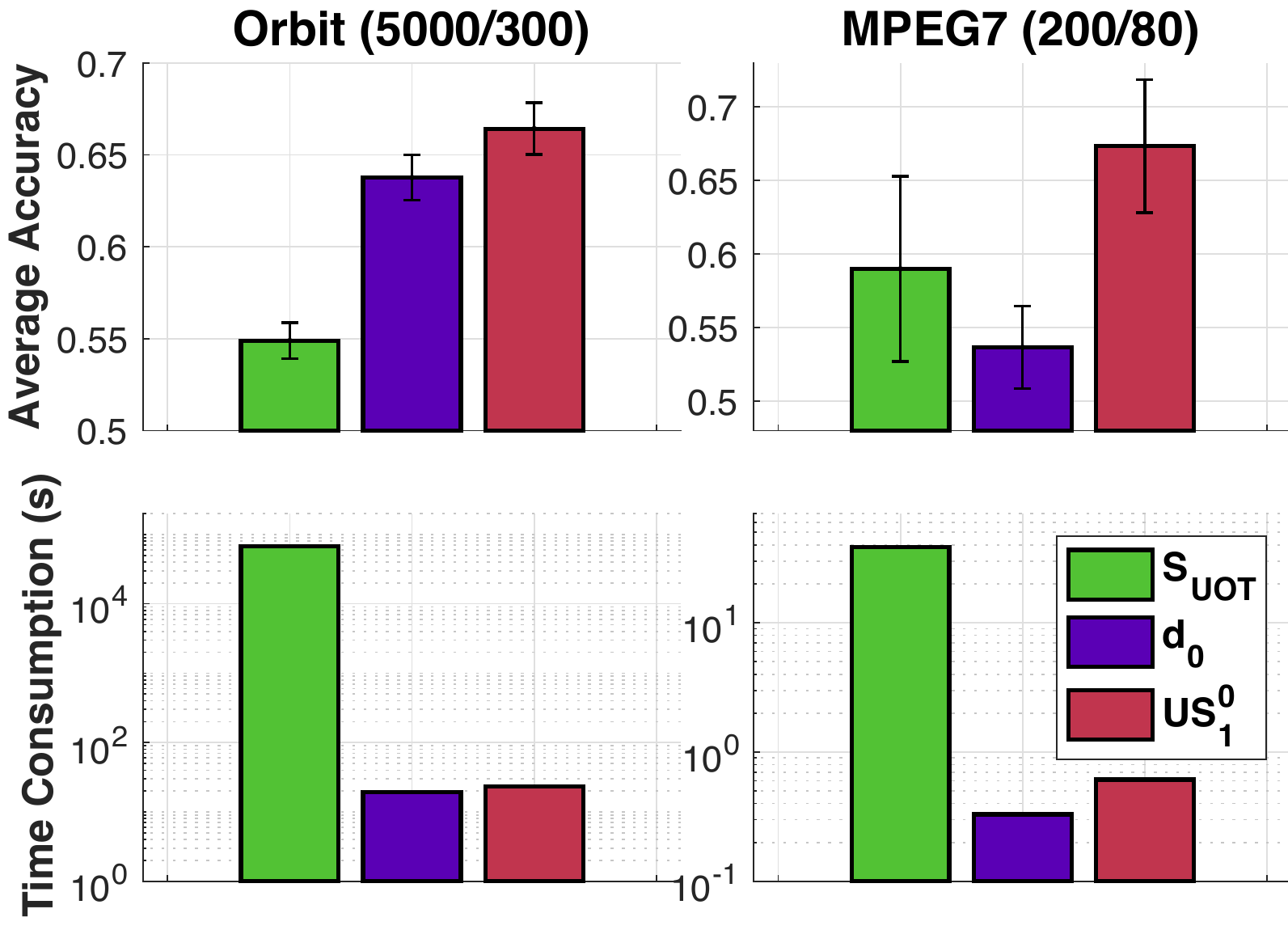}
  \end{center}
  \caption{SVM results and time consumption for kernel matrices in TDA with graph $\G_{\text{Sqrt}}$. For each dataset, the numbers in the parenthesis are respectively the number of PD; and the maximum number of points in PD.}
  \label{fg:TDA_mix1K100Sqrt_main}
\end{figure}

\textbf{TDA.} We carry out two tasks: orbit recognition on \texttt{Orbit} dataset and object shape recognition on \texttt{MPEG7} dataset. For \texttt{Orbit} dataset, it is synthesized as in \citep{adams2017persistence} for link twist map which are discrete dynamical systems to model flows in DNA microarrays~\citep{hertzsch2007dna}. There are five classes of orbits in the dataset. For each class, we generated $1000$ orbits where each orbit contains $1000$ points. For $\texttt{MPEG7}$ dataset~\citep{latecki2000shape}, we consider its 10-class subset where each class has $20$ samples as in \citep{le2018persistence}. The characteristics of the considered \texttt{Orbit} and \texttt{MPEG7} datasets are summarized in Figure~\ref{fg:TDA_mix1K100Sqrt_main}. We use the same procedure as in~\citep{le2021ept} to extract persistence diagram (PD) for orbits and object shapes. PD are multisets of points in $\R^2$. Each point in PD summarizes the lifespan (i.e., birth and death time) of a topological feature (e.g., connected component, ring, cavity). We represent each PD as a measure by regarding each $2$-dimensional point in PD as its support with a unit mass. Consequently, \emph{persistence diagrams having a different number of topological features are represented as measures with different total mass}.

Notice that supports in document classification simulations are in high-dimensional spaces (i.e., in $\R^{300}$) while supports in TDA simulations are in low-dimensional spaces (i.e., in $\R^2$). Therefore, we can observe the effects of dimensions to the proposed UST and other unbalanced transport baselines from these simulations. We next describe various graph settings (i.e., the assumed graph metric spaces for measures) considered in our experiments.

\textbf{Graph settings.} We use the same graph settings (i.e., $\G_{\text{Log}}$ and $\G_{\text{Sqrt}}$) employed  in~\citep[\S5]{le2022st} for our simulations on document classification and TDA. For these graphs, we consider the number of nodes: $M \!=\! 10^2, 10^3, 10^4, 4\!\times\!10^4$. We note that these graphs satisfy the assumptions in \S\ref{sec:pre}. Similar to the observations in~\citep{le2022st}, each node in these graphs has a high probability to satisfy the root node condition, i.e., the uniqueness property of the shortest path (see Appendix \S\ref{app:subsec:discussion} for a further discussion).

\textbf{Root node $z_0$ for UST.} The UST is defined over graph $\G$ with a root node $z_0$. From  Definition~\ref{def:Sobolev}, the root node $z_0$ imposes its own geometry by characterizing the graph derivative of functions on $\G$. To alleviate this dependency, we follow the sliced approach in~\citep{le2022st} for Sobolev transport by averaging over different choices of the root node $z_0$ in graph $\G$, which can be viewed as a sliced variant for UST.

\begin{figure}[h]
  \begin{center}
    \includegraphics[width=0.45\textwidth]{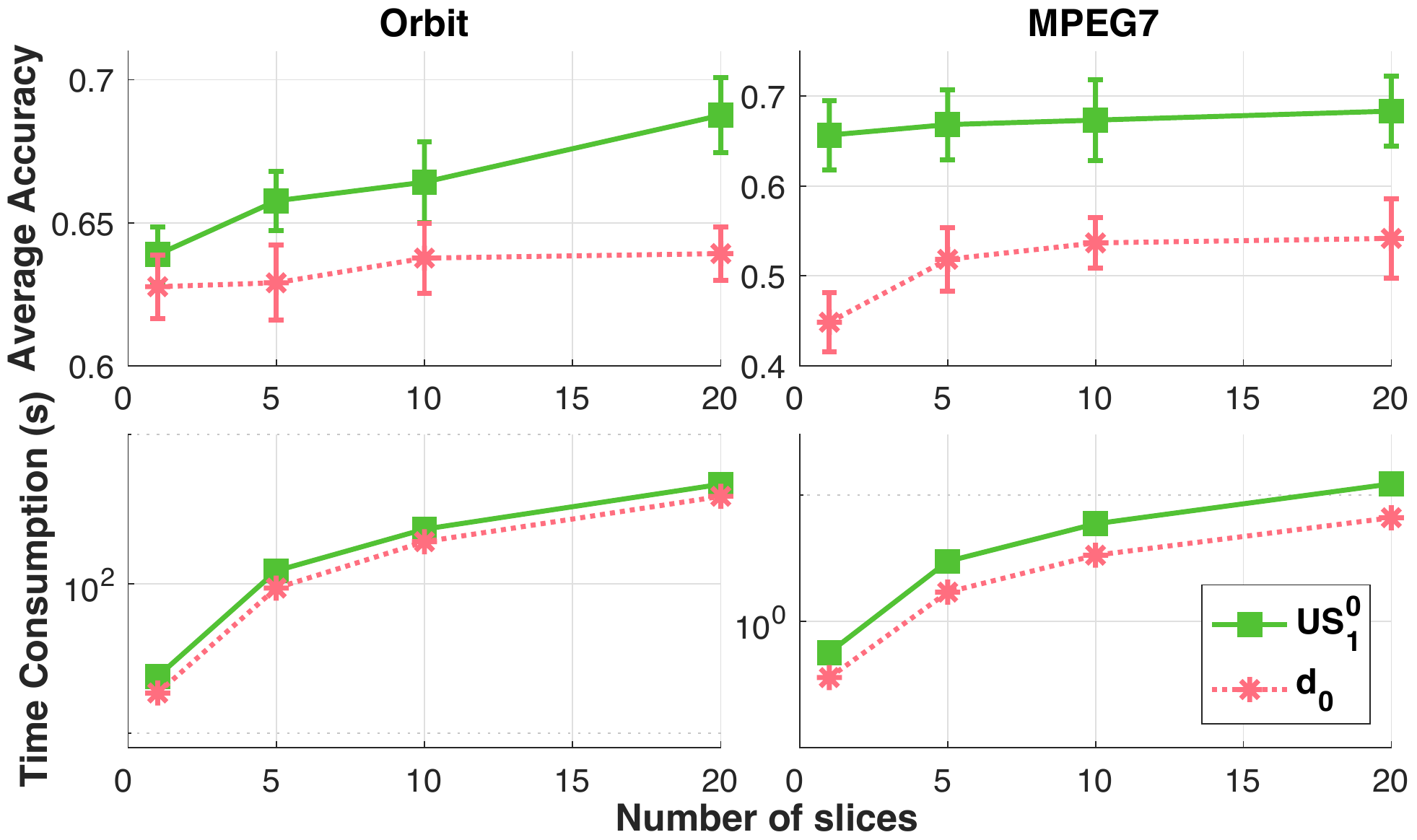}
  \end{center}
  \caption{SVM results and time consumption for kernel matrices of slice variants in TDA with graph $\G_{\text{Sqrt}}$.}
  \label{fg:TDA_10KSqrt_SLICE_main}
\end{figure}

\textbf{Baselines, and experimental setup.} We consider two typical UOT approaches for measures with unequal mass and supported on a graph metric space as baselines: (i) the Sinkhorn-based UOT~\citep{frogner2015learning, chizat2018scaling} ($\mathrm{S}_{\mathrm{UOT}}$)\footnote{\cite{sejourne2019sinkhorn} derived a debiased version for $\mathrm{S}_{\mathrm{UOT}}$ which may be helpful in applications. The debiased version is also empirically indefinite and has the same complexity as $\mathrm{S}_{\mathrm{UOT}}$. } with a graph metric ground cost, and (ii) the distance $d_{\alpha}$ of EPT on a \emph{tree}~\cite[Eq. (9)]{le2022st} (see Proposition~\ref{prop:D-via-ET} for its relation with $\mathrm{US}_p^{\alpha}$) where the tree structures are randomly sampled from graph $\G$. From results in Lemma~\ref{lm:lower-bound-part1} and Proposition~\ref{prop:lower} and for simplicity, we consider $\alpha=0$ and $p=1$ (and $d_0$ for EPT on a \emph{tree} as in~\citep{le2021ept})\footnote{One may tune these parameters for further improvements.}. We further note that there are different approaches for simulations on document classification and TDA. However, that is \emph{not} the purpose of our empirical simulations which compare different unbalanced transports for measures with unequal mass on a graph in the same settings. 

We apply the kernel approach in the form $\exp(-t \bar{d})$, where $\bar{d}$ is a discrepancy for unbalanced measures on a \emph{graph} and $t>0$, with support vector machines (SVM) for the simulations on document classification and TDA. Note that kernels for $\mathrm{US}_p^{\alpha}$ and $d_{\alpha}$ are positive definite, but kernels for $\mathrm{S}_{\mathrm{UOT}}$ is empirically indefinite (see \cite[\S8.3]{peyre2019computational}). Similar to~\citep{le2021ept}, we regularized the Gram matrices for kernels with $\mathrm{S}_{\mathrm{UOT}}$ by adding a sufficiently large diagonal term.

For simplicity, we employ the same setup for the EPT problem in~\citep{le2021ept}, i.e., using $\lambda \!=\! b \!=\! 1$ for the EPT. From Corollary~\ref{cor:closed-form} and Proposition~\ref{prop:negative_definite}, we consider the weight functions $w_1(x) \!=\! w_2(x) \!=\! a_1 d_{\G}(z_0, x) + a_0$ where $a_1\!=\! b$ and $a_0\!=\!1$.

For kernel SVM, we use the same setting as in~\citep{le2021ept}. In each dataset, we randomly split it into $70\%/30\%$ for training and test with $10$ repeats. We use 1-vs-1 strategy for SVM with multiclass data. Hyperparameters are typically chosen by cross validation. For kernel hyperparameter, we choose $1/t$ from $\{q_{s}, 2q_{s}, 5q_{s}\}$ with $s \!=\! 10, 20, \dotsc, 90$ where $q_s$ is the $s\%$ quantile of a random subset of corresponding distances on training data. For SVM regularization hyperparameter, we choose it from $\left\{0.01, 0.1, 1, 10, 100\right\}$. For $\mathrm{S}_{\mathrm{UOT}}$, we choose the entropic regularization from $\left\{0.01, 0.1, 1, 10\right\}$. The reported time consumption for each kernel matrices also includes the corresponding preprocessing, e.g., compute shortest paths on graph $\G$ for $\mathrm{US}_p^{\alpha}$ and $\mathrm{S}_{\mathrm{UOT}}$, or sampling random tree structures from $\G$ for $d_{\alpha}$ of EPT on a tree.

\textbf{Results of SVM, time consumption and discussions.} We illustrate the SVM results and time consumption for kernel matrices for document classification and TDA in Figure~\ref{fg:DOC_10KSqrt_main} and Figure~\ref{fg:TDA_mix1K100Sqrt_main} with $M\!=\!10^4$ for document datasets, $M\!=\!10^3$ for \texttt{Orbit} and $M\!=\!10^2$ for \texttt{MPEG7} for graph $\G_{\text{Sqrt}}$. The performances of kernels for our proposed UST compare favorably with other approaches (except $\mathrm{S}_{\mathrm{UOT}}$ on \texttt{RECIPE}). Additionally, the time consumption of $\mathrm{US}_1^0$ and $d_0$ is several-order faster than that of $\mathrm{S}_{\mathrm{UOT}}$. Recall that kernels for $\mathrm{S}_{\mathrm{UOT}}$ is indefinite, which may affect performances of $\mathrm{S}_{\mathrm{UOT}}$ in some datasets (e.g., \texttt{Orbit}, \texttt{TWITTER}). In Figure~\ref{fg:TDA_10KSqrt_SLICE_main}, we illustrate the effects of the number of slices (i.e., the number of root nodes used for averaging) for $\mathrm{US}_1^0$ and $d_{0}$ for TDA. Generally, performances of those approaches are improved with more slices but with a trade-off on time consumption. We observe that $10$ slices give a good trade-off in applications. Extensive further empirical results can be seen in Appendix~\S \ref{app:sec:furtherempiricalresults}, e.g., for various graph structures, graph sizes $M$, and different orders $p$ of UST.

\section{CONCLUSION}\label{sec:conclusion}

In this work, we proposed unbalanced Sobolev transport (UST) 
for measures with unequal mass on a \emph{graph}.
UST is the \emph{first} variant of UOT having a \emph{closed-form} formula for a fast computation. Additionally, UST is negative definite which allows to build positive definite kernels, required for kernel-dependent frameworks. Since UST exploits the graph metric structure of supports, it may restrict to applications with prior graph structures, or applications where one can build graphs from supports. On the other hand, we have not forseen any negative societal impacts of our work.

\section*{Acknowledgements}
We thank anonymous reviewers and area chairs for their comments. KF has been supported in part by Grant-in-Aid for Transformative Research Areas (A) 22H05106. The research of TN is supported in part by a grant from the Simons Foundation ($\#318995$). TL gratefully acknowledges the support of JSPS KAKENHI Grant number 20K19873. Finally, this research was enabled in part by computational support provided by Makoto Yamada.


\balance
\bibliographystyle{apalike}
\bibliography{bibEPT21, bibSobolev22}


\appendix
\onecolumn



In the appendix, we give further theoretical results and detailed proofs in \S \ref{app:theory_proof}. Additionally, we also give brief reviews about important definitions used in our work, additional discussions and further empirical results in \S \ref{app:results_discussions}.

\paragraph{Notations.} Besides the notations in the main manuscript, we further denote $\langle x_1, x_2\rangle$ as the line segment in $\R^n$ connecting two points $x_1, x_2$ and $( x_1, x_2)$ as the same line segment but without its two end-points.

\section{PROOFS AND ADDITIONAL THEORETICAL RESULTS}\label{app:theory_proof}

In this section, we give detailed proofs for the theoretical results in the main manuscript. We also provide some additional results for the unbalanced Sobolev transport (UST).

\subsection{Further Theoretical Results}\label{app:subsec:furthertheoreticalresults}
We include here  some additional results for the transport problems and the 
unbalanced Sobolev transport $\mathrm{US}_p^\alpha$.
 
\subsubsection{The Connection between Problem  \eqref{original} and Problem \eqref{P1}}
We show the connection between problem  \eqref{original} and problem \eqref{P1} for EPT on a \emph{graph} by following a similar reasoning as EPT on a \emph{tree}~\citep{le2021ept}. It is a direct extension of results in~\citep{le2021ept}.
\begin{theorem}\label{thm:m-via-lambda} Let 
$H(\lambda) \Let -\mathrm{ET}_{c,\lambda}(\mu,\nu)$ for $\lambda\in\R$, and denote 
\[
\partial H(\lambda) \Let \Big\{p\in \R: H(t)\geq H(\lambda) + p(t-\lambda), \forall t\in\R \Big\}
\]
for the set of all subgradients of $H$ at $\lambda$. Also,  set $\partial H(\R) \Let\cup_{\lambda\in \R} \partial H(\lambda)$. Then, we have 
\begin{itemize}
\item[i)] $H$ is a convex function on $\R$, and 
 \[
 \partial H(\lambda) =\big\{ b\, \gamma(\G\times\G): \gamma\in \Gamma^0(\lambda)\big\}
 \quad \forall \lambda\in\R,
 \]
 where we write $\Gamma^0$ for a set of all optimal plans $\gamma$. Also if $\lambda_1<\lambda_2$, then $m_1 \leq m_2$ for every $m_1\in \partial H(\lambda_1)$ and $m_2\in \partial H(\lambda_2)$. 


\item[ii)] $H$ is differentiable at $\lambda$ if and only if every optimal plan in $\Gamma^0(\lambda)$ has the same mass. When this happens, we also have 
\[
H'(\lambda) =b \, \gamma(\G\times\G),
\]
for any  $\gamma\in \Gamma^0(\lambda)$.

\item[iii)] If there exists a constant $M>0$ such that 
\[
w_1(x) +w_2(y) \leq b\, [ c(x,y) + M],
\]
for all $x,y\in \G$, then  $\partial H(\R)=[0,b\, \bar m]$. Moreover, \[
H(\lambda)=-\int_\G  w_1 \mu(\dd x) 
- \int_\G  w_2  \nu(\dd x),
\]
when $\lambda<-M$, and 
$H'(\lambda)=b\, \bar m $ 
for $\lambda> \|c\|_{L^\infty(\G\times\G)}$.   
\end{itemize}
\end{theorem}

The proof is placed in \S\ref{proof:thm:m-via-lambda}.

For any $m\in [0, \bar m]$,  part iii) of  Theorem~\ref{thm:m-via-lambda} implies  that there exists $\lambda\in \R$ such that $b\, m \in \partial H(\lambda)$. It then follows from part i) of this theorem  that  $m=  \gamma^*(\G\times\G)$ for some $\gamma^*\in \Gamma^0(\lambda)$. It is also clear that this $\gamma^*$ is an optimal plan for $ \mathrm W_{c,m}(\mu,\nu)$, and
\begin{align*} 
&\mathrm W_{c,m}(\mu,\nu)=\mathrm{ET}_{c,\lambda}(\mu,\nu) +\lambda b\,  m.
\end{align*}
Thus solving the auxiliary problem \eqref{P1} gives us a solution  to  the original problem  \eqref{original}.
When $H$ is differentiable, the relation between $m$ and $\lambda$ is given explicitly as 
\[
H'(\lambda)=b \, m.
\]
Note that the above selection of $\lambda$ is unique only if the function $H$ is strictly convex. Nevertheless, it enjoys the following monotonicity regardless of the uniqueness: if $m_1< m_2$, then $\lambda_1 \leq \lambda_2$. Indeed, we have  $m_1=  \gamma^1(\G\times\G)$ and $m_2=  \gamma^2(\G\times\G)$ for some $\gamma^1\in \Gamma^0(\lambda_1)$ and $\gamma^2\in \Gamma^0(\lambda_2)$. Since $\gamma^1(\G\times\G)<\gamma^2(\G\times\G)$, one has  $\lambda_1\leq \lambda_2$ by i) of Theorem~\ref{thm:m-via-lambda}.

\subsubsection{$W^{1,\infty}(\G, \omega^*)$ versus  Lipschitz space}\label{1infinity-via-lipschitz}
 We describe the connection between 
the Sobolev space $W^{1,\infty}(\G, \omega^*)$ and the space of Lipschitz continuous functions. The definition of the length measure $\omega^*$  is reviewed in \S\ref{sec:length_measure}).  
\begin{lemma}\label{lm:lipschitz-vs-sobolev}
Let $\omega^*$ be the length measure on  graph $\G$, and let $f: \G \to \R$ be a function.  We have:
\begin{enumerate}
    \item[i)]  If $
   |f(x)-f(y) |\leq b \, d_\G(x,y)$,  $\forall x,\, y\in\G$, then $f\in W^{1,\infty}(\G, \omega^*)$ with 
  $\|f'\|_{L^{\infty}(\G, \omega^*)}\leq b$.
  \item[ii)] Assume in addition that $\G$ is a tree. Then, $f\in W^{1,\infty}(\G, \omega^*)$ with 
  $\|f'\|_{L^{\infty}(\G, \omega^*)}\leq b$ implies that
  $
   |f(x)-f(y) |\leq b \, d_\G(x,y)$ for every $x,\, y\in\G$.
\end{enumerate}
\end{lemma}

The proof is placed in \S\ref{proof:1infinity-via-lipschitz}.

\begin{remark}\label{general-measure}
Our proof for Lemma~\ref{lm:lipschitz-vs-sobolev} (in \S\ref{proof:1infinity-via-lipschitz}) also shows that the result in part ii) of Lemma~\ref{lm:lipschitz-vs-sobolev} in fact holds for every measure $\omega$. Precisely, let $\omega$ be a nonnegative Borel measure  on a tree $\G$. Then, we have $f\in W^{1,\infty}(\G, \omega)$ with 
  $\|f'\|_{L^{\infty}(\G, \omega)}\leq b$ implies that
  $
   |f(x)-f(y) |\leq b \, \omega([x,y])$ for every $x,\, y\in\G$.
\end{remark}

\subsubsection{Comparison between Sobolev Spaces with Diferent Exponents}\label{sec:different-p}
We derive a comparison between UST with different exponent~$p$, and its proof is a direct consequence of our closed-form formula given in Proposition~\ref{prop:closed-form}.
\begin{proposition}[Relation for different $p$] \label{prop:upper_for_US}
Assume that $\omega$ is a  nonnegative Borel measure on $\G$. 
Then for any $1\leq p \leq  q\leq \infty$ and $\mu, \nu\in \calM(\G)$,  we have
\[
 \mathrm{US}_p^\alpha(\mu,\nu ) - \Theta|\mu(\G)-\nu(\G)|
\leq \omega(\G)^{\frac{1}{p} - \frac{1}{q} } \, \, \Big[\mathrm{US}_q^\alpha(\mu,\nu ) -  \Theta|\mu(\G)-\nu(\G)|\Big],
\]
where $\Theta$ is the constant defined by \eqref{def:M}.
\end{proposition}
\begin{proof}[Proof of Proposition~\ref{prop:upper_for_US}]
The case $p=q$ is trivial, so let us consider $1\leq p < q\leq \infty$. Then by
using Proposition~\ref{prop:closed-form} and H\"older's inequality, we obtain
\begin{align*}
  \mathrm{US}_p^\alpha(\mu,\nu ) - \Theta|\mu(\G)-\nu(\G)|
 &= b\, \left(\int_{\G} | \mu(\Lambda(x)) -  \nu(\Lambda(x))|^p \, \omega(\dd x)\right)^\frac{1}{p}\\
 &\leq b\, \omega(\G)^{\frac{1}{p} - \frac{1}{q} } \left(\int_{\G} | \mu(\Lambda(x)) -  \nu(\Lambda(x))|^q \, \omega(\dd x)\right)^\frac{1}{q}\\
 &= \omega(\G)^{\frac{1}{p} - \frac{1}{q} } \Big[\mathrm{US}_q^\alpha(\mu,\nu ) -  \Theta|\mu(\G)-\nu(\G)|\Big].
\end{align*}
\end{proof}

\subsubsection{Lower Bound for $\mathrm{US}_p^0$}

We derive a lower bound for $\mathrm{US}_p^0$ which is a generalization of the result for $p=1$ in Proposition~\ref{prop:lower}.
\begin{proposition}[Lower bound for $\mathrm{US}_p^0$] \label{prop:lower-more-general}
Let  $\omega^*$ be the length measure    on $\G$, and assume that $w_1$ and $w_2$ are $b$-Lipschitz w.r.t. $d_\G$.
Then by taking $\omega=\omega^*$, we have for every $1\leq p\leq \infty$ that
 \begin{equation*}
 \mathrm{US}_p^0(\mu,\nu )
 \geq \omega^*(\G)^{-\frac{1}{p'}  } \Big\{  \mathrm{ET}_\lambda(\mu,\nu) + \frac{b\lambda}{2}\big[ \mu(\G) +  \nu(\G)\big]  \Big\} + \Theta [1- \omega^*(\G)^{-\frac{1}{p'}  }] |\mu(\G)-\nu(\G)|
 \end{equation*}
 for every  $\mu, \, \nu\in \calM(\G)$. Here $\Theta$ is the constant defined by \eqref{def:M}.
\end{proposition}
\begin{proof}
This is a consequence of Corollary~\ref{cor:duality},  Lemma~\ref{lm:lower-bound-part1}, and Proposition~\ref{prop:upper_for_US}.
\end{proof}

\subsubsection{The Special Case of Balanced Mass}\label{app:subsec:balancedmass}
Observe that for the case $\mu(\G) = \nu(\G)$, the constraint $f(z_0) \in I_\alpha$ in the definition of $\mathbb U_{p'}^\alpha$
is redundant. Indeed, we have:

\begin{lemma}\label{lm:sam-sobolev}
Let $\omega$ be a nonnegative Borel measure on $\G$. Assume that $\mu, \nu\in \calM(\G)$ satisfy $\mu(\G) =\nu(\G)$. Then,
 \begin{equation*}
  \mathrm{US}_p^\alpha(\mu,\nu)=\sup \Big\{ \int_\G f(\mu-\nu): \, f\in W^{1,p'}(\G, \omega),\, 
   \|f'\|_{L^{p'}(\G, \omega)}\leq b\Big\}. 
 \end{equation*}
In particular, $\mathrm{US}_p^\alpha(\mu,\nu)$ is independent of the parameters $\alpha$, $\lambda$ and the weights $w_1$, $w_2$.
\end{lemma}
\begin{proof}
This follows from the fact that Definition~\ref{def:discrepancy}
is unchanged in the case $\mu(\G) =\nu(\G)$ when the critic function $f$ is translated by a constant.
\end{proof}
From Lemma~\ref{lm:sam-sobolev}, we see that for the case $\mu(\G) =\nu(\G)$, our proposed unbalanced Sobolev transport $\mathrm{US}_p^\alpha$ with $b=1$ coincides with the balanced Sobolev transport $\calS_p$ (defined in \cite[Definition~3.2]{le2022st}).

\subsubsection{Infinite Divisibility for Unbalanced Sobolev Transport Kernel}

Recall that given $t>0$ and $1 \le p \le 2$, the unbalanced Sobolev transport kernel $k_{\text{US}_p^{\alpha}}(\mu, \nu) \Let \exp(-t \text{US}_p^{\alpha}(\mu, \nu) )$ is positive definite (see \S\ref{sec:Prop_US} and Proposition~\ref{prop:negative_definite}).

For $i \in \N^*$, the kernel $k_{\text{US}_{pi}^{\alpha}}(\mu, \nu) \Let \exp(-\frac{t}{i} \text{US}_p^{\alpha}(\mu, \nu) )$ is positive definite. Additionally, $k_{\text{US}_p^{\alpha}}(\mu, \nu) = \left[ k_{\text{US}_{pi}^{\alpha}}(\mu, \nu) \right]^i$. Therefore, $k_{\text{US}_p^{\alpha}}$ is indefinitely divisible following \cite[Definition 2.6 in \S3]{Berg84}.  

Hence, one does not need to recompute the Gram matrix for unbalanced Sobolev transport kernel $k_{\text{US}_p^{\alpha}}$ for different values of $t$. Indeed, it is suffice to compute the Gram matrix of $k_{\text{US}_p^{\alpha}}$ once for some fixed $t$ and leverage its indefinite divisibility for other values of $t$.

\subsection{Detailed Proofs}\label{app:subsec:detailedproofs}
In this section, we give  detailed proofs for our theoretical results.

\subsubsection{Proof of Theorem~\ref{thm:m-via-lambda}}\label{proof:thm:m-via-lambda}

\begin{proof}[Proof of Theorem~\ref{thm:m-via-lambda}]
We employ a similar reasoning for EPT on a tree~\citep{le2021ept} to prove the relation between problem  \eqref{original} and problem \eqref{P1} for EPT on a \emph{graph} as follow:

i) Note that $\lambda \mapsto  \mathrm{ET}_{c,\lambda}(\mu,\nu)$ is a concave function  since it is the infimum of a family of concave functions in $\lambda$. Therefore,  $ H$ is   convex on $\R$. In particular, $H$ is differentiable almost everywhere on $\R$. 

Let $\lambda\in\R$, recall the definition of $\mathcal{C}_{\lambda } (\gamma)$ in Equation~\eqref{easier-form}. Then for any $\gamma\in \Gamma^0(\lambda)$,  we have 
\begin{align}\label{sub-ineq}
  \mathrm{ET}_{c,\lambda +\delta}(\mu,\nu) 
  \leq \mathcal{C}_{\lambda +\delta} (\gamma)=\mathcal{C}_{\lambda } (\gamma)- b \delta \gamma(\G\times\G)
  = \mathrm{ET}_{c,\lambda }(\mu,\nu)- b \delta \gamma(\G\times\G) \,\,\,  \forall \delta\in\R.
 \end{align}
This implies that 
\[
\big\{ b\, \gamma(\G\times\G): \gamma\in \Gamma^0(\lambda)\big\} \subset \partial H(\lambda).
\]
We next show that the opposite  inclusion is also true, i.e., $\big\{ b\, \gamma(\G\times\G): \gamma\in \Gamma^0(\lambda)\big\} = \partial H(\lambda)$. This is obviously holds if $\partial H(\lambda) $ is singleton, which holds for example when $H$ is differentiable at $\lambda$. Hence  we only need to consider  $\lambda$ for which  the convex set $\partial H(\lambda)$ has more than one element. 

Let $m \in \partial H(\lambda)$, then $m$ can be expressed as a convex combination of extreme points $m_1, \dotsc, m_N$ of $\partial H(\lambda)$, i.e.,  $m= \sum_{i=1}^N t_i m_i$ with $0\leq t_i \leq 1$ and $\sum_{i=1}^N t_i=1$. As $m_i$ is an extreme point of $\partial H(\lambda)$, there exists a sequence $\lambda_n\to \lambda$ such that $\lambda_n$ is a differentiable point of $H$ and $H'(\lambda_n)\to m_i$. 

Let $\gamma^n \in \Gamma^0(\lambda_n)$, then $b\, \gamma^n(\G\times \G)=H'(\lambda_n)\to m_i$. By compactness, there exists a subsequence $\{\gamma^{n_k}\}$ and $\tilde\gamma^i\in \Pi_{\leq}(\mu,\nu)$ such that  $\gamma^{n_k}\to \tilde \gamma^i$ weakly. It follows that  $\gamma^{n_k}(\G\times \G)\to \tilde \gamma^i(\G\times \G)$, and hence we must have $b\, \tilde\gamma^i(\G\times \G)=m_i$. We have
\begin{align*}
   \mathcal{C}_{\lambda_{n_k}} (\gamma^{\lambda_{n_k}})
  = \mathcal{C}_{\lambda } (\gamma^{\lambda_{n_k}})
 + b (\lambda -\lambda_{n_k} )  \gamma^{n_k}(\G\times \G) 
  &\geq \mathrm{ET}_{c,\lambda }(\mu,\nu)+ b (\lambda -\lambda_{n_k} )  \gamma^{n_k}(\G\times \G)\\
  &\geq \mathrm{ET}_{c,\lambda }(\mu,\nu)- b \bar m |\lambda -\lambda_{n_k}|
 \end{align*}
and for any $\gamma \in \Gamma^0(\lambda)$, there holds
\begin{align*}
   \mathcal{C}_{\lambda_{n_k}} (\gamma^{\lambda_{n_k}})\leq \mathcal{C}_{\lambda_{n_k}} (\gamma)
   =\mathcal{C}_{\lambda } (\gamma)+ b (\lambda -\lambda_{n_k} )  \gamma(\G\times \G) 
  = \mathrm{ET}_{c,\lambda }(\mu,\nu)+b (\lambda -\lambda_{n_k} )  \gamma(\G\times \G).
 \end{align*}
 We thus deduce that $\lim_{k\to\infty} \mathcal{C}_{\lambda_{n_k}} (\gamma^{\lambda_{n_k}})=\mathrm{ET}_{c,\lambda }(\mu,\nu)$. These together with the lower semicontinuity of $\mathcal{C}_{\lambda }$ give
\begin{align*}
\mathrm{ET}_{c,\lambda }(\mu,\nu) =\liminf_{k\to\infty} \mathcal{C}_{\lambda_{n_k}} (\gamma^{\lambda_{n_k}})
&=\liminf_{k\to\infty}\Big[ \mathcal{C}_{\lambda } (\gamma^{\lambda_{n_k}})
 + b (\lambda -\lambda_{n_k} )  \gamma^{n_k}(\G\times \G)\Big]\\
 &=\liminf_{k\to\infty}\mathcal{C}_{\lambda } (\gamma^{\lambda_{n_k}})\geq \mathcal{C}_{\lambda } (\tilde \gamma^i).
\end{align*}
Therefore,  $\tilde \gamma^i \in \Gamma^0(\lambda)$ with mass $b\, \tilde \gamma^i(\G\times \G)=m_i$. Due to the convexity of $\Gamma^0(\lambda)$, we have $\bar\gamma :=\sum_{i=1}^N t_i \tilde \gamma^i \in \Gamma^0(\lambda)$ with $b\, \bar  \gamma(\G\times \G)=\sum_{i=1}^N t_i m_i=m$. That is, 
\[
 \partial H(\lambda) \subset \big\{ b\, \gamma(\G\times\G): \gamma\in \Gamma^0(\lambda)\big\},
 \]
and we thus infer that $\big\{ b\, \gamma(\G\times\G): \gamma\in \Gamma^0(\lambda)\big\} = \partial H(\lambda)$ for all $\lambda\in\R$.

In order to prove the second part of i), let $ \gamma\in \Gamma^0(\lambda_1)$ and $ \tilde \gamma\in \Gamma^0(\lambda_2)$ be arbitrary.  We have 
 \begin{align}\label{another-sub-ineq}
  \mathrm{ET}_{c,\lambda_2}(\mu,\nu) 
  = \mathcal{C}_{\lambda_2} (\tilde\gamma)
 & = \mathcal{C}_{\lambda_1 } (\tilde \gamma)
  - b (\lambda_2 -\lambda_1) \tilde \gamma(\G\times\G)\nonumber\\
  &\geq \mathrm{ET}_{c,\lambda_1 }(\mu,\nu)-b (\lambda_2 -\lambda_1) \tilde \gamma(\G\times\G).
 \end{align}
Hence by combining with  \eqref{sub-ineq}, we deduce that
\begin{align*}
 \mathrm{ET}_{c,\lambda_1 }(\mu,\nu)- b (\lambda_2-\lambda_1) \tilde \gamma(\G\times\G) \leq  \mathrm{ET}_{c,\lambda_2}(\mu,\nu) 
  \leq  \mathrm{ET}_{c,\lambda_1 }(\mu,\nu)- b (\lambda_2-\lambda_1) \gamma(\G\times\G),
 \end{align*}
which yields $\gamma(\G\times\G)\leq \tilde \gamma(\G\times\G)$. This together with the above characterization of $\partial H(\lambda)$ implies the second part of i).

ii) If $H$ is differentiable at $\lambda$, then $\partial H(\lambda)$ is a singleton set. However, as $\partial H(\lambda) =\big\{ b\, \gamma(\G\times\G): \gamma\in \Gamma^0(\lambda)\big\}$  by i), we thus infer that the mass $\gamma(\G\times\G)$ must be the same for every $\gamma\in \Gamma^0(\lambda)$.

Next assume that every element  in $\Gamma^0(\lambda)$ has the same  mass, say $m$. For $\delta\neq 0$,
 let  $\gamma^{\lambda +\delta}\in \Gamma^0(\lambda+\delta)$ and  $m(\lambda+\delta)\Let \gamma^{\lambda +\delta}(\G\times\G)$.
 Then, we claim that 
 \begin{equation}\label{m-cont}
 \lim_{\delta\to 0} m(\lambda +\delta) =m.
 \end{equation}
 Assume the claim for the moment, and let $\delta> 0$. Then, as in \eqref{sub-ineq}--\eqref{another-sub-ineq}, we have
 \begin{align*}
  \mathrm{ET}_{c,\lambda +\delta}(\mu,\nu) 
  \leq \mathrm{ET}_{c,\lambda }(\mu,\nu)- b \delta m\quad
  \mbox{and}\quad 
  \mathrm{ET}_{c,\lambda +\delta}(\mu,\nu) 
  \geq \mathrm{ET}_{c,\lambda }(\mu,\nu)- b \delta m(\lambda+\delta).
 \end{align*}
  It follows that
 \[
 - b  m(\lambda+\delta) \leq 
 \frac{\mathrm{ET}_{c,\lambda +\delta}(\mu,\nu)-\mathrm{ET}_{c,\lambda }(\mu,\nu)}{\delta}
 \leq - b  m.
 \]
This together with  claim \eqref{m-cont} gives $\lim_{\delta\to 0^+} \frac{\mathrm{ET}_{c,\lambda +\delta}(\mu,\nu)-\mathrm{ET}_{c,\lambda }(\mu,\nu)}{\delta} =- b  m$. By the same argument, we also have 
 $\lim_{\delta\to 0^-} \frac{\mathrm{ET}_{c,\lambda +\delta}(\mu,\nu)-\mathrm{ET}_{c,\lambda }(\mu,\nu)}{\delta} =- b  m$.
 Thus, we infer that $H$ is differentiable at $\lambda$ with  
 $H'(\lambda)= b m$.
Therefore, it remains to prove claim \eqref{m-cont}. 

Indeed, by compactness there exists a subsequence, still labeled by $\gamma^{\lambda+\delta}$, and $\gamma \in \Pi_{\leq}(\mu,\nu)$ such that 
$\gamma^{\lambda+\delta}\to \gamma$ weakly as $\delta\to 0$. As in i), we can show  that $\gamma\in \Gamma^0(\lambda)$.
Then, as the mass functional  is weakly continuous, we obtain $m(\lambda+\delta)=\gamma^{\lambda+\delta}(\G\times \G)\to \gamma(\G\times \G)=m$. We in fact have shown that any subsequence of $\{m(\lambda+\delta)\}_\delta$ has a further subsequence converging to the same number $m$. Therefore, the full sequence $\{m(\lambda+\delta)\}_\delta$ must converge to $m$, and hence \eqref{m-cont} is proved.

iii) For any $\lambda\in\R$, we have by i) that $ \partial H(\lambda) = \big\{ b\, \gamma(\G\times\G): \gamma\in \Gamma^0(\lambda)\big\}\subset [0,b\, \bar m]$. Thus, we only need to prove $[0,b\, \bar m]\subset \partial H(\R)$. First, note that as $ \partial H(\lambda)\subset\R$ is a compact and convex set, it must be a finite and closed interval. Therefore, if we let 
\[
\gamma^\lambda_{min} :=\argmin_{\gamma\in \Gamma^0(\lambda)} \gamma(\G\times\G)\quad \mbox{and}\quad 
\gamma^\lambda_{max} :=\argmax_{\gamma\in \Gamma^0(\lambda)} \gamma(\G\times\G),
\]
then it follows from ii) that $ \partial H(\lambda)=\big[b\,\gamma^\lambda_{min}(\G\times\G), b\, \gamma^\lambda_{max}(\G\times\G)\big] $ for every $\lambda\in\R$.
From Equation~\eqref{easier-form}, it is clear that $ \partial H(\lambda)=\{0\}$ for $\lambda$ negative enough. Indeed, if we take $\lambda < -M$, then as $w_1(x) +w_2(y) \leq b\,  [c(x,y) +M]$, we have 
$0<  b \, [c(x,y)-\lambda]- w_1(x) -w_2(y)$
for all $x,y\in \G$. Then, we obtain from 
 Equation~\eqref{easier-form}  that $\mathcal{C}_\lambda(  0)\leq \mathcal{C}_\lambda(  \gamma)$ for every  $\gamma\in \Pi_{\leq}(\mu,\nu)$ and the strict inequality holds if $\gamma \neq 0$. Thus, $ \Gamma^0(\lambda)=\{0\}$ which gives  $\partial H(\lambda)=\{0\}$ and $H(\lambda)=-\int_\G  w_1 \mu(\dd x) 
- \int_\G  w_2  \nu(\dd x)$. 

We next show that $ \partial H(\lambda)=\{b\, \bar m\}$ for $\lambda$ positive enough.
Since $c(x,y)$ is bounded due to its continuity on $\G\times\G$, we can choose $\lambda\in\R$ such that $c(x,y)-\lambda<0$ for all $x,y\in \G$. Let $\gamma \in \Gamma^0(\lambda)$. We claim that  either $\gamma_1=\mu$ or $\gamma_2=\nu$. Indeed, since otherwise we have  $\gamma_1(A_0)<\mu(A_0)$ and  $\gamma_2(B_0)<\nu(B_0)$ for some Borel sets $A_0, B_0\subset \G$. Let $\tilde \gamma := \gamma +  [(\mu-\gamma_1)\chi_{A_0}]\otimes [(\nu-\gamma_2)\chi_{B_0}]$. Then, for any Borel set $A\subset \G$ we have
\begin{align*}
 \tilde\gamma_1(A) = \gamma_1(A) + \mu(A\cap A_0) - \gamma_1(A\cap A_0) & =\gamma_1(A\setminus  A_0) +\mu(A\cap A_0)\\
 &\leq     \mu(A\setminus  A_0) +\mu(A\cap A_0)=\mu(A).
\end{align*}
Likewise, $\tilde\gamma_2(B)\leq \nu(B)$ for any Borel set $B\subset\G$. Thus 
$\tilde \gamma \in \Pi_{\leq}(\mu,\nu)$. On the other hand, it is clear from \eqref{easier-form} 
 and the facts $\gamma_1\leq \tilde\gamma_1$,  $\gamma_2\leq \tilde\gamma_2$, and  $c-\lambda <0$ that $\mathcal{C}_\lambda(  \tilde \gamma)< \mathcal{C}_\lambda(  \gamma)$. This is impossible and so the claim is proved. That is, either $\gamma_1=\mu$ or $\gamma_2=\nu$. It follows that $\gamma(\G\times \G) = \bar m$ for every $\gamma\in \Gamma^0(\lambda)$, and hence $\partial H(\lambda)=\{b\, \bar m\}$ due to i). This also means that $H$ is differentiable at $\lambda$ with  $H'(\lambda)=b\, \bar m$. 

Therefore, it remains to show that 
\begin{equation}\label{sub-inclusion}
(0,b\, \bar m)\subset \partial H(\R)= \bigcup_{\lambda\in\R}\big[b\, \gamma^\lambda_{min}(\G\times\G), b\, \gamma^\lambda_{max}(\G\times\G)\big].
\end{equation}
Assume by contradiction that there exists $m\in (0,b\, \bar m)$ such that $m\not\in \partial H(\lambda)$ for every $\lambda \in \R$. For convenience, we adopt the following notation: for sets $A, B\subset \R$ and $r\in\R$, we write $A< r$ if $a<r$ for every $a\in A$, and $A<B$ if $a<b$ for every $a\in A$ and $b\in B$. Let us consider the following two sets
\[
S_1 := \{\lambda: \partial H(\lambda) <m\}\quad \mbox{and}\quad S_2 := \{\lambda: \partial H(\lambda) >m\}.
\]
Then $\lambda\in S_1$ if $\lambda$ is negative enough, and  $\lambda\in S_2$ if $\lambda$ is positive enough. For any $\lambda_1\in S_1$ and $\lambda_2\in S_2$, we have $\partial H(\lambda_1) <m<\partial H(\lambda_2)$, and hence $\lambda_1 < \lambda_2$  by the monotonicity in i). That is,   $S_1 <S_2$ and so we obtain 
\[
\lambda^*:=\sup\{\lambda: \lambda\in S_1\}\leq \inf\{\lambda: \lambda\in S_2\}=:\lambda^{**}.
\]
If $\lambda^* <\lambda^{**}$, then for any $\lambda \in (\lambda^*, \lambda^{**})$ we have $\lambda \not\in S_1$ and $\lambda \not\in S_2$. Therefore, $\partial H(\lambda) \not< m$ and $\partial H(\lambda) \not> m$. Hence, we can find $m_1, m_2\in \partial H(\lambda) $ such that $m_1\geq m$ and $m_2 \leq m$. Thus, $m\in [m_2, m_1]\subset \partial H(\lambda)$ due to the convexity of the set $\partial H(\lambda)$. This contradicts our hypothesis, and we conclude that $\lambda^* =\lambda^{**}$. 

We next select sequences $\{\lambda^1_n\}\subset S_1$ and $\{\lambda^2_n\}\subset S_2$ such that  $\lambda^1_n\to \lambda^*$
and $\lambda^2_n\to \lambda^{**}=\lambda^*$. For each $n$, let 
\[
\gamma^n_{min} :=\argmin_{\gamma\in \Gamma^0(\lambda^1_n)} \gamma(\G\times\G)\quad \mbox{and}\quad 
\gamma^n_{max} :=\argmax_{\gamma\in \Gamma^0(\lambda^2_n)} \gamma(\G\times\G).
\]
 By compactness, there exist subsequences, still labeled as $\{\gamma^n_{min}\}$ and $\{\gamma^n_{max}\}$, and $\gamma^*, \gamma^{**}\in \Pi_{\leq}(\mu,\nu)$ such that  $\gamma^n_{min} \to \gamma^*$ weakly and $\gamma^n_{max} \to \gamma^{**}$ weakly. By arguing exactly as in i), we then obtain $\gamma^*, \gamma^{**}\in \Gamma^0(\lambda^*)$, $\gamma^n_{min}(\G\times \G) \to  \gamma^*(\G\times \G)$, and $\gamma^n_{max}(\G\times \G) \to  \gamma^{**}(\G\times \G)$. As $b\, \gamma^n_{min}(\G\times \G)<m$ due to $\lambda^1_n\in S_1$, we must have $b\, \gamma^*(\G\times \G)\leq m$. Likewise, we have $b\, \gamma^{**}(\G\times \G)\geq m$ as $b\, \gamma^n_{max}(\G\times \G)>m$ for all $n$.
 Hence, $m\in [b\, \gamma^*(\G\times \G), b\, \gamma^{**}(\G\times \G)]$. Since $\gamma^*, \gamma^{**}\in \Gamma^0(\lambda^*)$, we infer that $m\in \partial H(\lambda^*)$. This is a contradiction and the proof is complete. We note that since $\lambda^1_n\leq \lambda^*\leq \lambda^2_n$, we have from the monotonicity in i) that 
 \[
 \gamma^n_{min}(\G\times \G) \leq \gamma(\G\times \G)\leq \gamma^n_{max}(\G\times \G)
 \]
 for every $\gamma\in \Gamma^0(\lambda^*)$. By sending $n$ to infinity,  it follows that $\gamma^*(\G\times \G) \leq \gamma(\G\times \G)\leq \gamma^{**}(\G\times \G)$ for every $\gamma\in \Gamma^0(\lambda^*)$. That is,
 $
\gamma^* =\gamma^{\lambda^*}_{min}$ and 
$\gamma^{**} =\gamma^{\lambda^*}_{max}
$.
\end{proof}

\subsubsection{Proof of Lemma~\ref{lm:lipschitz-vs-sobolev}}\label{proof:1infinity-via-lipschitz}


\begin{proof}[Proof of Lemma~\ref{lm:lipschitz-vs-sobolev}]
Let us define 
\begin{align*}
  A &\Let\Big\{f\in C(\G):\, 
   |f(x)-f(y) |\leq b \, d_\G(x,y)\Big\}.
\end{align*}
and
\begin{align*}
B &\Let \Big\{f\in W^{1,\infty}(\G, \omega^*):\, 
   \|f'\|_{L^{\infty}(\G, \omega^*)}\leq b\Big\}
 \end{align*}

i) The statement of this part is equivalent to showing that  $A\subset  B$. Let $f\in A$. Then $f$ is continuous on $\G$, and 
\begin{equation}\label{L-d}
 |f(x)-f(y) |\leq b \, d_\G(x,y) \quad \forall x,y \in \G.  
\end{equation}
On each edge $e$ and similar to the real line, the Lipschitz condition \eqref{L-d} implies that there exists a function $h_e: e \to \R$
with the following properties: $|h_e(z)| \leq b$ for $\omega^*$-a.e. $z\in e$, and 
\begin{equation*}
 f(x) = f(y) + \int_{\langle y, x\rangle} h_e(z) \, \omega^*(\dd z) \quad \forall x,y \in e,
\end{equation*}
where we recall that $\langle y, x\rangle$ denotes the line segment in $\R^n$ connecting $y$ and  $x$ (noting that for general graph, $\langle y, x\rangle$ might not be the same as the shortest path  $[y,x]$).
Let us  glue  them together by  taking $h(z) = h_e(z)$ if $z$ is an interior point of an edge $e$. Then $h:\G\to \R$ is a function satisfying:  $|h(z)| \leq b$ for $\omega^*$-a.e. $z\in G$. That is, $h\in L^\infty(\G, \omega^*)$ with $\|h\|_{L^\infty(\G, \omega^*)}\leq b$.
Moreover, for every edge $e$ in $\G$ we have
\begin{equation}\label{on-each-edge}
 f(x) = f(y) + \int_{\langle y, x\rangle } h(z) \, \omega^*(\dd z) \quad \forall x,y \in e.
\end{equation}
Now let $x\in \G$ be arbitrary. Let us break the unique shortest path $[z_0, x]$ connecting $z_0$ and $x$ into sub line segments $\langle z_0, y_0\rangle, \, \langle y_0, y_1\rangle,...,  \langle y_{m-1}, y_m\rangle,\, \langle y_m, x\rangle$ such that each of  them is contained in exactly one edge. Then by applying \eqref{on-each-edge} to each of these sub line segments, we obtain
\begin{align*}
   f(x) -  f(z_0) 
   &= [f(x) - f(y_m)] +  [f(y_m) - f(y_{m-1})] + \cdots 
    +[f(y_0) - f(z_0)]\\
    &= \int_{\langle y_m, x\rangle} h(z) \, \omega^*(\dd z) + \int_{\langle y_{m-1}, y_m\rangle} h(z) \, \omega^*(\dd z)
    +\cdots 
    + \int_{\langle z_0, y_0\rangle} h(z) \, \omega^*(\dd z)\\
 &= \int_{[z_0, x]} h(z) \, \omega^*(\dd z).  \end{align*}
Thus, we have proved that
\begin{equation*}
f(x) = f(z_0) + \int_{[z_0, x]} h(z) \, \omega^*(\dd z) \quad \forall x\in \G.
\end{equation*}
Therefore, according to Definition~\ref{def:Sobolev} we conclude that $f\in W^{1,\infty}(\G, \omega^*)$ with $\|f'\|_{L^{p'}(\G, \omega^*)}\leq b$. It then follows that  $f\in B$, and hence $A \subset \mathbb B$ as desired.

ii) Assume that $\G$ is a tree. We can and will assume that $z_0$ is the root of this tree. We need to show that $B\subset A$. For this, let $f\in B$. Then by Definition~\ref{def:Sobolev}, we have $\|f'\|_{L^\infty(\G, \omega^*)}\leq b$ and 
\[
f(x) = f(z_0) + \int_{[z_0, x]} f'(z) \, \omega^*(\dd z) \quad \forall x\in \G.
\]
Thus for any two points $x,y \in\G$, we obtain 
\begin{equation}\label{difference}
  |f(x) - f(y)| =  \left| \int_{[z_0, x]} f'(z) \, \omega^*(\dd z) - \int_{[z_0, y]} f'(z) \, \omega^*(\dd z)\right|. 
\end{equation}
Let $\hat z$ be the deepest node on the tree that belongs to both path $[z_0, x]$ and path $[z_0, y]$.  Due to the tree structure,  the joining of path 
$[x, \hat z]$ and  path $[\hat  z, y]$ constitutes  the shortest path $[x,y]$ connecting the points $x$ and $y$. These together with \eqref{difference} imply that
\begin{align*}
  |f(x) - f(y)| 
  &=  \left| \int_{[\hat z, x]} f'(z) \, \omega^*(\dd z) - \int_{[\hat z, y]} f'(z) \, \omega^*(\dd z)\right|\\
  &\leq   \int_{[x, \hat z]} |f'(z)| \, \omega^*(\dd z) + \int_{[\hat z, y]} |f'(z)| \, \omega^*(\dd z)\\
  &= \int_{[x, y]} |f'(z)| \, \omega^*(\dd z)
  \leq \|f'\|_{L^\infty(\G, \omega^*)} \omega^*([x,y])\leq b \, \omega^*([x,y]).
\end{align*}
By the property of the length measure given in
Lemma~\ref{lem:length-measure}, we then infer that $|f(x) - f(y)|\leq b\, d_\G(x,y) $ for every $x,y\in\G$. It follows that $f\in A$. Therefore, we have proved that $B \subset A$ as desired. 
\end{proof}

\subsubsection{Proof of Theorem~\ref{thm:duality} }\label{app:dual-formula}

The proof of Theorem~\ref{thm:duality} is based on two auxiliary lemmas. Before stating these lemmas, let us describe the the  setting and associated problem. 

First, in order to investigate  problem \eqref{P1},  we recast it as the standard complete OT problem by using an observation in \citep{CM}. More precisely, let $\hat s$ be a point outside graph $\G$ and consider the set  $\hat\G:= \G \cup \{\hat s\}$. We next  extend the cost function to $\hat \G\times \hat \G$ as follow 
\begin{equation*}
\hat c(x,y) \Let
\left\{\begin{array}{lr}
\!\!b[c(x,y)-\lambda] \hspace{1 em} \mbox{ if } x,y\in \G,\\
\!\!w_1(x) \hspace{4 em} \mbox{ if }  x\in \G \mbox{ and } y=\hat s,\\
 \!\! w_2(y) \hspace{4 em}  \mbox{ if }  x=\hat s \mbox{ and } y\in \G,\\
  \!\! 0 \hspace{6 em} \mbox{ if }  x=y=\hat s.
\end{array}\right.
\end{equation*}
The measures $\mu, \nu$ are extended accordingly by adding a Dirac mass at the isolated point $\hat s$: $\hat\mu = \mu +\nu(\G) \delta_{\hat s}$ and $\hat\nu = \nu +\mu(\G) \delta_{\hat s}$. As $\hat\mu, \hat\nu$ have the same total mass on $\hat \G$, we can consider the standard complete OT problem between $\hat\mu, \hat\nu$ as follow
\begin{align}\label{P2}
\mathrm{KT}(\hat \mu,\hat \nu) \Let \inf_{\hat \gamma \in \Gamma(\hat\mu,\hat \nu)}  \int_{\hat \G\times \hat \G} \hat c(x,y) \hat\gamma(\dd x, \dd y),
\end{align}
where 
\[\Gamma(\hat \mu,\hat \nu) \Let \Big\{ \hat\gamma \in \calM( \hat\G \times \hat \G): \hat \mu(U) =\hat\gamma(U\times \hat \G),\, \hat\nu(U)= \hat\gamma(\hat \G\times U) \mbox{ for all Borel sets } U\subset \hat \G\Big\}.
\]

This reformulation under an observation in \citep{CM} helps us to transform an unbalanced optimal transport (EPT) on a graph into a corresponding standard complete OT. Therefore, we can not only bypass all the issues coming from the unbalanced setting, but also rely on many results in the standard setting for OT. 

We then adapt the procedure in~\citep{CM} to derive the dual formulation for the EPT on a \emph{graph}.

Additionally, we have a one-to-one correspondence between $\gamma\in \Pi_{\leq}(\mu,\nu)$ and $\hat \gamma\in \Gamma(\hat \mu,\hat \nu)$ as follow 
\begin{align}\label{one-to-one}
\hat \gamma = \gamma + [(1-f_1) \mu] \otimes \delta_{\hat s} +\delta_{\hat s} \otimes [(1-f_2) \nu] 
+\gamma(\G\times \G) \delta_{(\hat s, \hat s)}.
\end{align}
Indeed, if $\gamma\in \Pi_{\leq}(\mu,\nu)$, then it is clear that $\hat\gamma$ defined by \eqref{one-to-one} satisfies $\hat \gamma\in \Gamma(\hat \mu,\hat \nu)$. The converse is guaranteed by the next technical result.
\begin{lemma}\label{rep-formula}
For $\hat \gamma\in \Gamma(\hat \mu,\hat \nu)$, let $\gamma$  be the restriction of  $\hat\gamma$ to $\G$. Then, relation \eqref{one-to-one} holds  and $\gamma\in \Pi_{\leq}(\mu,\nu)$.
 \end{lemma}
\begin{proof}
We first observe for any Borel set $A\subset \G$ that
\begin{align*}
\hat \gamma(A\times \{\hat s\}) = \hat \gamma(A\times\hat \G) - \hat \gamma(A\times \G)=\hat \mu(A) - \gamma(A\times \G)=\mu(A) -\gamma_1(A)=\int_A (1-f_1) \mu(\dd x).
\end{align*}
For the same reason, we have  $\hat \gamma( \{\hat s\}\times B)=  \int_B (1-f_2) \nu(dx)$ for any set
 Borel set $B\subset \G$. Also,
 \begin{align*}
\hat \gamma(\{\hat s\}\times \{\hat s\}) 
&= \hat \gamma(\hat\G \times \{\hat s\}) -\hat \gamma(\G\times \{\hat s\})  \\
&= \hat \gamma(\hat\G\times \hat\G)
- \hat \gamma(\hat\G\times \G)   -\big[  \hat\gamma(\G\times \hat\G) -  \hat \gamma(\G\times \G) \big]\\
&= \hat\mu(\hat\G) - \hat\nu(\G) - \hat\mu(\G) +\gamma(\G\times \G)
=\gamma(\G\times \G).
\end{align*}
 
Since \eqref{one-to-one} 
 is obviously true for sets of the form $A\times B$ with $A,B\subset \G$ being Borel sets, we only need to verify it for sets of the following three forms:  $(A\cup 
 \{\hat s\})\times B$, $A\times (B\cup \{\hat s\})$, $(A\cup \{\hat s\})\times (B\cup\{\hat s\})$ for Borel sets  $A,B\subset \G$. We check it case by case as follows.
 
$\bullet$ (i) For $(A\cup 
 \{\hat s\})\times B$: Using the above observation, we have 
 \begin{align*}
\hat \gamma((A\cup 
 \{\hat s\})\times B) 
&=  \hat \gamma(A\times B) + \hat \gamma(\{\hat s\}\times B) =  \gamma(A\times B) + \int_B (1-f_2) \nu(\dd x).
\end{align*}
 Therefore, \eqref{one-to-one} 
  holds in this case.
 
$\bullet$ (ii) For $A \times (B\cup 
 \{\hat s\}))$:  \eqref{one-to-one} is also true for this case because 
 \begin{align*}
\hat \gamma(A \times (B\cup 
 \{\hat s\})) 
&=  \hat \gamma(A\times B) + \hat \gamma(A\times \{\hat s\}) 
=  \gamma(A\times B) + \int_A (1-f_1) \mu(\dd x).
\end{align*}

$\bullet$ (iii) For $(A\cup 
 \{\hat s\}) \times (B\cup 
 \{\hat s\})$:   \eqref{one-to-one} is true as well since 
 \begin{align*}
\hat \gamma((A\cup 
 \{\hat s\}) \times (B\cup 
 \{\hat s\})) 
&=  \hat \gamma(A\times B) + \hat \gamma(A\times \{\hat s\})  + \hat \gamma(\{\hat s\}\times B) +\hat \gamma(\{\hat s\}\times \{\hat s\})\\
&=  \gamma(A\times B)  + \int_A (1-f_1) \mu(\dd x)
+\int_B (1-f_2) \nu(\dd x)
+\gamma(\G\times \G).
\end{align*}

Now as \eqref{one-to-one}  
 holds, we obviously have $\gamma(U\times \G)\leq \hat \gamma(U\times \G)\leq \hat \gamma(U\times \hat \G)=\hat\mu(U)=\mu(U)$ for any Borel set $U\subset \G$. Likewise, $\gamma( \G\times U )\leq \nu(U)$ for any Borel set $U\subset \G$. Therefore, $\gamma\in \Pi_{\leq}(\mu,\nu)$.
\end{proof}

These observations in particular display the following connection between the EPT problem on a graph \eqref{P1} and the corresponding standard complete OT problem \eqref{P2}.
\begin{lemma}[EPT on a graph versus its corresponding complete OT]\label{distance-agree}
For every $\mu, \nu \in \calM(\calT )$, we have
 $\mathrm{ET}_{c,\lambda}(\mu,\nu)=\mathrm{KT}(\hat \mu,\hat\nu)$. Moreover, relation \eqref{one-to-one} gives a one-to-one correspondence between  optimal solution  $\gamma$ for EPT problem \eqref{P1} and optimal solution $\hat \gamma$ for standard complete OT problem \eqref{P2}.
\end{lemma}
\begin{proof}
We derive two parts as follow:

$\bullet$ (i) We show that $\mathrm{KT}(\hat \mu,\hat\nu)  \leq\mathrm{ET}_{c,\lambda}(\mu,\nu) $:  

For any  $\gamma\in \Pi_{\leq}(\mu,\nu)$, let $\hat\gamma$ be given by \eqref{one-to-one}. Then, $\hat \gamma\in \Gamma(\hat \mu,\hat \nu)$ and 
\begin{align*}
\mathrm{KT}(\hat \mu,\hat\nu) \leq 
\int_{\hat \G\times \hat \G} \hat c(x,y) \hat\gamma(\dd x, \dd y)
&=b \int_{ \G\times  \G} [c(x,y)-\lambda] \gamma(\dd x, \dd y)\\
&\quad +\int_\G  w_1 [1-f_1(x)] \mu(\dd x)  
+ \int_\G  w_2 [1-f_2(x)] \nu(\dd x).
\end{align*}
It follows that $\mathrm{KT}(\hat \mu,\hat\nu)  \leq\mathrm{ET}_{c,\lambda}(\mu,\nu)$.
 
$\bullet$ (ii) We show that $\mathrm{KT}(\hat \mu,\hat\nu)\geq \mathrm{ET}_{c,\lambda}(\mu,\nu)$:  

To see this, for any  $\hat \gamma\in \Gamma(\hat \mu,\hat \nu)$ we let 
$\gamma$ be the restriction of  $\hat\gamma$ to $\calT$. Then by Lemma~\ref{rep-formula}, we have $\gamma\in \Pi_{\leq}(\mu,\nu)$ and \eqref{one-to-one}  holds. Consequently,
\begin{align*}
\int_{\hat \G\times \hat \G} \hat c(x,y) \hat\gamma(\dd x, \dd y)
&=b \int_{ \G\times  \G} [c(x,y)-\lambda] \gamma(\dd x, \dd y)\\
&\quad +\int_\G  w_1 [1-f_1(x)] \mu(\dd x)  
+ \int_\G  w_2 [1-f_2(x)] \nu(\dd x)\\
&\geq  \mathrm{ET}_{c,\lambda}(\mu,\nu).
\end{align*}
By taking the infimum over $\hat \gamma$, we infer that $\mathrm{KT}(\hat \mu,\hat\nu)  \geq\mathrm{ET}_{c,\lambda}(\mu,\nu)$. 

Thus, from the above two parts, we obtain 
\[
\mathrm{KT}(\hat \mu,\hat\nu)  = \mathrm{ET}_{c,\lambda}(\mu,\nu). 
\]
The relation about the optimal solutions also follows from the above arguments.
\end{proof}

Given the above two lemmas, we are ready to present the proof of Theorem~\ref{thm:duality}.
\begin{proof}[Proof of Theorem~\ref{thm:duality} ]
From Lemma~\ref{distance-agree} 
 and the dual formulation for $\mathrm{KT}(\hat \mu,\hat\nu)$ proved in \cite[Corollary~2.6] {CM}, we have
\begin{align*}
\mathrm{ET}_{c,\lambda}(\mu,\nu)=\sup_{\substack{\hat u \in L^1(\hat \mu),\, \hat v\in L^1(\hat \nu)\\\hat u(x) +\hat v(y)\leq \hat c(x,y)}} \int_{\hat\G} \hat u(x) \hat \mu(\dd x) +  \int_{\hat\G} \hat v(x) \hat \nu(\dd x)=: I.
\end{align*}
Therefore, it is enough to prove that  $I=J$ where
\begin{equation*}
J \Let
\sup_{(u,v) \in \K} \Big[ \int_{\G}  u(x) \mu(\dd x) +  \int_{\G}  v(x)  \nu(\dd x)\Big].
\end{equation*}

For  $(u,v)$ satisfying $u\leq w_1$, $v\leq w_2$ and $ u(x) + v(y)\leq  b[c(x,y) - \lambda]$, we extend  it to $\hat \G$ by taking $\hat u(\hat s)=0$ and $\hat v(\hat s)=0$. Then, it is clear that $\hat u(x) +\hat v(y)\leq \hat c(x,y)$ for $x,y\in \hat\G$, and 
\begin{align*}
I\geq \int_{\hat\G} \hat u(x) \hat \mu(\dd x) +  \int_{\hat\G} \hat v(x) \hat \nu(\dd x) 
=\int_{\G}  u(x) \mu(\dd x) +  \int_{\G}  v(x)  \nu(\dd x).
\end{align*}
It follows that $I\geq J$. In order to prove the converse, let $(\hat u,\hat v)$ be a maximizer for $I$. Then, by considering $(\hat u - \hat u(\hat s),\hat v +\hat u(\hat s))$, we can assume  that $\hat u(\hat s)=0$. Also, if we let 
$
v(y) :=\inf_{x\in\hat\G} [\hat c(x,y)-\hat u(x)]$,
then $(\hat u,v)$ is still in the admissible class for $I$ and $\hat v(y)\leq v(y)$.  This implies that  $(\hat u,v)$ is also a maximizer for $I$. For these reasons,  we can assume w.l.g. that the maximizer $(\hat u,\hat v)$  has the following additional properties: $\hat u(\hat s)=0$ and 
\[
\hat v(y) =\inf_{x\in\hat\G} [\hat c(x,y)-\hat u(x)]\quad \forall y\in \hat\G.
\]
In particular, $\hat v(\hat s) = \inf_{x\in\hat\G} [\hat c(x,\hat s)-\hat u(x)]$. For convenience, define $w_1(\hat s)=0$ and consider the following two possibilities.

$\bullet$ (i) For $\inf_{x\in\hat \G} [w_1(x) - \hat u(x)]\geq 0$:

Since $\hat c(\hat s,\hat s)-\hat u(\hat s)=0$ and $ \inf_{x\in\G} [\hat c(x,\hat s)-\hat u(x)]=\inf_{x\in\G} [w_1(x) - \hat u(x)]\geq 0$, we have  $\hat v(\hat s) =0$. 

Also, $\hat v(y) \leq \hat c(\hat s,y)-\hat u(\hat s)\leq w_2(y)$ for all $ y\in \hat\G$.
    For each $ y\in\G$, by using  the facts $\hat  u \leq w_1$ and $\hat c(\hat s,y)-w_1(\hat s)=w_2(y)\geq 0$  we get 
\begin{align*}
\hat  v(y) \geq \inf_{x\in \hat \G} [\hat c(x,y)-w_1(x)]=\inf_{x\in  \G} \{b[c(x,y)-\lambda]-w_1(x)\}
= -b \lambda + \inf_{x\in \G} [b\, c(x,y) -w_1(x)].
\end{align*}
  Thus $(\hat  u, \hat  v)\in\K$ and 
\begin{align*}
I = \int_{\hat\G} \hat u(x) \hat \mu(\dd x) +  \int_{\hat\G} \hat v(x) \hat \nu(\dd x) 
&=\int_{\G}  \hat u(x) \hat \mu(\dd x) +  \int_{\G}  \hat v(x)  \hat \nu(\dd x)
+ \hat v(\hat s) \mu(\G) \\
&=  \int_{\G}  \hat u(x)  \mu(\dd x) +  \int_{\G}  \hat v(x)   \nu(\dd x)\leq  J.
\end{align*}

$\bullet$ (ii) For $\inf_{x\in\hat \G} [w_1(x) - \hat u(x)]< 0$: 

By arguing as in the above case (i),  we have  $\hat v(\hat s) =\inf_{x\in\G} [w_1(x) - \hat u(x)]<0$ and
\begin{align}\label{a-connection}
I=\int_{\G}  \hat v(x)  \nu(\dd x) + \int_{\G}  \hat u(x)  \mu(\dd x)   
+\mu(\G)  \inf_{\G} [w_1 - \hat u].
\end{align}
Let $\tilde u(x) := \min\{\hat u(x), w_1(x)\}$. Then, it is obvious that  $\tilde  u(x) +\hat v(y)\leq \hat c(x,y)$ and $\tilde  u(\hat s)=0$.
Since $\inf_{x\in\G} [w_1(x) - \hat u(x)]< 0$, there exists $x_0\in \G$ such that $w_1(x_0) < \hat u(x_0)$. Thus, $\tilde u(x_0) = w_1(x_0)$ and hence  $\inf_{\G} [w_1 -\tilde u] \leq 0$. As $\tilde u\leq w_1$, we infer further that $\inf_{\G} [w_1 -\tilde u] = 0$.
 We also have
\begin{align*}
& \int_{\G}  \hat u(x) \mu(\dd x)   
+\mu(\G)  \inf_{\G} [w_1 - \hat u]\\
&= \int_{\G}  \tilde  u(x)  \mu(\dd x)   
+ \int_{\G: \hat  u > w_1 }  [\hat u(x) -w_1(x)] \mu(\dd x)  
+ \mu(\G)  \inf_{\G} [w_1 - \hat u] \leq \int_{\G}  \tilde  u(x)  \mu(\dd x).
\end{align*}
This together with \eqref{a-connection} gives
\begin{align*}
I\leq \int_{\G}  \tilde  u(x)  \mu(\dd x)   +\int_{\G}  \hat v(x)   \nu(\dd x).
\end{align*}
Now let 
$\tilde  v(y) =\inf_{x\in \hat \G} [\hat c(x,y)-\tilde  u(x)]$ for $ y\in \G$. Then, $\hat v(y)\leq \tilde  v(y) \leq \hat c(\hat s,y)-\tilde  u(\hat s)=w_2(y)$ for $y\in \G$.  For each $ y\in\G$, by using  the facts $\tilde u \leq w_1$ and $\hat c(\hat s,y)-w_1(\hat s)=w_2(y)\geq 0$  we also get 
\begin{align*}
\tilde  v(y) \geq \inf_{x\in \hat \G} [\hat c(x,y)-w_1(x)]=\inf_{x\in  \G} \{b[c(x,y)-\lambda]-w_1(x)\}
= -b \lambda + \inf_{x\in \G} [b\, c(x,y) -w_1(x)].
\end{align*}
It follows that $(\tilde u, \tilde v)\in \K$ and 
\begin{align*}
I\leq \int_{\G}  \tilde  u(x)  \mu(\dd x)   +\int_{\G}  \tilde  v(x)  \nu(\dd x)\leq J.
\end{align*}

Thus we conclude that $I=J$ and the theorem follows.
\end{proof}

\subsubsection{Proof of Corollary~\ref{cor:duality}}

\begin{proof}[Proof of Corollary~\ref{cor:duality}]

Notice that as $w_i$ ($i=1,2$) is  $b$-Lipschitz w.r.t. $d_\G$, we have for every $x\in\G$ that 
\begin{equation}\label{w-lipschitz}
-w_i(x) \leq \inf_{y\in\G} \big[b\, d_\G(x,y)- w_i(y)\big].
\end{equation}
Let $\K$ be the set defined in the statement of Theorem~\ref{thm:duality}.
Then for each $(u,v)\in\K$, let
\begin{align*}
v^*(x) 
&:= \inf_{y\in\G} \big\{b[d_\G(x,y)-\lambda]- v(y)\big\} = -b\lambda + \inf_{y\in\G} \big[b\, d_\G(x,y)- v(y)\big]\geq u(x),\\
 v^{**}(y) 
 &:= \inf_{x\in\G} \big\{b[d_\G(x,y)-\lambda]- v^*(x)\big\}=-b\lambda + \inf_{x\in\G} \big[b\, d_\G(x,y) - v^*(x)\big]\geq v(y).
\end{align*}
By using  $ -b \lambda + \inf_{x\in \G} [b\, d_\G(x,y) -w_1(x)]\leq v(y)\leq w_2(y)$ and  \eqref{w-lipschitz}, we obtain  for every $x\in\G$ that
\begin{align*}
v^*(x)
&\leq -b\lambda -v(x)\leq -\inf_{y\in \G} [b\, d_\G(x,y) -w_1(y)]\leq w_1(x),\\
v^*(x)
&\geq -b\lambda + \inf_{y\in\G} \big[b\, d_\G(x,y)- w_2(y)\big]\geq -b\lambda -  w_2(x).
\end{align*}
 We also have  $v^*$ is  $b$-Lipschitz, i.e., $|v^*(x_1)-v^*(x_2) |\leq b \, d_\G(x_1,x_2)$. Indeed, let $x_1, x_2\in \G$. Then for any $\e>0$, there exists $y_1\in\G$ such that 
 \[
 b\, d_\G(x_1,y_1)- v(y_1) < v^*(x_1) +b\lambda +\e.
 \]
 It follows that
\[
v^*(x_2) -v^*(x_1) \leq b\, d_\G(x_2,y_1)- v(y_1)
+\e - [b\, d_\G(x_1,y_1)- v(y_1)]\leq b \, d_\G(x_1, x_2) +
\e.
\]
Since this holds for every  $\e>0$, we get  
\[
v^*(x_2) -v^*(x_1) \leq b \, d_\G(x_1, x_2).
\]
By interchanging the role of $x_1$ and $x_2$, we also obtain $v^*(x_1) -v^*(x_2) \leq b \, d_\G(x_1, x_2)$. Thus,  
\[
|v^*(x_1)-v^*(x_2) |\leq b \, d_\G(x_1,x_2).
\]
Hence, we have shown  that $v^*\in  \mathbb{U^*}$ with
\[
\mathbb{U^*} :=\Big\{f\in C(\G):\, 
 -b\lambda -w_2 \leq f \leq    w_1, \, |f(x)-f(y) |\leq b \, d_\G(x,y)\Big\}.
 \]

We next claim $v^{**}=  - b\lambda - v^*$. For this, it is clear from the definition that  $v^{**}(y) \leq - b\lambda - v^*(y)$. On the other hand, from the Lipschitz property of $v^*$ we obtain 
\[
- v^*(y) \leq b \, d_\G(x,y) - v^*(x) \quad \forall x\in \G,
\]
which gives  $- b\lambda - v^*(y) \leq v^{**}(y)$. Thus, we conclude that $v^{**}=  - b\lambda - v^*$ as claimed.

From these, we obtain that
\begin{align*}
\int_{\G}  u(x) \mu(\dd x) +  \int_{\G}  v(x)  \nu(\dd x)
&\leq \int_{\G}  v^*(x) \mu(\dd x) +  \int_{\G}  v^{**}(x)  \nu(\dd x)\\
&= \int_{\G}  v^*(x) \mu(\dd x) - \int_{\G}  v^{*}(x)  \nu(\dd x) -b\lambda \nu(\G)\\
&\leq - b\lambda \nu(\G) + \sup \left\{ \int_\G f (\mu - \nu) :\, f\in \mathbb{U^*}  \right\}.
\end{align*}
This together with Theorem~\ref{thm:duality}  in the main text implies that 
\[
\mathrm{ET}_\lambda(\mu,\nu)\leq  - b\lambda \nu(\G) + \sup \left\{ \int_\G f (\mu - \nu) :\, f\in \mathbb{U^*} \right\}.
\]
To prove the converse, let $ f\in \mathbb{U^*}$.
Define $u:= f$ and $ v:= -b\lambda -f$. Then,
we have 
\[
u(x) \leq w_1(x),
\]
\[
v(x)\leq -b\lambda-[-b\lambda -w_2(x)]=w_2(x),
\]
and 
\[
v(x)\geq -b\lambda- w_1(x)\geq -b\lambda + \inf_{y\in \G} [b\, d_\G(x,y) -w_1(y)].
\]
Also, the Lipschitz property of $f$ gives 
\[
u(x) + v(y) =-b\lambda + f(x)- f(y)\leq b[ d_\G(x,y)-\lambda]\quad \forall x,y\in\G.
\]
Thus  $(u,v) \in \K$, and hence we obtain from Theorem~\ref{thm:duality} in the main text that
 \begin{align*}
 - b\lambda \nu(\G) +  \int_\G f (\mu - \nu) 
 =
\int_{\G}  u(x) \mu(\dd x) +  \int_{\G}  v(x)  \nu(\dd x)
\leq \mathrm{ET}_\lambda(\mu,\nu).
\end{align*}
As this holds for every $f\in \mathbb{U^*}$, we get 
\[
- b\lambda \nu(\G) + \sup \left\{ \int_\G f (\mu - \nu) :\, f\in \mathbb{U^*} \right\}
\leq \mathrm{ET}_\lambda(\mu,\nu).
\]
Thus, we have shown that
\begin{equation}\label{non-summetric}
\mathrm{ET}_\lambda(\mu,\nu)= - b\lambda \nu(\G) + \sup \left\{ \int_\G f (\mu - \nu) :\, f\in \mathbb{U^*} \right\}.
\end{equation}
Now consider $f= \tilde f - \frac{b\lambda}{2}$. Then, $f\in \mathbb{U^*} $ if and only if $\tilde f\in \mathbb{U}$. Moreover,
 \[
 \int_\G f (\mu - \nu)  =- \frac{b\lambda}{2}\big[\mu(\G) -\nu(\G)\big] + \int_\G \tilde f (\mu - \nu) .
 \]
Therefore, the conclusion of the corollary follows from \eqref{non-summetric}.
\end{proof}

\subsubsection{Proof of Lemma~\ref{lm:lower-bound-part1}}

\begin{proof}[Proof of Lemma~\ref{lm:lower-bound-part1}]
By using part i) of Lemma~\ref{lm:lipschitz-vs-sobolev}, we see that 
\begin{align}\label{set-inclusion}
 \mathbb{U}_0 
 &\subset 
 \Big\{f\in W^{1,\infty}(\G, \omega^*):\, 
  -w_2(z_0) - \frac{b\lambda}{2}\leq f(z_0) \leq    w_1(z_0) + \frac{b\lambda}{2}, \, \|f'\|_{L^{\infty}(\G, \omega^*)}\leq b\Big\} =
 \mathbb U_{\infty}^0.
\end{align}
As a consequence, we obtain 
\begin{align*}
\mathrm{US}_1^0(\mu,\nu )
= \sup \Big[ \int_\G f (\mu - \nu):\, f\in  \mathbb U_{\infty}^0 \Big] 
\geq  \sup \Big[ \int_\G f (\mu - \nu):\, f\in \mathbb{U}_0  \Big].
\end{align*}
Thus the first statement of the lemma is proved. Now if $\G$ is a tree. Then Lemma~\ref{lm:lipschitz-vs-sobolev} implies that the inclusion in \eqref{set-inclusion} is actually the equality. That is,  $\mathbb{U}_0 
 =
 \mathbb U_{\infty}^0$.
Therefore, we get the desired identity
\begin{align*}
\mathrm{US}_1^0(\mu,\nu )
=   \sup \Big[ \int_\G f (\mu - \nu):\, f\in \mathbb{U}_0  \Big].
\end{align*}
\end{proof}

\subsubsection{Proof of Proposition~\ref{prop:closed-form}}

\begin{proof}[Proof of Proposition~\ref{prop:closed-form}]
It follows from Definition~\ref{def:discrepancy} and the representation \eqref{alternative_representation} for $f$ that
\begin{align*}
\mathrm{US}_p^\alpha(\mu,\nu )
&= \sup \Big\{
s[\mu(\G) -\nu(\G)] :\, s \in \big[ - \frac{b\lambda}{2} -w_2(z_0)+\alpha, w_1(z_0) + \frac{b\lambda}{2} -\alpha\big] \Big\}\\
&\hspace{9em} + \sup \left\{ \int_\G \Big[  \int_{[z_0,x]} h(y) \omega(\dd y)\Big] (\mu - \nu)(dx) :\,    \|h\|_{L^{p'}(\G, \omega)}\leq b \right\}.
\end{align*}
The first supremum equals to $[w_1(z_0) +\frac{b\lambda}{2}-\alpha] [ \mu(\G)- \nu(\G)]$  if $ \mu(\G)\geq \nu(\G)$ and equals to $-[w_2(z_0) +\frac{b\lambda}{2}-\alpha] [ \mu(\G)- \nu(\G)]$ if 
 $\mu(\G)< \nu(\G)$. 
 
 On the other hand,
by the same arguments as in the proof of \cite[Proposition~3.5]{le2022st}  we see that the second supremum  equals to 
$b \left(\int_{\G} | \mu(\Lambda(x)) -  \nu(\Lambda(x))|^p \, \omega(\dd x)\right)^\frac1p$. Putting them together, we obtain the desired formula for $\mathrm{US}_p^\alpha(\mu,\nu )$.
\end{proof}

\subsubsection{Proof of Corollary~\ref{cor:closed-form} }

\begin{proof}[Proof of Corollary~\ref{cor:closed-form}]
 We first recall that $\langle u,  v\rangle$ denotes the line segment in $\R^n$ connecting two points $u, v$, 
while $( u, v)$ means the same line segment but without its two end-points.
Then as $\omega(\{x\})= 0$ for every $x\in\G$, we have
\[
\int_{\G} | \mu(\Lambda(x)) -  \nu(\Lambda(x))|^p \, \omega(\dd x) 
=\sum_{e=\langle u,v\rangle\in E}   \int_{(u,v)} | \mu(\Lambda(x)) -  \nu(\Lambda(x))|^p \, \omega(\dd x).
\]
Since  $\mu$  and $\nu$ are supported on nodes, we can rewrite the above identity as
\begin{align*}
\int_{\G} | \mu(\Lambda(x)) -  \nu(\Lambda(x))|^p \, \omega(\dd x)  =
\hspace{-0.3em}\sum_{e=\langle u,v\rangle\in E}   \int_{(u,v)} \hspace{-1em}| \mu(\Lambda(x)\setminus (u,v)) -  \nu(\Lambda(x)\setminus (u,v))|^p \, \omega(\dd x). 
\end{align*}
For $e=\langle u,v\rangle$ and $x\in (u,v)$, we observe that  $y\in \G\setminus (u,v)$ belongs to $\Lambda(x)$ if and only if $y\in \gamma_e$. It follows that $\Lambda(x)\setminus (u,v) =\gamma_e$, and thus we deduce from the above identity that
\begin{align*}
\int_{\G} | \mu(\Lambda(x)) -  \nu(\Lambda(x))|^p \, \omega(\dd x)
&=\hspace{-0.3em}\sum_{e=\langle u,v\rangle\in E}   \int_{(u,v)} \hspace{-1em}| \mu(\gamma_e) -  \nu(\gamma_e)|^p \, \omega(\dd x)\\
&=\sum_{e\in E}    \big| \mu(\gamma_e) -  \nu(\gamma_e)\big|^p \omega(e).
\end{align*}
This together with  Proposition~\ref{prop:closed-form} yields the postulated result.
\end{proof}

\subsubsection{Proof of Proposition~\ref{geodesic-space}}\label{sec:app_geodesic-space-part2}

We begin with the following  auxiliary result.
\begin{lemma}\label{equal-measure}
Let $\mu,\nu\in \calM(\G)$. Then, $\mu=\nu$ if and only if  $\mu(\Lambda(x)) = \nu(\Lambda(x))$ for every $x$ in $\G$.
\end{lemma}
\begin{proof}
It is obvious that $\mu=\nu$ implies that $\mu(\Lambda(x)) = \nu(\Lambda(x))$  for every $x$ in $\G$.
Now assume that $\mu(\Lambda(x)) = \nu(\Lambda(x))$ for every $x$ in $\G$. We first claim that  $\mu(\{a\}) = \nu(\{a\})$ for any $a\in\G$.
Let $a\in\G$ be arbitray. Then there are two possibility for $a$: either $a$ is a node or $a$ is an interior point of an edge. We consider these two cases saperately.

$\bullet$ (i) $a$ is an interior point of an edge $e\in E$ (i.e. $a$ is not a node): 

Let $\{a_n\}_{n=1}^\infty$ be a sequence of distinct points on the same edge $e$ as $a$ such that $d_\G(a_n, z_0) > d_\G(a, z_0)$ for every $n\geq 1$ and 
$a_n \to a$ as $n\to \infty$. It follows that  $  \Lambda(a_n) \subset  \Lambda(a)$ and  $ \Lambda(a)\setminus \Lambda(a_n) \downarrow \{a\}$ as $n\to \infty$. As a consequence, we have
\begin{align*}
\mu(\{a\}) 
=  \lim_{n\to\infty}  \mu(\Lambda(a)\setminus \Lambda(a_n) )
=\lim_{n\to\infty} \big[ \mu(\Lambda(a)) -\mu(\Lambda(a_n))\big].
\end{align*}
But as $\mu(\Lambda(x)) = \nu(\Lambda(x))$ for every $x$ in $\G$, we thus obtain
\begin{align*}
\mu(\{a\}) 
=\lim_{n\to\infty} \big[ \nu(\Lambda(a)) -\nu(\Lambda(a_n))\big]
=  \lim_{n\to\infty}  \nu(\Lambda(a)\setminus \Lambda(a_n) ) =\nu(\{a\})
\end{align*}
as claimed.

$\bullet$ (ii) $a$ is a node: 

We can assume that $a$ is a common node for edges $e_1,..., e_k$.
Then for each $i\in \{1,...,k\}$, let 
$\{a^i_n\}_{n=1}^\infty$ be a sequence of distinct points on edge $e_i$ such that  $d_\G(a^i_n, z_0) > d_\G(a, z_0)$ for every $n\geq 1$ and 
$a^i_n \to a$ as $n\to \infty$. These choices yield
 $  \Lambda(a^i_n) \subset  \Lambda(a)$
and  $ \Lambda(a)\setminus \cup_{i=1}^k\Lambda(a^i_n) \downarrow \{a\}$ as $n\to \infty$.  Using this and the assumption $\mu(\Lambda(x)) = \nu(\Lambda(x))$ for every $x$ in $\G$, we obtain
\begin{align*}
  \mu(\{a\}) = \lim_{n\to\infty} \big[ \mu(\Lambda(a)) -\sum_{i=1}^k \mu(\Lambda(a^i_n))\big]
=\lim_{n\to\infty} \big[ \nu(\Lambda(a)) -\sum_{i=1}^k\nu(\Lambda(a^i_n))\big]=\nu(\{a\}). 
\end{align*}

Thus, we have  proved the claim  that  $\mu(\{a\}) = \nu(\{a\})$ for every $a\in\G$. 

On the other hand, for any points $x,y$ belonging to the same edge
\[
\mu(\langle x,y))= \mu(\Lambda(x))-\mu( \Lambda(y))=\nu(\Lambda(x))-\nu( \Lambda(y))
= \nu(\langle x,y)),
\] 
where $\langle x, y)$ denotes the line segment in $\R^n$ connecting two points $x, y$  but without its right end-point $x$ (while $\langle x, y\rangle$ include both end-points).

Thus, by combining them, we infer further that 
$\mu(\langle x,y\rangle)=  \nu(\langle x,y\rangle)$ for any $x,y\in e$ and for any edge $e\in E$. 
It follows that $\mu=\nu$, and the proof is complete.
\end{proof}

\begin{proof}[Proof of Proposition~\ref{geodesic-space}]
We note first that the quantity $\mathrm{US}_p^\alpha$ depends only on the values of the weights at the root $z_0$ of the graph. This comes from the fact that only $w_1(z_0)$ and $w_2(z_0)$ are used in the definition of $\mathbb U_{p'}^{\alpha}$. 

i)  This follows immediately from Proposition~\ref{prop:closed-form} in the main text.

ii) It follows from Definition~\ref{def:discrepancy}
that     $\mathrm{US}_p^\alpha(\mu,\mu)= 0$ and $\mathrm{US}_p^\alpha$ satisfies the triangle inequality. As  the constant function $f=0$ belongs to the constraint set $\mathbb U_{p'}^{\alpha}$,
we also have  $\mathrm{US}_p^\alpha(\mu,\nu)\geq  0$.
Next, assume that $\mathrm{US}_p^\alpha(\mu,\nu)=0$.  Then by  Proposition~\ref{prop:closed-form} in the main text, we get
\[
 b \left(\int_{\G} | \mu(\Lambda(x)) -  \nu(\Lambda(x))|^p \, \omega(\dd x)\right)^\frac1p +  \Theta |\mu(\G)-\nu(\G)|  = 0.
\]
As $ \Theta> 0$ by our assumption of $\alpha$, we  must have 
\[
\mu(\G) =\nu(\G) \quad \mbox{and}\quad \int_{\G} | \mu(\Lambda(x)) -  \nu(\Lambda(x))|^p \, \omega(\dd x)
 =0.
 \]
 Therefore,  $\mu(\Lambda(x)) = \nu(\Lambda(x))$ for every  $x\in\G$.
By using Lemma~\ref{equal-measure},  we then conclude that $\mu=\nu$. 

iii) Due to the assumption $w_1(z_0) = w_2(z_0)$ we have  $f\in \mathbb U_{p'}^{\alpha}$ if and only if $-f\in \mathbb U_{p'}^{\alpha}$. Hence we obtain from Definition~\ref{def:discrepancy}   that  $\mathrm{US}_p^\alpha(\mu,\nu)= \mathrm{US}_p^\alpha(\nu,\mu)$. This together with ii) implies that $(\calM(\G), \mathrm{US}_p^\alpha)$  is a metric space. Its completeness follows from \cite[Proposition 4]{P1}.  As a complete metric space, it is well known that $(\calM(\G), \mathrm{US}_p^\alpha)$ is a geodesic space if and only if for every $\mu,\nu \in\calM(\G)$ there exists $\sigma\in \calM(\G)$ such that 
\[
\mathrm{US}_p^\alpha(\mu,\sigma)= \mathrm{US}_p^\alpha(\nu,\sigma) =\frac12 \mathrm{US}_p^\alpha(\mu,\nu).
\]
To verify the latter, take $\sigma :=\frac{\mu +\nu }{2}$. Then using Definition~\ref{def:discrepancy}  in the main text, we  obtain 
\[
\mathrm{US}_p^\alpha(\mu,\sigma)= \frac12 \sup_{f\in \mathbb U_{p'}^{\alpha} } \int_\G f(\mu-\nu) = \frac12 \mathrm{US}_p^\alpha(\mu,\nu)
\]
and
\[
\mathrm{US}_p^\alpha(\nu,\sigma)= \frac12 \sup_{f\in \mathbb U_{p'}^{\alpha}} \int_\G f(\nu-\mu) = \frac12 \mathrm{US}_p^\alpha(\nu,\mu)= \frac12 \mathrm{US}_p^\alpha(\mu,\nu).
\]
\end{proof}

\subsubsection{Proof of Proposition~\ref{prop:D-via-ET}}

\begin{proof}[Proof of Proposition~\ref{prop:D-via-ET}]


i) From its definition, we have  $\mathbb{U}_\infty^\alpha = \mathbb{L}_\alpha$ with $\mathbb{L}_\alpha$ being the set defined in \cite[Section~3.2]{le2021ept}. As a consequence, we obtain $\mathrm{US}_1^\alpha(\mu,\nu )= d_\alpha(\mu,\nu)$. 
On the other hand, Proposition~\ref{prop:upper_for_US} yields for any $1\leq p \leq \infty$ that
\[
 \mathrm{US}_1^\alpha(\mu,\nu ) - \Theta|\mu(\G)-\nu(\G)|
\leq \omega^*(\G)^{\frac{1}{p'} } \, \, \Big[\mathrm{US}_p^\alpha(\mu,\nu ) -  \Theta|\mu(\G)-\nu(\G)|\Big].
\]
Therefore, we conclude that 
\[
\omega^*(\G)^{-\frac{1}{p'} } \Big[ d_\alpha(\mu,\nu) - \Theta|\mu(\G)-\nu(\G)|\Big]
\leq  \mathrm{US}_p^\alpha(\mu,\nu ) -  \Theta|\mu(\G)-\nu(\G)|.
\]
By moving and combining terms we arrive at
\[
 \mathrm{US}_p^\alpha(\mu,\nu ) \geq \omega^*(\G)^{-\frac{1}{p'}  } d_\alpha(\mu,\nu ) + \Theta [1- \omega^*(\G)^{-\frac{1}{p'}  }] |\mu(\G)-\nu(\G)|.
\]

ii) Let $\bar m \Let \mu(\G) =\nu(\G)$. From the definition of the $p$-Wasserstein distance, we have 
\begin{align*}
\calW_p(\mu,\nu)^p 
&= \inf_{\gamma \in \Pi(\mu,\nu)}\int_{\G\times\G} d_\G(x,y)^p \gamma(\dd x, \dd y)\\
&\leq    \big[\sup_{x,y\in\G} d_\G(x,y)\big]^{p-1}
\inf_{\gamma \in \Pi(\mu,\nu)}\int_{\G\times\G} d_\G(x,y) \gamma(\dd x, \dd y)\\
&= \big[\sup_{x,y\in\G} d_\G(x,y)\big]^{p-1} \calW_1(\mu,\nu),
\end{align*}
where 
\[
\Pi(\mu,\nu) \Let \Big\{ \gamma \in \calM(\G \times \G): \, \gamma(\G\times \G) = \bar{m}, \, \gamma_1= \mu, \, \gamma_2= \nu \Big\}.
\]
Therefore, the first statement will follow if we can show that
\begin{equation}\label{reduce-to-1}
 \mathrm{US}_p^\alpha(\mu,\nu ) \geq b \,  \calW_1(\mu,\nu).
\end{equation}

\smallskip
Since $\mu(\G) = \nu(\G)$, we have from Lemma~\ref{lm:sam-sobolev} that
 \begin{align*}
  \mathrm{US}_p^\alpha(\mu,\nu)
  &=\sup \Big\{ \int_\G f(\mu-\nu): \, f\in W^{1,p'}(\G, \omega),\, 
   \|f'\|_{L^{p'}(\G, \omega)}\leq b\Big\}.
 \end{align*}
 Hence by taking $g \Let f/b$, we can rewrite this identity as
 \begin{align*}
  \mathrm{US}_p^\alpha(\mu,\nu)
  &= b\sup \Big\{ \int_\G g(\mu-\nu): \, g\in W^{1,p'}(\G, \omega),\, 
   \|g'\|_{L^{p'}(\G, \omega)}\leq 1\Big\}\\
&=b\, \calS_p(\mu,\nu ),
 \end{align*}
 where  $\calS_p$ is the balanced Sobolev transport distance defined in \cite[Definition~3.2]{le2022st}.
 On the other hand,  we have 
$\calS_p(\mu,\nu ) \geq \omega^*(\G)^{-\frac{1}{p'}  } \calW_1(\mu,\nu)$ by \cite[Lemma~4.3]{le2022st}.
Therefore, we obtain  \eqref{reduce-to-1} as desired.

Alternatively, we can derive  \eqref{reduce-to-1}
as follows. 
 By using $\mathbb{U}_\infty^\alpha = \mathbb{L}_\alpha$ as in the proof of part i) and the observation about the translation invariant in the proof of Lemma~\ref{lm:sam-sobolev}, 
 we see that
\begin{align*}
 d_\alpha(\mu,\nu ) 
 &= \sup \Big\{  \int_\G  f(\mu-\nu):\, f\in \mathbb{U}_\infty^\alpha\Big\}\\
 &=\sup \Big\{ \int_\G f(\mu-\nu): \, f\in W^{1,\infty}(\G, \omega^*),\, 
   \|f'\|_{L^{\infty}(\G, \omega^*)}\leq b\Big\}.
\end{align*}
Then due to Lemma~\ref{lm:lipschitz-vs-sobolev}, we can further rewrite as
\begin{align*}
    d_\alpha(\mu,\nu ) 
    &=\sup \Big\{ \int_\G f(\mu-\nu): \, f\in C(\G),\, 
   |f(x)-f(y) |\leq b \, d_\G(x,y)\Big\}\\
    &=b \sup \Big\{ \int_\G g(\mu-\nu): \, g\in C(\G),\, 
   |g(x)-g(y) |\leq 1 \, d_\G(x,y)\Big\}\\
   &= b \, \calW_1(\mu,\nu ).
\end{align*}
On the other hand,  part i) above gives 
\[
\mathrm{US}_p^\alpha(\mu,\nu ) \geq \omega^*(\G)^{-\frac{1}{p'}  } d_\alpha(\mu,\nu ).
\]

Therefore, we  obtain
\[
\mathrm{US}_p^\alpha(\mu,\nu ) \geq b\, \omega^*(\G)^{-\frac{1}{p'}  } \calW_1(\mu,\nu ),
\]
for every $1\leq p \leq \infty$. 

\smallskip
For $p=1$, the equality happens since $p' = \infty$ and
\[
\mathrm{US}_1^\alpha(\mu,\nu )
= \sup \big\{  \int_\G  f(\mu-\nu):\, f\in \mathbb{U}_\infty^\alpha\big\} = b\, \calW_1(\mu,\nu ).
\]
Thus, the second statement follows.

\end{proof}

\subsubsection{Proof of Proposition~\ref{prop:negative_definite}}

\begin{proof}[Proof of Proposition~\ref{prop:negative_definite}]
We first prove that  $\ell_p$ distance is negative definite for $1 \le p \le 2$, where
\[
\ell_p(x, z) \Let \left(\sum_{i=1}^m \left| x_{(i)} - z_{(i)} \right|^p \right)^{1/p} \quad \mbox{for}\quad x, z \in \R^{m}.
\]

It is easy to see that the function $(u, v) \mapsto (u - v)^2$ is negative definite for $u, v \in \R$. Using this and by applying \cite[Corollary 2.10]{Berg84}, the function $(u, v) \mapsto (u - v)^p$ is negative definite for $1 \le p \le 2$. 

Therefore, for $1 \le p \le 2$, the function $\ell_p^p$ is negative definite since it is a sum of negative definite functions. Using this and by applying \cite[Corollary 2.10]{Berg84}, we have $\ell_p$ is negative definite for $1 \le p \le 2$.

We are now ready to prove the Proposition~\ref{prop:negative_definite}. From Proposition~\ref{prop:closed-form}, we have
\begin{equation*}
\mathrm{US}_p^\alpha(\mu,\nu ) = b\, \Big(\sum_{e \in E} w_e \left|\mu(\gamma_e) - \nu(\gamma_e) \right|^{p}\Big)^{\frac{1}{p}}+   \Theta |\mu(\G)-\nu(\G)|.
\end{equation*}
Let $m = |E|$. Then, $\mu \mapsto \Big\{w_e^{\frac{1}{p}}  \mu(\gamma_e) \Big\}_{e \in E}$ can be regarded as a feature map for measure $\mu$ onto $\R_{+}^{m}$. Therefore, the first term of $\mathrm{US}_p^\alpha$ is equivalent to $b$ times the $\ell_p$ distance between two feature maps of measures $\mu, \nu$ on $\R^{m}_{+}$ respectively. Recall that $b \ge 0$. Thus, the first term of $\mathrm{US}_p^\alpha$ is negative definite for $1 \le p \le 2$.

Additionally, the second term of $\mathrm{US}_p^\alpha$ is $\Theta$ times the $\ell_1$ distance between $\mu(\G)$ and $\nu(\G)$. Since $w_1(z_0) = w_2(z_0)$ and $\alpha \le \frac{b\lambda}{2} +w_1(z_0)$,
we also have from \eqref{def:M} that $\Theta = w_1(z_0) +\frac{b\lambda}{2} -\alpha\ge 0$. Therefore, the second term of $\mathrm{US}_p^\alpha$ is also negative definite.

Hence, $\mathrm{US}_p^\alpha$ is negative definite for any $1 \le p \le 2$.
\end{proof}

\section{FURTHER RESULTS AND DISCUSSIONS}\label{app:results_discussions}

\subsection{Brief Reviews}\label{app:subsec:review}

We give brief reviews for some definitions used in our work.


\subsubsection{Length Measure on Graphs}
\label{sec:length_measure}

We recall the definition and properties in \cite[\S4.1]{le2022st} about the length measure on graphs.

\begin{definition}[Length measure] \label{def:measure} 
Let $ \omega^*$ be the unique Borel measure on $\G$ such that the restriction of $\omega^*$ on any edge is the length measure of that edge. That is, $\omega^*$  satisfies:
\begin{enumerate}
\item[i)] For  any edge $e$ connecting two nodes $u$ and $v$, we have 
 $\omega^*(\langle x,y\rangle) = (t-s) w_e$ 
 whenever $x = (1-s) u + s v$ and $y = (1-t)u + t v$ for $s,t \in [0,1)$ with $s \leq t$. Here, $\langle x,y\rangle$ is the line segment in $e$ connecting $x$ and $y$.
 \item[ii)] For any Borel set $F \subset \G$, we have
 \[
 \omega^*(F) = \sum_{e\in E} \omega^*(F\cap e).
 \]
\end{enumerate}
\end{definition}

The next lemma asserts that $\omega^*$ is closely connected to the graph metric $d_\G$, and thus justifies the terminology of a length measure.
\begin{lemma}[$\omega^*$ is the length measure on graph] \label{lem:length-measure}
Suppose that $\G$ has no short cuts, namely, any edge $e$ is a shortest path connecting its two end-points. Then, $\omega^*$ is a length measure in the sense that
\[
\omega^*([x,y]) = d_\G(x,y)
\]
for  any  shortest path   $[x,y]$ connecting $x$ and $y$. In particular, $\omega^*$ has no atom in the sense that $\omega^*(\{x\})=0$ for every $x$ in $\G$. 
\end{lemma}

\subsubsection{Wasserstein distances}
We recall here the definition of the $p$-Wasserstein distances with graph metric ground cost on $\G$.

\begin{definition}
Let $1\leq p <\infty$. Suppose that  $\mu$ and $\nu$ are  two nonnegative Borel measures on $\G$ satisfying $\mu(\G) =\nu(\G)$. Then the  $p$-Wasserstein distance between $\mu$ and $\nu$ is defined by  
\begin{align*}
\calW_p(\mu,\nu)^p 
&= \inf_{\gamma \in \Pi(\mu,\nu)}\int_{\G\times\G} d_\G(x,y)^p \gamma(\dd x, \dd y),
\end{align*}
where 
\[
\Pi(\mu,\nu) \Let \Big\{ \gamma \in \calM(\G \times \G): \, \gamma(\G\times \G) = \bar{m}, \, \gamma_1= \mu, \, \gamma_2= \nu \Big\}
\]
with $\bar m \Let \mu(\G) = \nu(\G)$.
\end{definition}

\subsubsection{Kernels}
 We review some important definitions  and theorems/corollaries about kernels that are used in our work. 
\begin{itemize}

\item \textbf{Positive Definite Kernels \cite[pp.~66--67]{Berg84}.} A kernel function $k: \Omega \times \Omega \rightarrow \R$ is called positive definite if for every positive integer $m$ and every points $x_1, x_2, ..., x_m \in \Omega$, we have 
\[
\sum_{i, j=1}^m c_i c_j k(x_i, x_j) \ge 0 \qquad \mbox{for every}\quad c_1,...,c_m \in \R.
\]

\item \textbf{Negative Definite Kernels \cite[pp.~66--67]{Berg84}.} A kernel function $k: \Omega \times \Omega \rightarrow \R$ is  called negative definite if for every integer $ m \ge 2$ and every points $x_1, x_2, ..., x_m \in \Omega$, we have 
\[
\sum_{i, j=1}^m c_i c_j k(x_i, x_j) \le 0, \qquad \mbox{for every}\quad c_1,...,c_m \in \R\,\, \text{ s.t. } \, \sum_{i=1}^m c_i = 0.
\]

\item \textbf{Theorem 3.2.2 in \cite[pp.~74]{Berg84}.}
Let  $\kappa$ be a \textit{negative definite} kernel. Then for every $t>0$, the  kernel 
\[
k_{t}(x, z) \Let \exp{\left(- t \kappa(x, z)\right)}
\]
is positive definite.

\item \textbf{Definition 2.6 in \cite[pp.~76]{Berg84}.}
A positive definite kernel $\kappa$ is called \emph{infinitely divisible} if for each $n \in {\N}^{*}$, there exists a positive definite kernel $\kappa_n$ such that 
\[
\kappa = (\kappa_n)^n.
\]

\item \textbf{Corollary 2.10 in \cite[pp.~78]{Berg84}.}
Let  $\kappa$ be a \textit{negative definite} kernel. Then for $0 < t < 1$, the  kernel 
\[
k_{t}(x, z) \Let \left[\kappa(x, z)\right]^{t}
\]
is negative definite.

\end{itemize}

\subsection{Further Discussions}\label{app:subsec:discussion}

In this subsection, we discuss some extension for our work and describe more details for some parts in the main manuscript.

\paragraph{Path length for points in $\G$.} We can canonically measure a path length connecting any two points  $x, y \in \G$ where $x, y$ are not necessary to be nodes in $V$. Indeed, for two points $x, y \in \R^n$ belonging to the same edge $e= \langle u, v\rangle$ which connects two nodes $u$ and $v$ in $V$, then we have 
\begin{eqnarray*}
& x = (1-s) u + s v, \\
& y = (1-t)u + t v,
\end{eqnarray*}
for some numbers $t,s\in [0,1]$. Therefore, the length of the path connecting $x$ and $y$ along the edge $e$ (i.e., the line segment $\langle x, y\rangle$) is defined by $|t-s| w_e$. Hence, the length for an arbitrary path in $\G$ can be similarly defined by breaking down into pieces over edges and summing over their corresponding lengths~\citep{le2022st}.

\paragraph{Lipschitz nonnegative weight function on graph $\G$.} An example of $b$-Lipschitz nonegative weight function on $\G$ is
\[
w(x) = a_1 d_{\G}(z_0, x) + a_0,
\]
for some constants $a_1 \in [0, b]$ and $a_0 \in [0, \infty)$.

\paragraph{Extension to measures supported on $\G$.} The closed-form formula for $\mathrm{US}_p^\alpha$ in \eqref{equ:closed_form_US} can be extended for measures with finite supports on $\G$ (i.e., measures which may have supports on edges) by using the same strategy to measure a path length connecting $z_0$ and y for any $z_0, y \in \G$ (see \S\ref{sec:pre}). More precisely, we break down edges containing supports into pieces and sum over their corresponding values instead of the sum over edges for $\mathrm{US}_p^\alpha$ in \eqref{equ:closed_form_US}.

\paragraph{About the assumption of uniqueness property of the shortest paths on $\G$.} As discussed in the supplementary of \citep{le2022st}, since $w_e \in \R$ for any edge $e \in E$ of graph $\G$., it is almost surely that every node in the graph can be regarded as unique-path root node (with a high probability, lengths of paths connecting any two nodes in graph $\G$ are different). Additionally, for some special graph, e.g., a grid of nodes, there is no unique-path root node for such graph. However, by perturbing each node of such graph (or lengths of edges in $\G$ in case $\G$ is a non-physical graph, i.e., $w_e$) with a small deviation $\varepsilon$, we can obtain a graph satisfying the unique-path root node assumption.

\paragraph{About the unbalanced Sobolev transport.} Similar to the work~\citep{le2022st}, we assume that we know the graph metric space (i.e., the graph structure) where supports of measures are belongs to. Giving such graph, we define the unbalanced Sobolev transport for measures which may have different total mass and are supported on that graph metric space. We leave a question to learn an optimal graph metric structure from data (i.e., supports of measures) for unbalanced Sobolev transport for future work.

\paragraph{About graphs $\G_{\text{Log}}$ and $\G_{\text{Sqrt}}$~\citep{le2022st}.} First, we use a clustering method, e.g., the farthest-point clustering, to partition supports of measures into at most $M$ clusters.\footnote{$M$ is the input number of clusters for the clustering method. Therefore, the result has at most $M$ clusters depending on input data.} Then, let $V$ denote the set of centroids of these clusters. For edges, in graph $\G_{\text{Log}}$, we randomly choose $M\log(M)$ edges; and $M^{3/2}$ edges for graph $\G_{\text{Sqrt}}$, we also denote the set of those sampled edges as $\tilde{E}$.  

For each edge $e$, its corresponding weight $w_e$ is computed by the Euclidean distance between the two corresponding nodes of $e$. Let $n_c$ be the number of connected components in the graph $\tilde{\G}(V, \tilde{E})$, we then randomly add $(n_c - 1)$ more edges between these $n_c$ connected components to construct a connected graph $\G$ from $\tilde{\G}$.Let $E_c$ be the set of these $(n_c - 1)$ added edges and denote set $E = \tilde{E} \cup E_c$, then $\G(V, E)$ is the considered graph.

\paragraph{Datasets and Computational Devices.} For document dataset (i.e., \texttt{TWITTER, RECIPE, CLASSIC, AMAZON}), orbit dataset (\texttt{Orbit}) and a $10$-class subset of \texttt{MPEG7} dataset, one can contact the authors of~\citep{le2022st} to access to these datasets. For computational devices, we run all of our experiments on commodity hardware.

\subsection{Further Empirical Results}\label{app:sec:furtherempiricalresults}

In this subsection, we provide further empirical results for our work.

\subsubsection{Extended Empirical Results for the Main Text}

\begin{figure}[h]
  \begin{center}
    \includegraphics[width=0.5\textwidth]{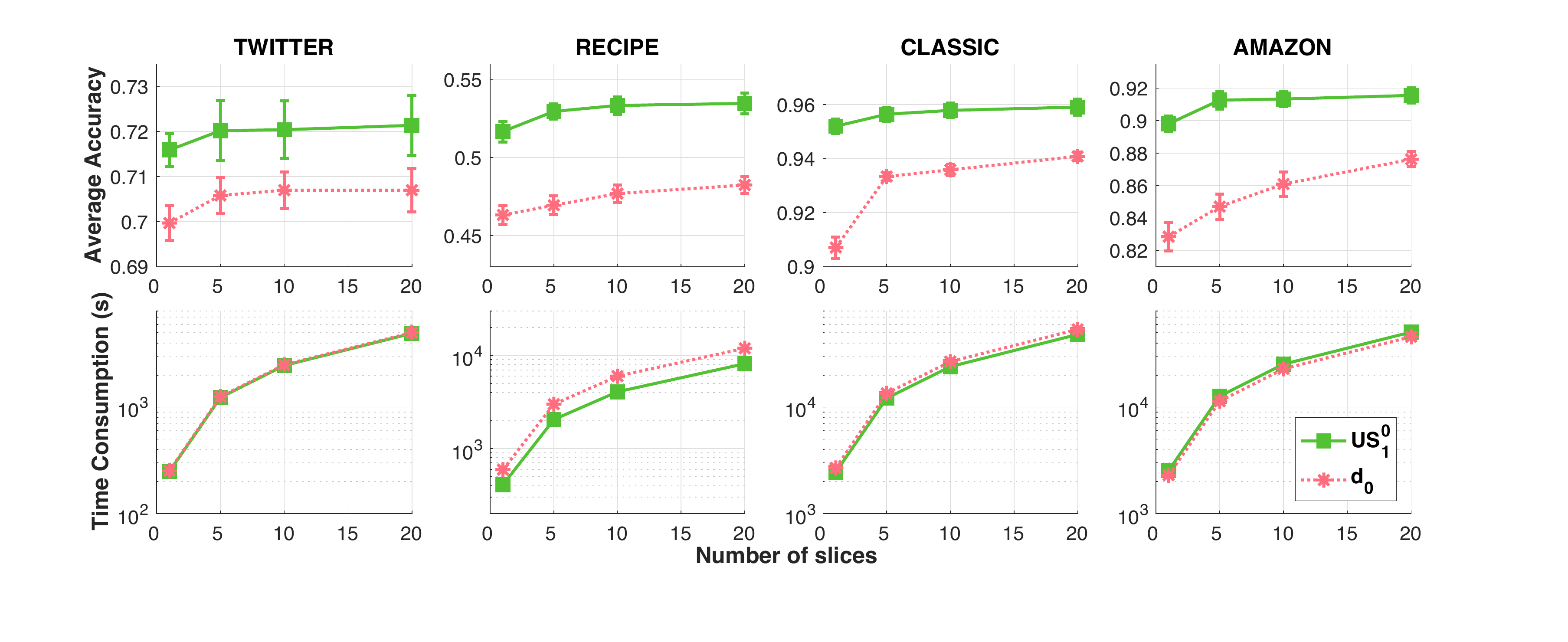}
  \end{center}
  \vspace{-10pt}
  \caption{SVM results and time consumption for kernel matrices of slice variants for UST and EPT on a tree in document classification with graph $\G_{\text{Sqrt}}$.}
  \label{fg:DOC_10KSqrt_SLICE_main_app}
 \vspace{-6pt}
\end{figure}

Similar to Figure 3 in the main text for TDA, we illustrate the effect of the number of slices for document classification with graph $\G_{\text{Sqrt}}$ in Figure~\ref{fg:DOC_10KSqrt_SLICE_main_app}.

\begin{figure}[h]
  \begin{center}
    \includegraphics[width=0.5\textwidth]{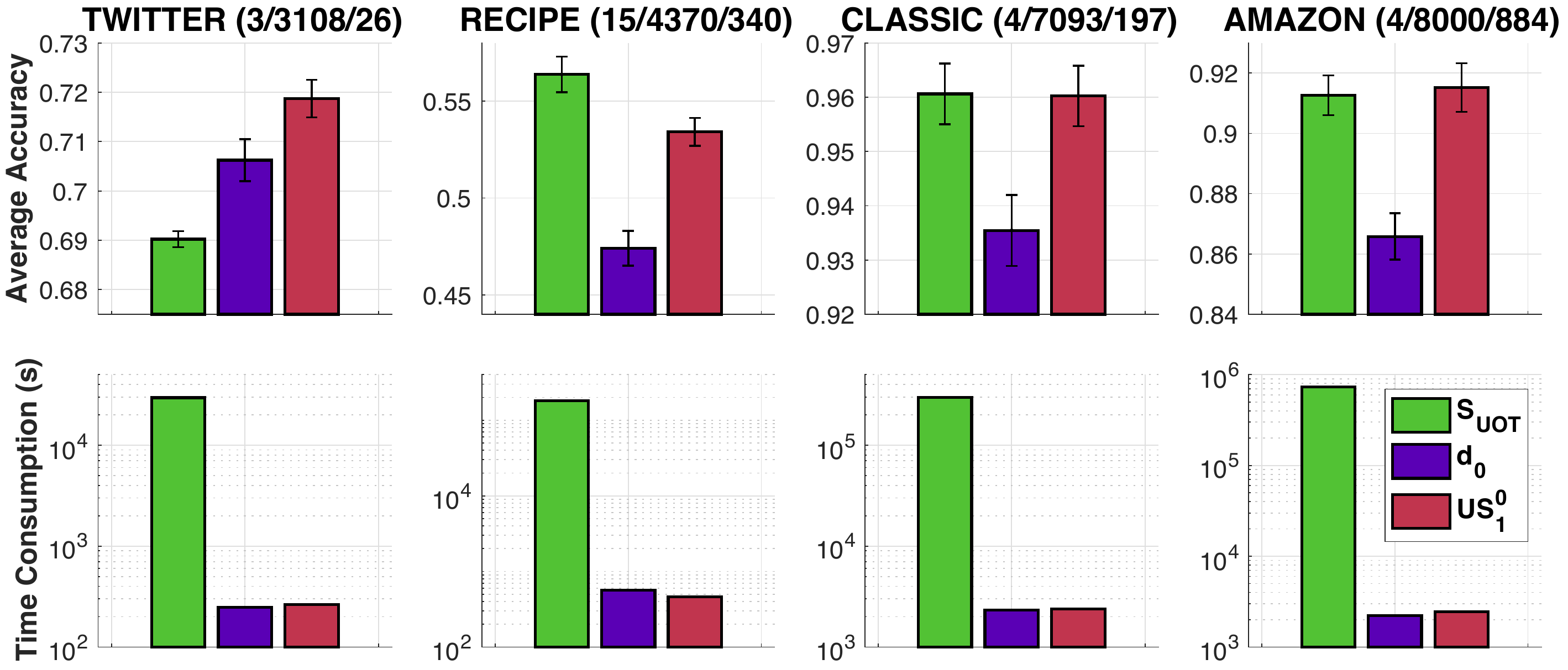}
  \end{center}
  \vspace{-10pt}
  \caption{SVM results and time consumption for kernel matrices in document classification with graph $\G_{\text{Log}}$. For each dataset, the numbers in the parenthesis are the number of classes; the number of documents; and the maximum number of unique words for each document respectively.}
  \label{fg:DOC_10KLog_main_app}
 \vspace{-6pt}
\end{figure}

\begin{figure}[h]
  \begin{center}
    \includegraphics[width=0.27\textwidth]{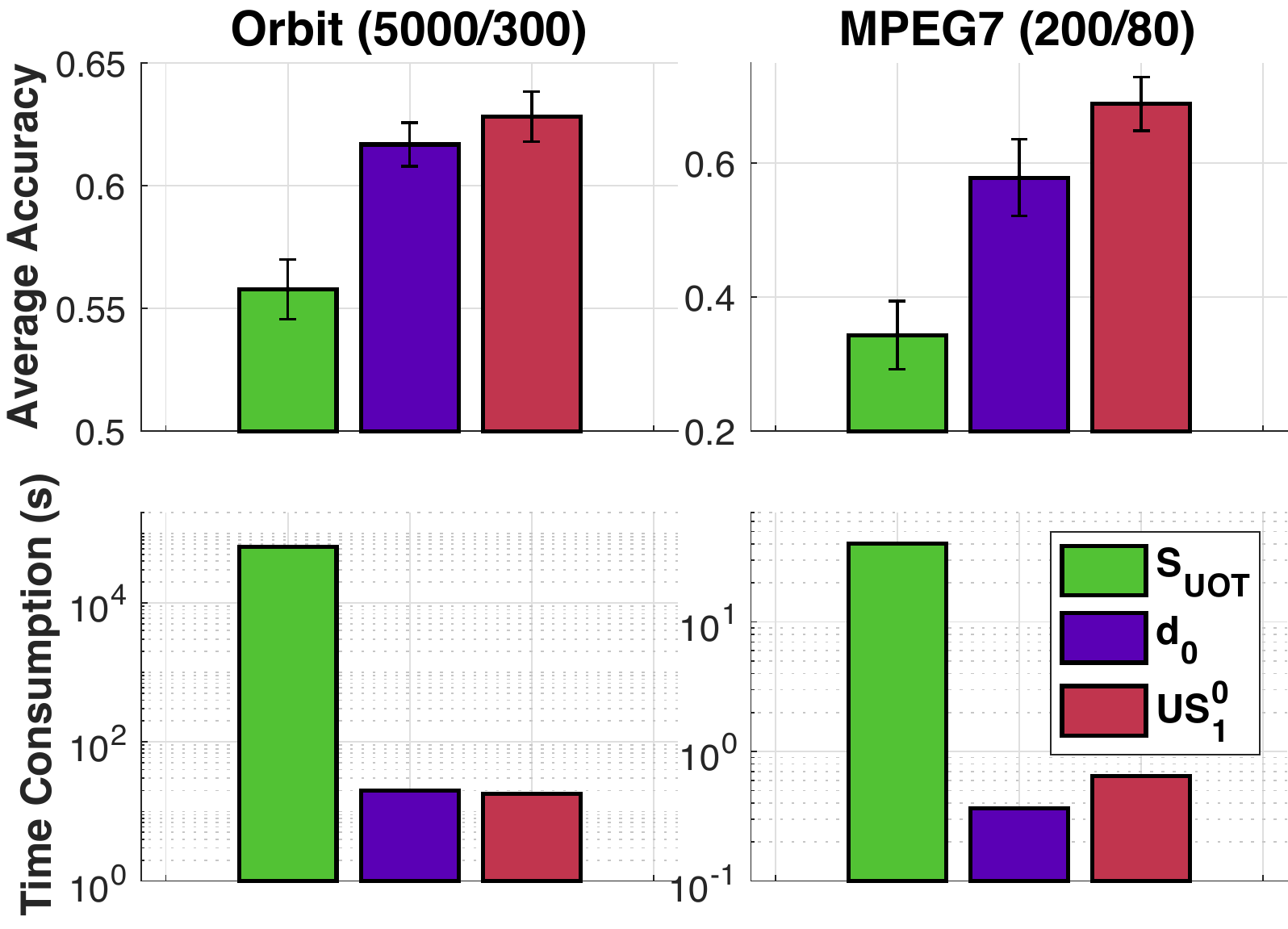}
  \end{center}
  \vspace{-10pt}
  \caption{SVM results and time consumption for kernel matrices in TDA with graph $\G_{\text{Log}}$. For each dataset, the numbers in the parenthesis are respectively the number of PD; and the maximum number of points in PD.}
  \label{fg:TDA_mix1K100Log_main_app}
 \vspace{-6pt}
\end{figure}

\begin{figure}[h]
  \begin{center}
    \includegraphics[width=0.5\textwidth]{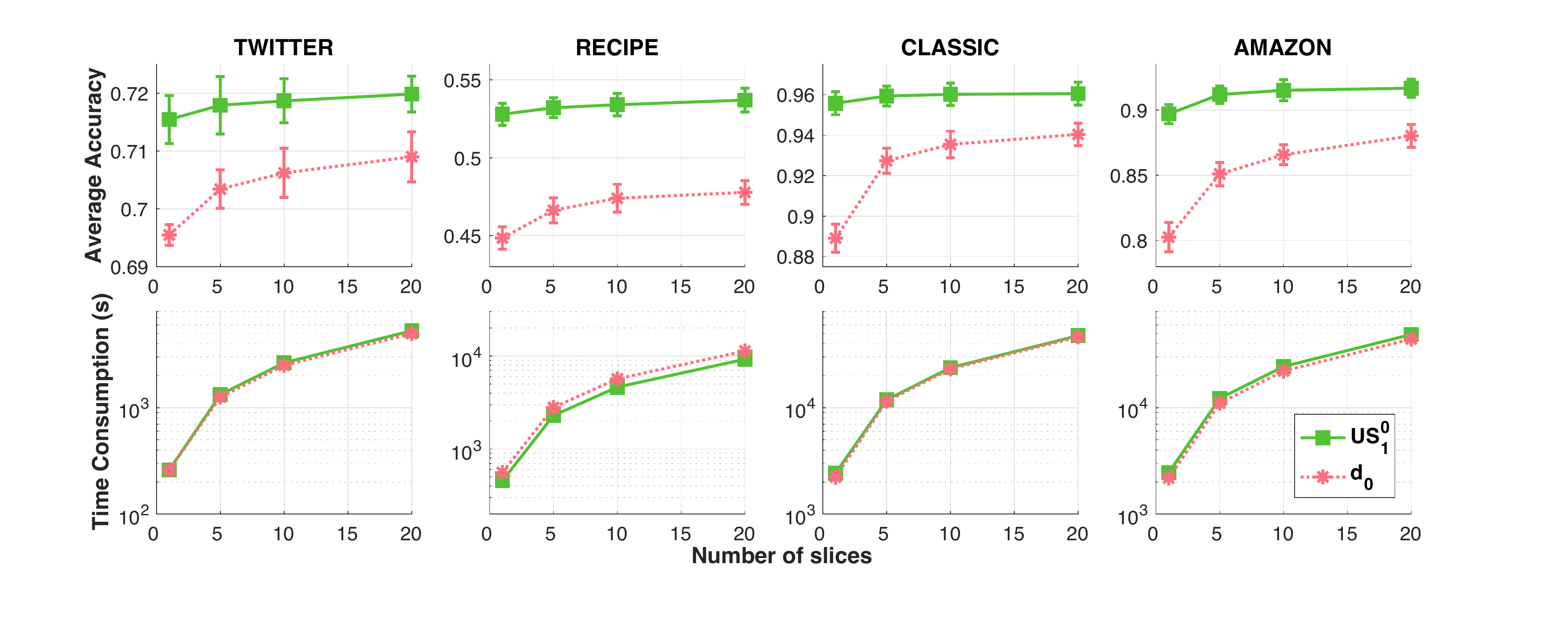}
  \end{center}
  \vspace{-10pt}
  \caption{SVM results and time consumption for kernel matrices of slice variants for UST and EPT on a tree in document classification with graph $\G_{\text{Log}}$.}
  \label{fg:DOC_10KLog_SLICE_main_app}
 \vspace{-6pt}
\end{figure}

\begin{figure}[h]
  \begin{center}
    \includegraphics[width=0.27\textwidth]{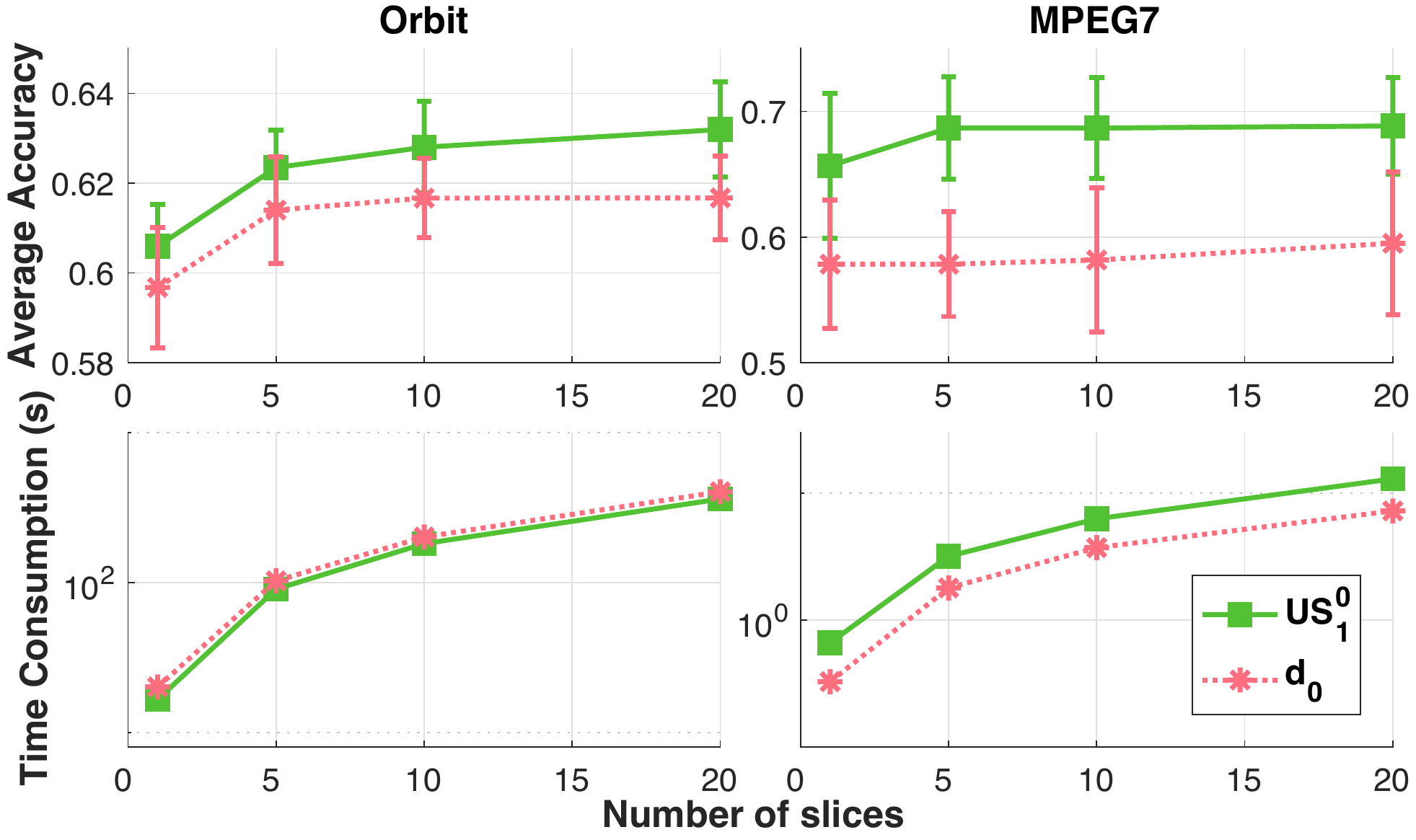}
  \end{center}
  \vspace{-10pt}
  \caption{SVM results and time consumption for kernel matrices of slice variants for UST and EPT on a tree in TDA with graph $\G_{\text{Log}}$.}
  \label{fg:TDA_10KLog_SLICE_main_app}
 \vspace{-6pt}
\end{figure}

We also consider a graph $\G$ with a different setting:$\G_{\text{Log}}$. Recall that for Figure 1, Figure 2, Figure 3 in the main text and Figure~\ref{fg:DOC_10KSqrt_SLICE_main_app}, results are for graph $\G_{\text{Sqrt}}$ where $M=10^4$ for document datasets, $M=10^3$ for \texttt{MPEG7} dataset and $M=10^2$ for \texttt{Orbit} dataset.\footnote{There is a typo in the main text (\S6): It should be $M=10^3$ is for \texttt{MPEG7} and $M=10^2$ is for \texttt{Orbit}.} We illustrate corresponding results for graph $\G_{\text{Log}}$ in Figure~\ref{fg:DOC_10KLog_main_app}, Figure~\ref{fg:TDA_mix1K100Log_main_app}, Figure~\ref{fg:DOC_10KLog_SLICE_main_app}, and Figure~\ref{fg:TDA_10KLog_SLICE_main_app} respectively.

\subsubsection{Further Empirical Results}

We also provides further results for document classification and TDA as follow:

\paragraph{For document classification.}
\begin{itemize}
    \item For $M=10^2$, we illustrate the SVM results and time consumption for kernels matrices and the effect of the number of slices for graph $\G_{\text{Sqrt}}$ in Figure~\ref{fg:DOC_100Sqrt_app} and Figure~\ref{fg:DOC_100Sqrt_SLICE_app} respectively. The corresponding results for graph $\G_{\text{Log}}$ are in Figure~\ref{fg:DOC_100Log_app} and Figure~\ref{fg:DOC_100Log_SLICE_app}.
 
 \begin{figure}[h]
  \vspace{-2pt}
  \begin{center}
    \includegraphics[width=0.5\textwidth]{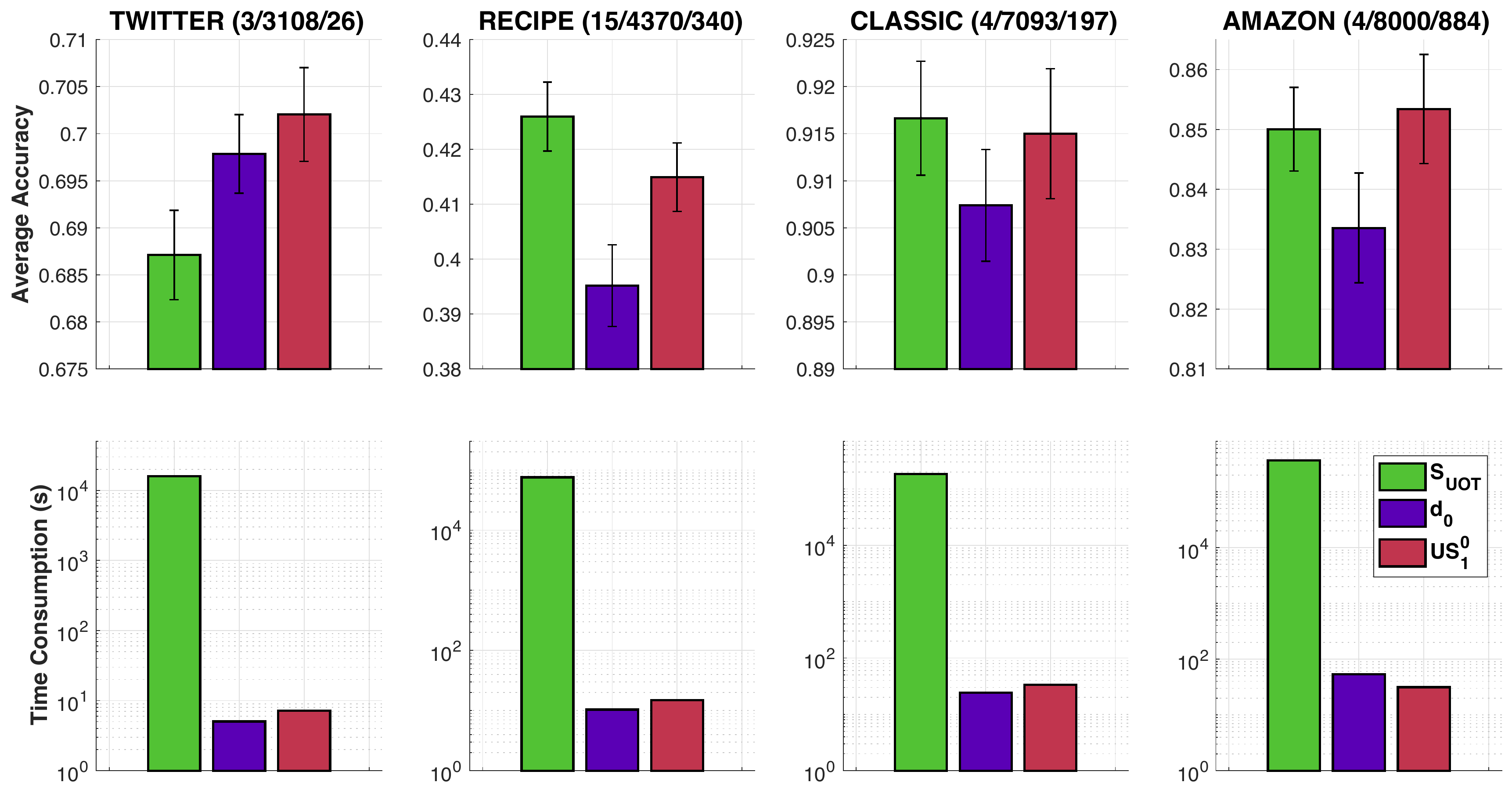}
  \end{center}
  \vspace{-10pt}
  \caption{SVM results and time consumption for kernel matrices in document classification with graph $\G_{\text{Sqrt}}$ with $M=10^2$.}
  \label{fg:DOC_100Sqrt_app}
 \vspace{-6pt}
\end{figure}

\begin{figure}[h]
  \vspace{-2pt}
  \begin{center}
    \includegraphics[width=0.5\textwidth]{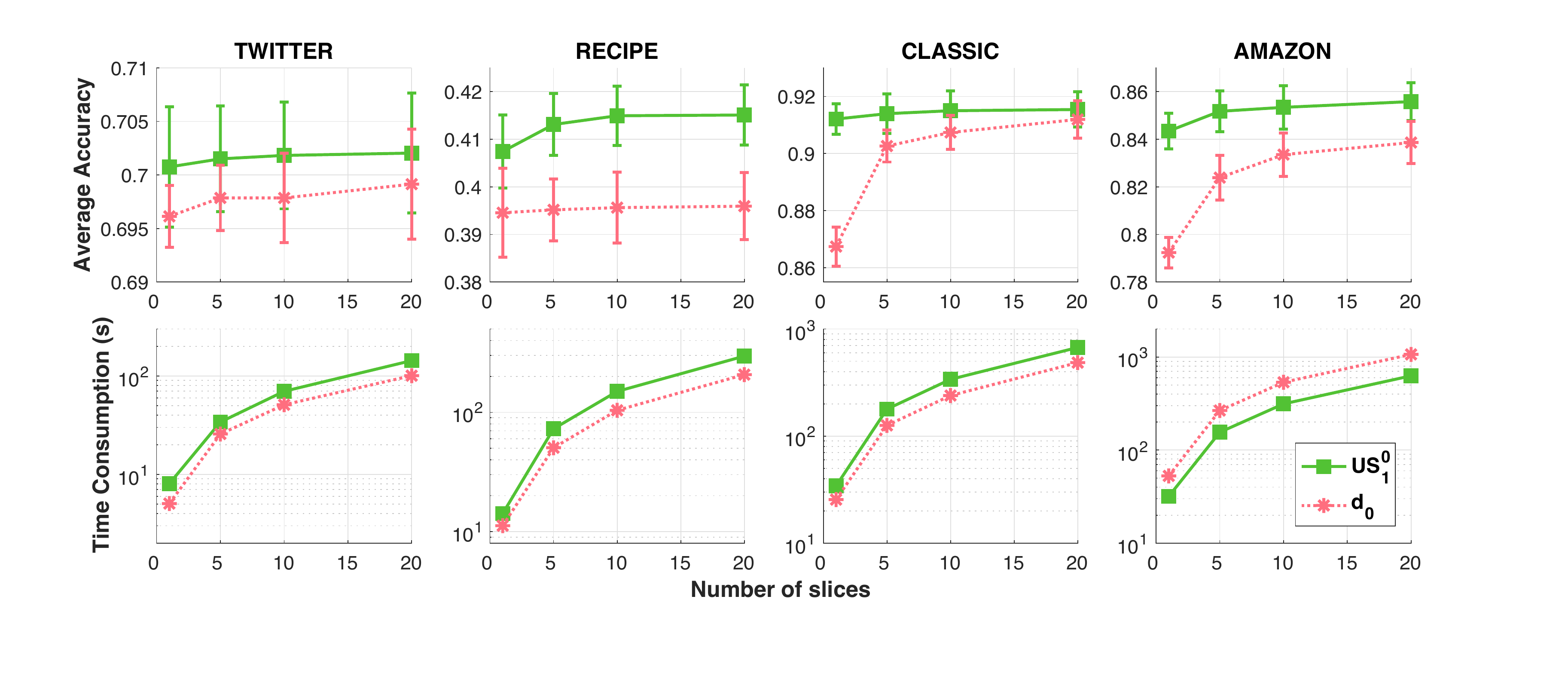}
  \end{center}
  \vspace{-10pt}
  \caption{SVM results and time consumption for kernel matrices of slice variants for UST and EPT on a tree in document classification with graph $\G_{\text{Sqrt}}$ with $M=10^2$.}
  \label{fg:DOC_100Sqrt_SLICE_app}
 \vspace{-6pt}
\end{figure}

 \begin{figure}[h]
  \vspace{-2pt}
  \begin{center}
    \includegraphics[width=0.5\textwidth]{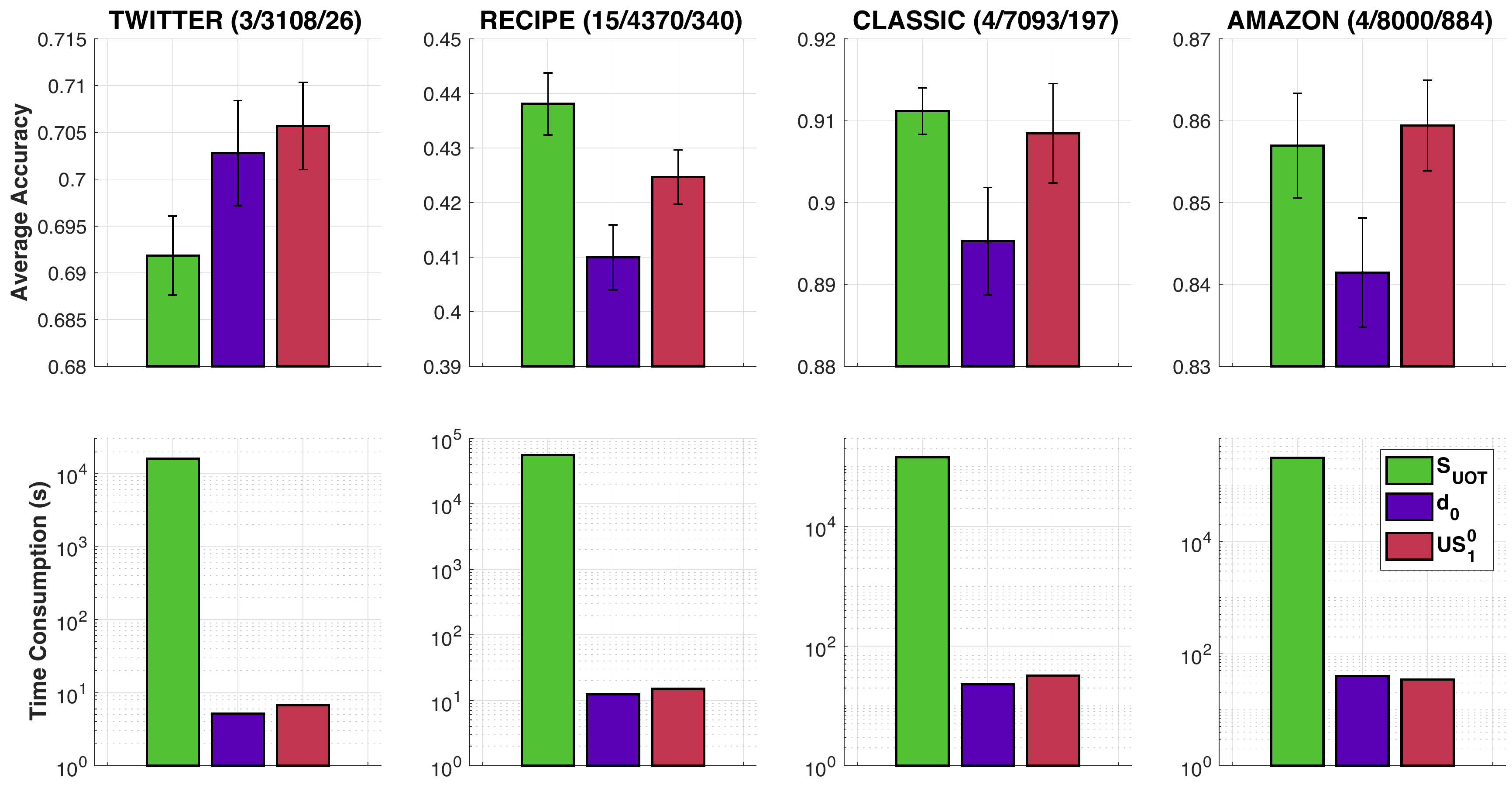}
  \end{center}
  \vspace{-14pt}
  \caption{SVM results and time consumption for kernel matrices in document classification with graph $\G_{\text{Log}}$ with $M=10^2$.}
  \label{fg:DOC_100Log_app}
 \vspace{-6pt}
\end{figure}

\begin{figure}[h]
  \vspace{-2pt}
  \begin{center}
    \includegraphics[width=0.5\textwidth]{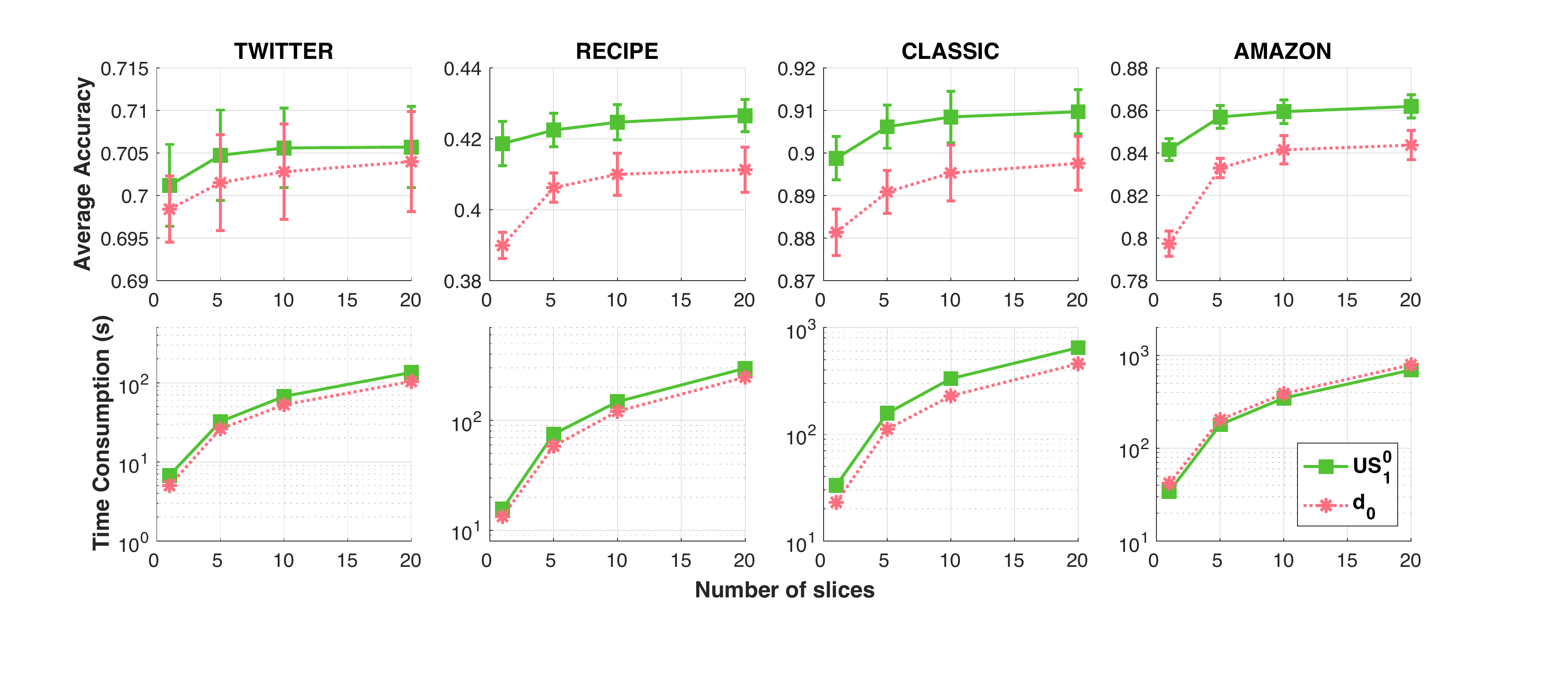}
  \end{center}
  \vspace{-14pt}
  \caption{SVM results and time consumption for kernel matrices of slice variants for UST and EPT on a tree in document classification with graph $\G_{\text{Log}}$ with $M=10^2$.}
  \label{fg:DOC_100Log_SLICE_app}
 \vspace{-6pt}
\end{figure}

    \item For $M=10^3$, we illustrate the SVM results and time consumption for kernels matrices and the effect of the number of slices for graph $\G_{\text{Sqrt}}$ in Figure~\ref{fg:DOC_1KSqrt_app} and Figure~\ref{fg:DOC_1KSqrt_SLICE_app} respectively. The corresponding results for graph $\G_{\text{Log}}$ are in Figure~\ref{fg:DOC_1KLog_app} and Figure~\ref{fg:DOC_1KLog_SLICE_app}.
 
 \begin{figure}[h]
  \vspace{-2pt}
  \begin{center}
    \includegraphics[width=0.5\textwidth]{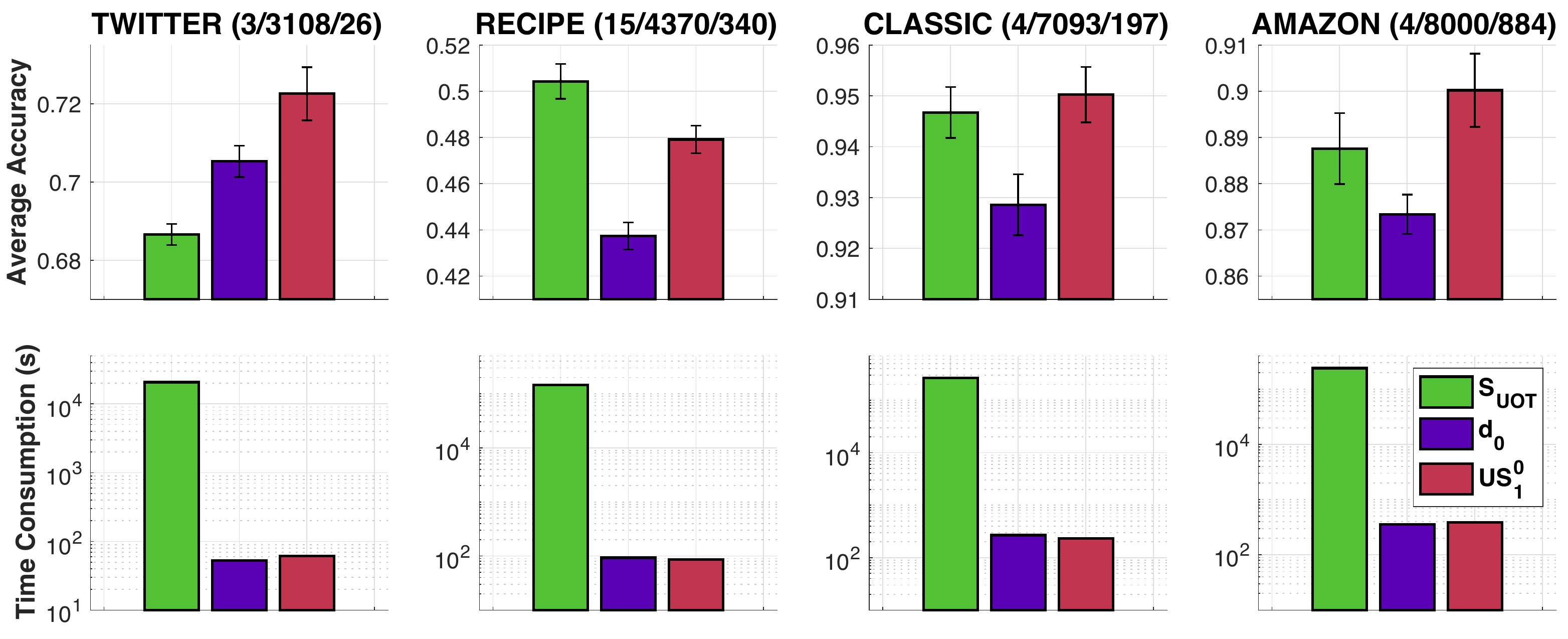}
  \end{center}
  \vspace{-14pt}
  \caption{SVM results and time consumption for kernel matrices in document classification with graph $\G_{\text{Sqrt}}$ with $M=10^3$.}
  \label{fg:DOC_1KSqrt_app}
 \vspace{-6pt}
\end{figure}

\begin{figure}[h]
  \vspace{-2pt}
  \begin{center}
    \includegraphics[width=0.5\textwidth]{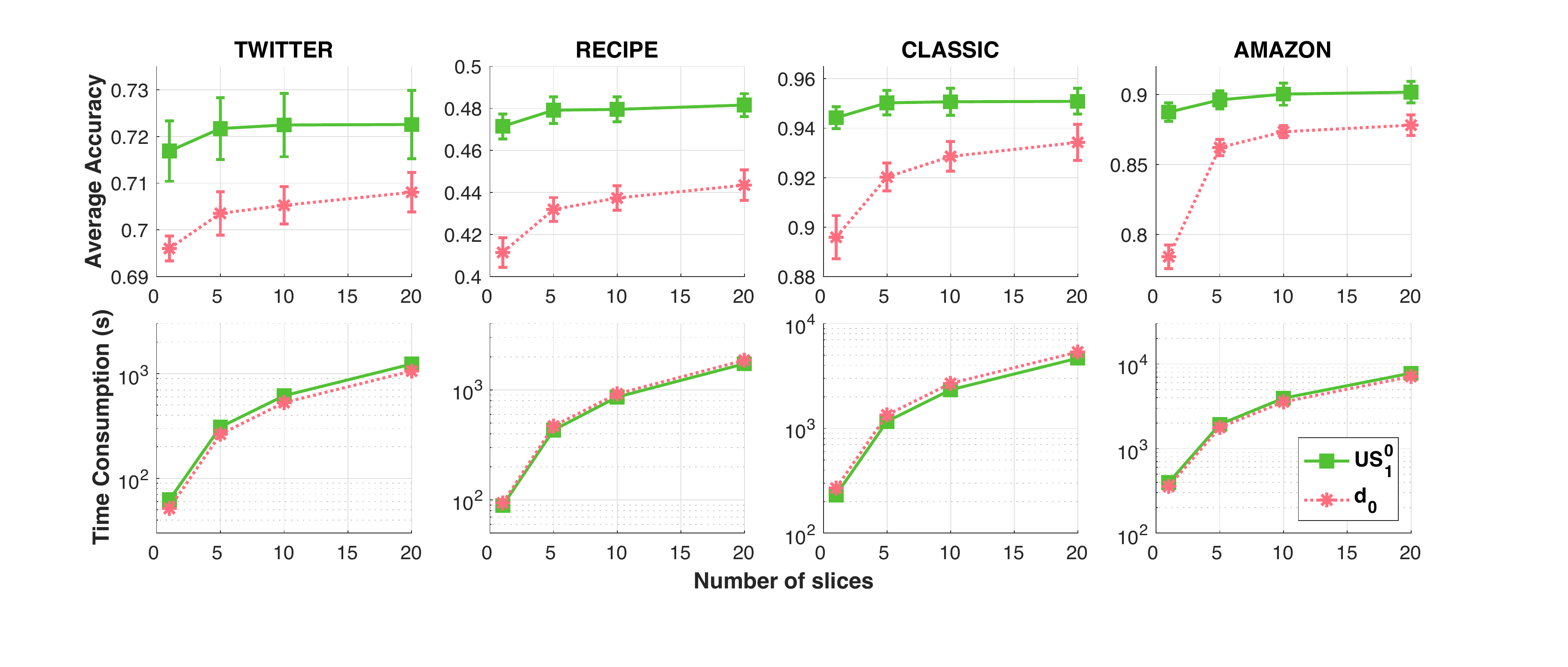}
  \end{center}
  \vspace{-14pt}
  \caption{SVM results and time consumption for kernel matrices of slice variants for UST and EPT on a tree in document classification with graph $\G_{\text{Sqrt}}$ with $M=10^3$.}
  \label{fg:DOC_1KSqrt_SLICE_app}
 \vspace{-6pt}
\end{figure}

 \begin{figure}[h]
  \vspace{-2pt}
  \begin{center}
    \includegraphics[width=0.5\textwidth]{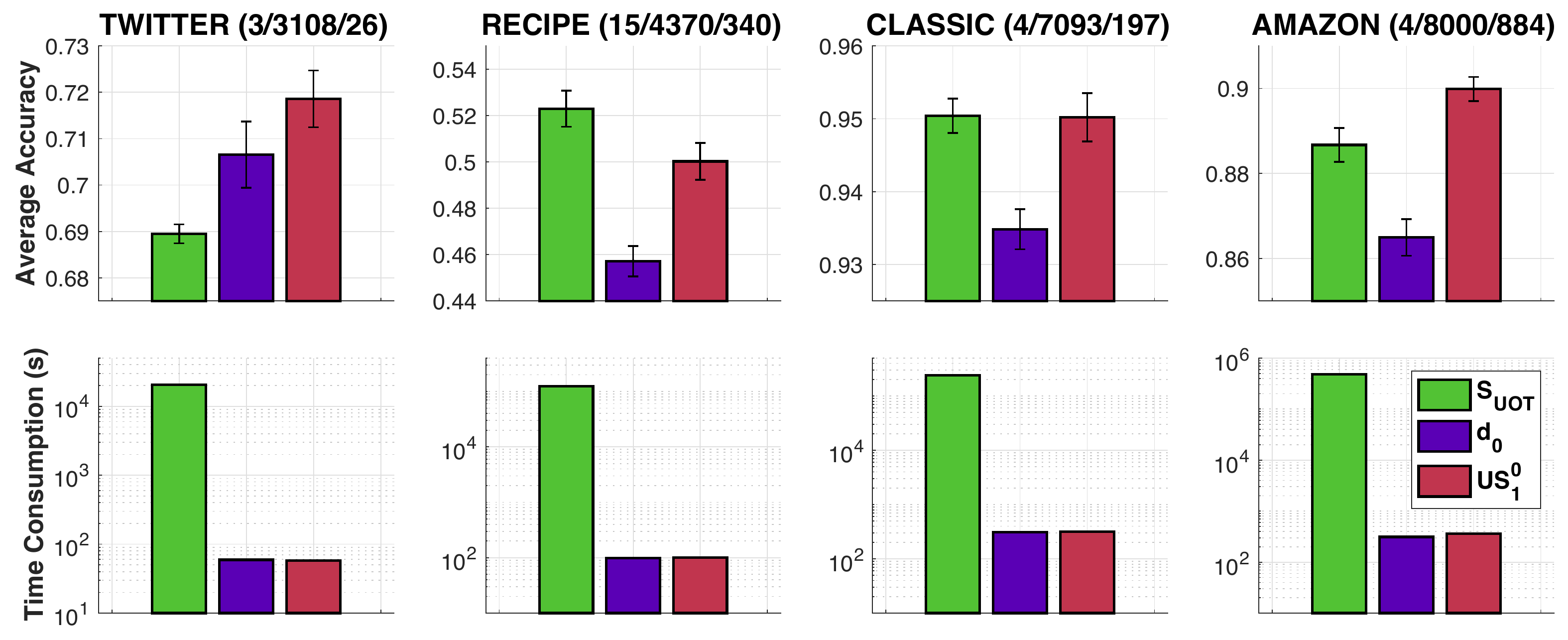}
  \end{center}
  \vspace{-14pt}
  \caption{SVM results and time consumption for kernel matrices in document classification with graph $\G_{\text{Log}}$ with $M=10^3$.}
  \label{fg:DOC_1KLog_app}
 \vspace{-6pt}
\end{figure}

\begin{figure}[h]
  \vspace{-2pt}
  \begin{center}
    \includegraphics[width=0.5\textwidth]{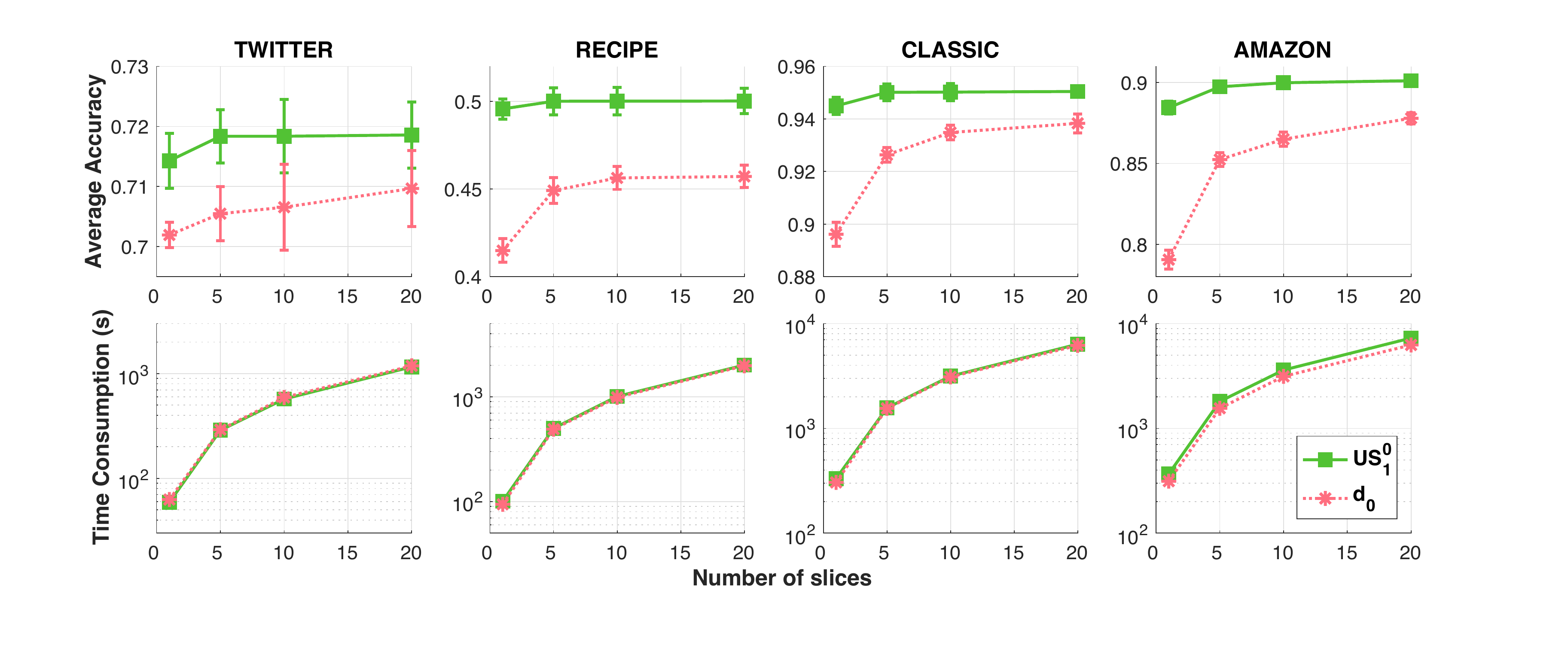}
  \end{center}
  \vspace{-14pt}
  \caption{SVM results and time consumption for kernel matrices of slice variants for UST and EPT on a tree in document classification with graph $\G_{\text{Log}}$ with $M=10^3$.}
  \label{fg:DOC_1KLog_SLICE_app}
 \vspace{-6pt}
\end{figure}
    
    \item For $M=10^4$, we illustrate the SVM results and time consumption for kernels matrices and the effect of the number of slices for graph $\G_{\text{Sqrt}}$ in Figure~\ref{fg:DOC_10KSqrt_app} and Figure~\ref{fg:DOC_10KSqrt_SLICE_app} respectively. The corresponding results for graph $\G_{\text{Log}}$ are in Figure~\ref{fg:DOC_10KLog_app} and Figure~\ref{fg:DOC_10KLog_SLICE_app}.
 
 \begin{figure}[h]
  \vspace{-2pt}
  \begin{center}
    \includegraphics[width=0.5\textwidth]{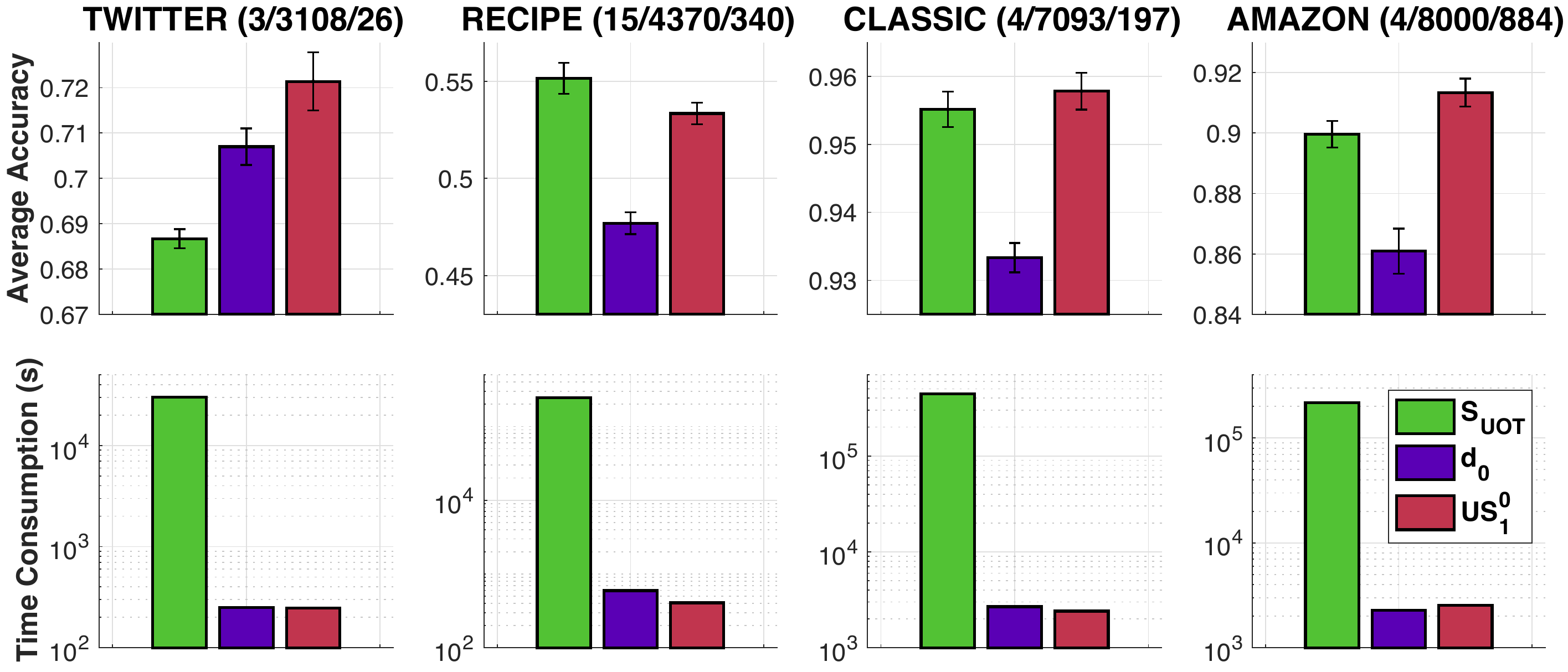}
  \end{center}
  \vspace{-14pt}
  \caption{SVM results and time consumption for kernel matrices in document classification with graph $\G_{\text{Sqrt}}$ with $M=10^4$.}
  \label{fg:DOC_10KSqrt_app}
 \vspace{-6pt}
\end{figure}

\begin{figure}[h]
  \vspace{-2pt}
  \begin{center}
    \includegraphics[width=0.5\textwidth]{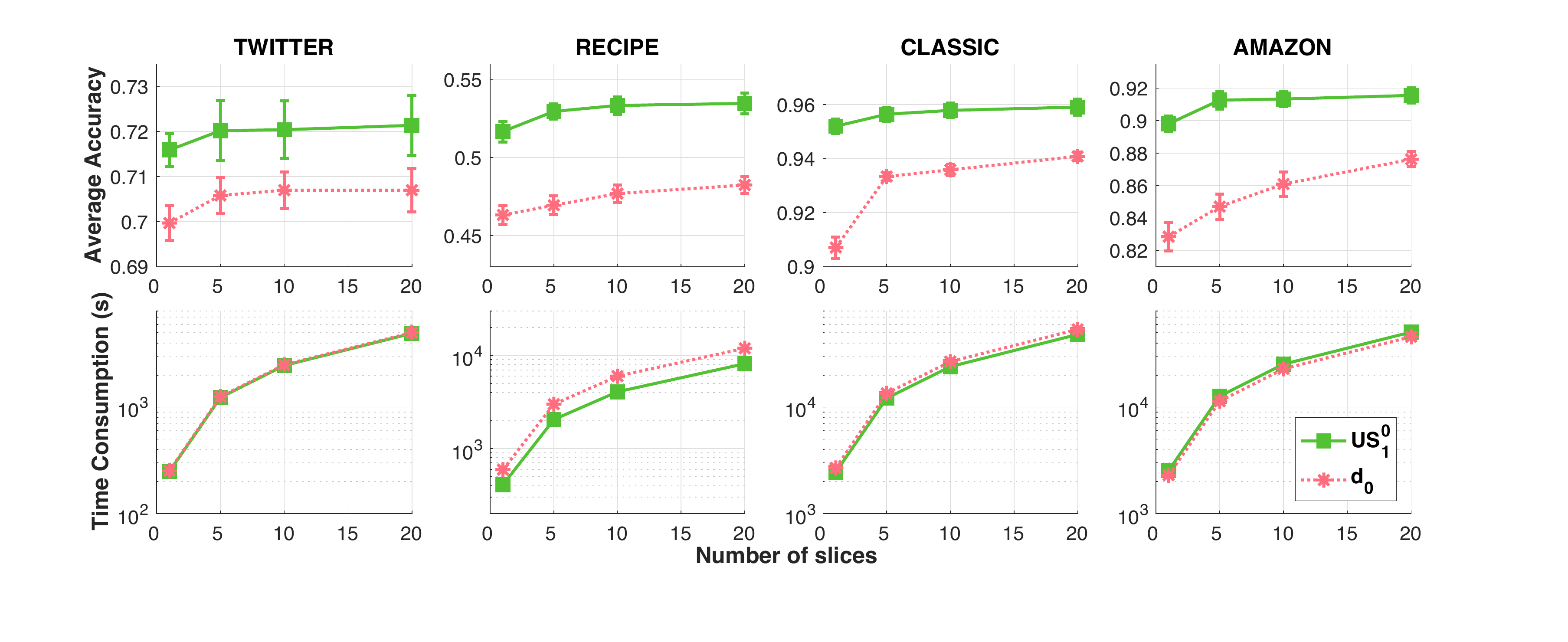}
  \end{center}
  \vspace{-14pt}
  \caption{SVM results and time consumption for kernel matrices of slice variants for UST and EPT on a tree in document classification with graph $\G_{\text{Sqrt}}$ with $M=10^4$.}
  \label{fg:DOC_10KSqrt_SLICE_app}
 \vspace{-6pt}
\end{figure}

 \begin{figure}[h]
  \vspace{-2pt}
  \begin{center}
    \includegraphics[width=0.5\textwidth]{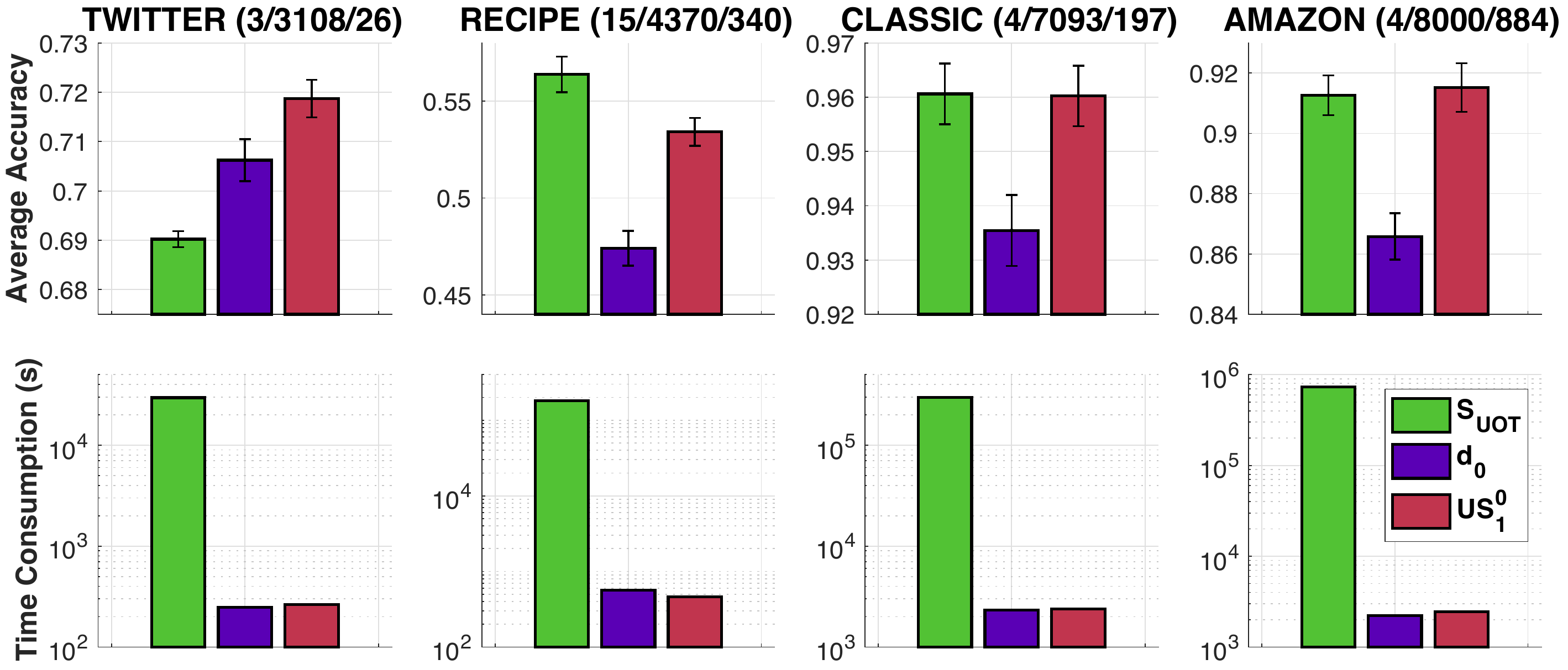}
  \end{center}
  \vspace{-14pt}
  \caption{SVM results and time consumption for kernel matrices in document classification with graph $\G_{\text{Log}}$ with $M=10^4$.}
  \label{fg:DOC_10KLog_app}
 \vspace{-6pt}
\end{figure}

\begin{figure}[h]
  \vspace{-2pt}
  \begin{center}
    \includegraphics[width=0.5\textwidth]{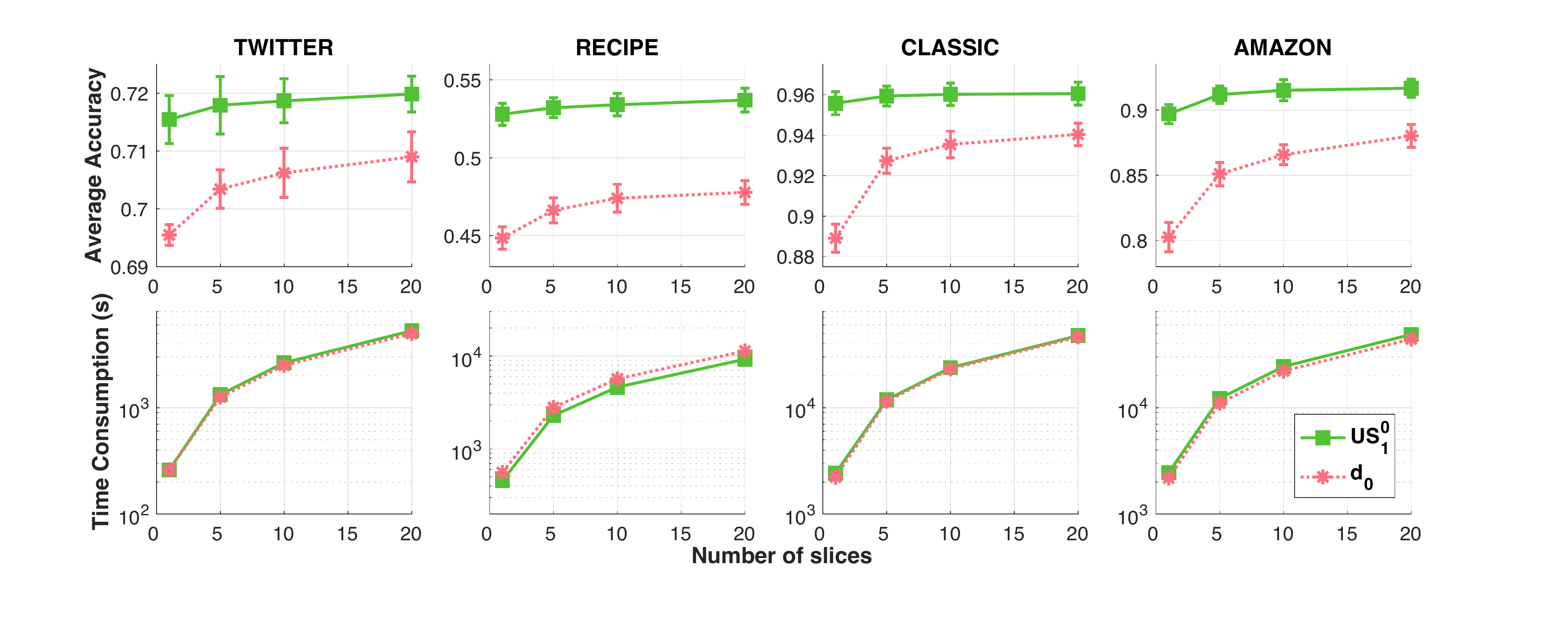}
  \end{center}
  \vspace{-14pt}
  \caption{SVM results and time consumption for kernel matrices of slice variants for UST and EPT on a tree in document classification with graph $\G_{\text{Log}}$ with $M=10^4$.}
  \label{fg:DOC_10KLog_SLICE_app}
 \vspace{-6pt}
\end{figure}


    \item For $M=4 \times 10^4$, we illustrate the SVM results and time consumption for kernels matrices and the effect of the number of slices for graph $\G_{\text{Sqrt}}$ in Figure~\ref{fg:DOC_40KSqrt_app} and Figure~\ref{fg:DOC_40KSqrt_SLICE_app} respectively. The corresponding results for graph $\G_{\text{Log}}$ are in Figure~\ref{fg:DOC_40KLog_app} and Figure~\ref{fg:DOC_40KLog_SLICE_app}.
 
 \begin{figure}[h]
  \vspace{-2pt}
  \begin{center}
    \includegraphics[width=0.3\textwidth]{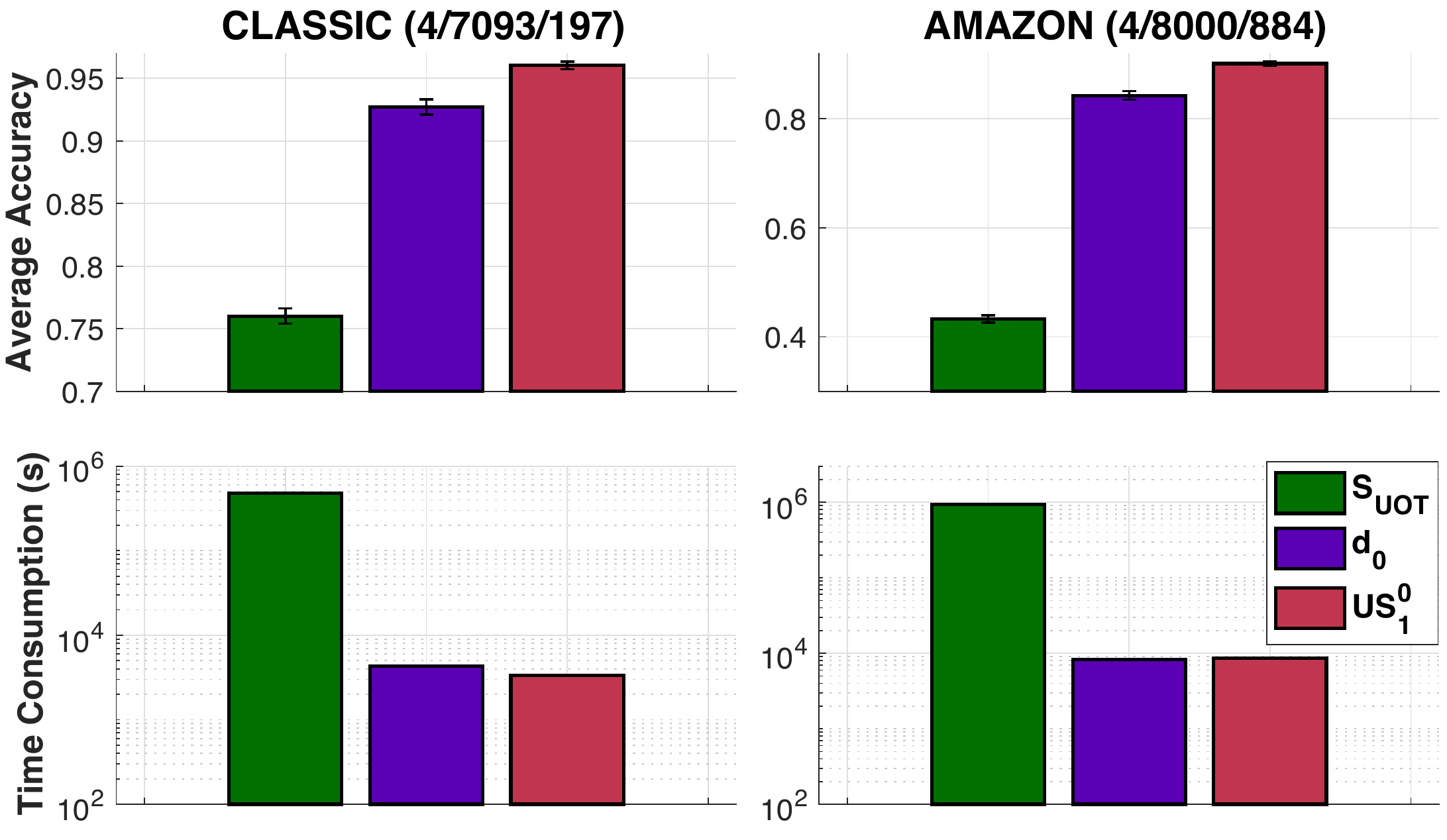}
  \end{center}
  \vspace{-14pt}
  \caption{SVM results and time consumption for kernel matrices in document classification with graph $\G_{\text{Sqrt}}$ with $M=4 \times 10^4$.}
  \label{fg:DOC_40KSqrt_app}
 \vspace{-6pt}
\end{figure}

\begin{figure}[h]
  \vspace{-2pt}
  \begin{center}
    \includegraphics[width=0.3\textwidth]{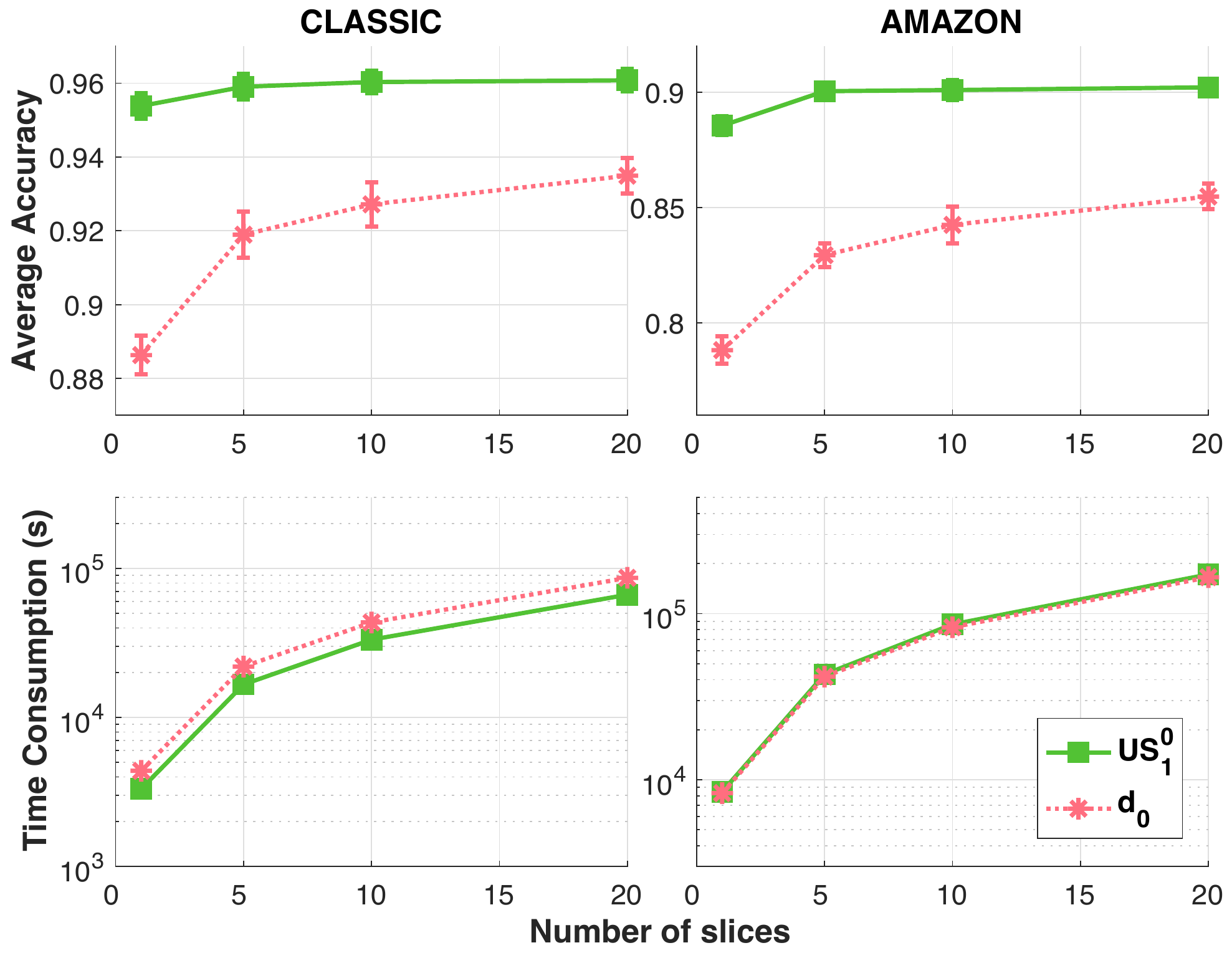}
  \end{center}
  \vspace{-14pt}
  \caption{SVM results and time consumption for kernel matrices of slice variants for UST and EPT on a tree in document classification with graph $\G_{\text{Sqrt}}$ with $M=4 \times 10^4$.}
  \label{fg:DOC_40KSqrt_SLICE_app}
 \vspace{-6pt}
\end{figure}

 \begin{figure}[h]
  \vspace{-2pt}
  \begin{center}
    \includegraphics[width=0.3\textwidth]{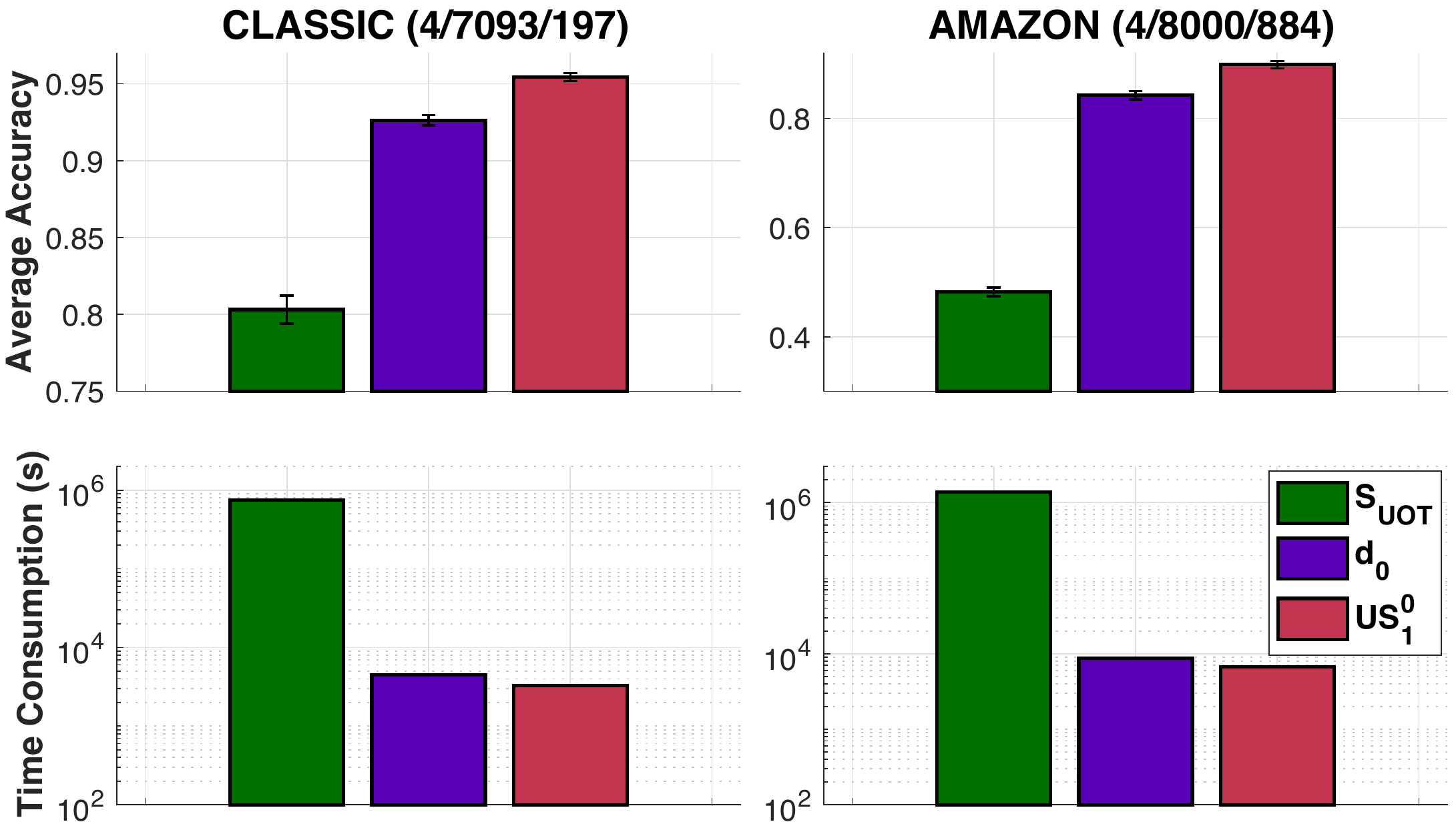}
  \end{center}
  \vspace{-14pt}
  \caption{SVM results and time consumption for kernel matrices in document classification with graph $\G_{\text{Log}}$ with $M=4 \times 10^4$.}
  \label{fg:DOC_40KLog_app}
 \vspace{-6pt}
\end{figure}

\begin{figure}[h]
  \vspace{-2pt}
  \begin{center}
    \includegraphics[width=0.3\textwidth]{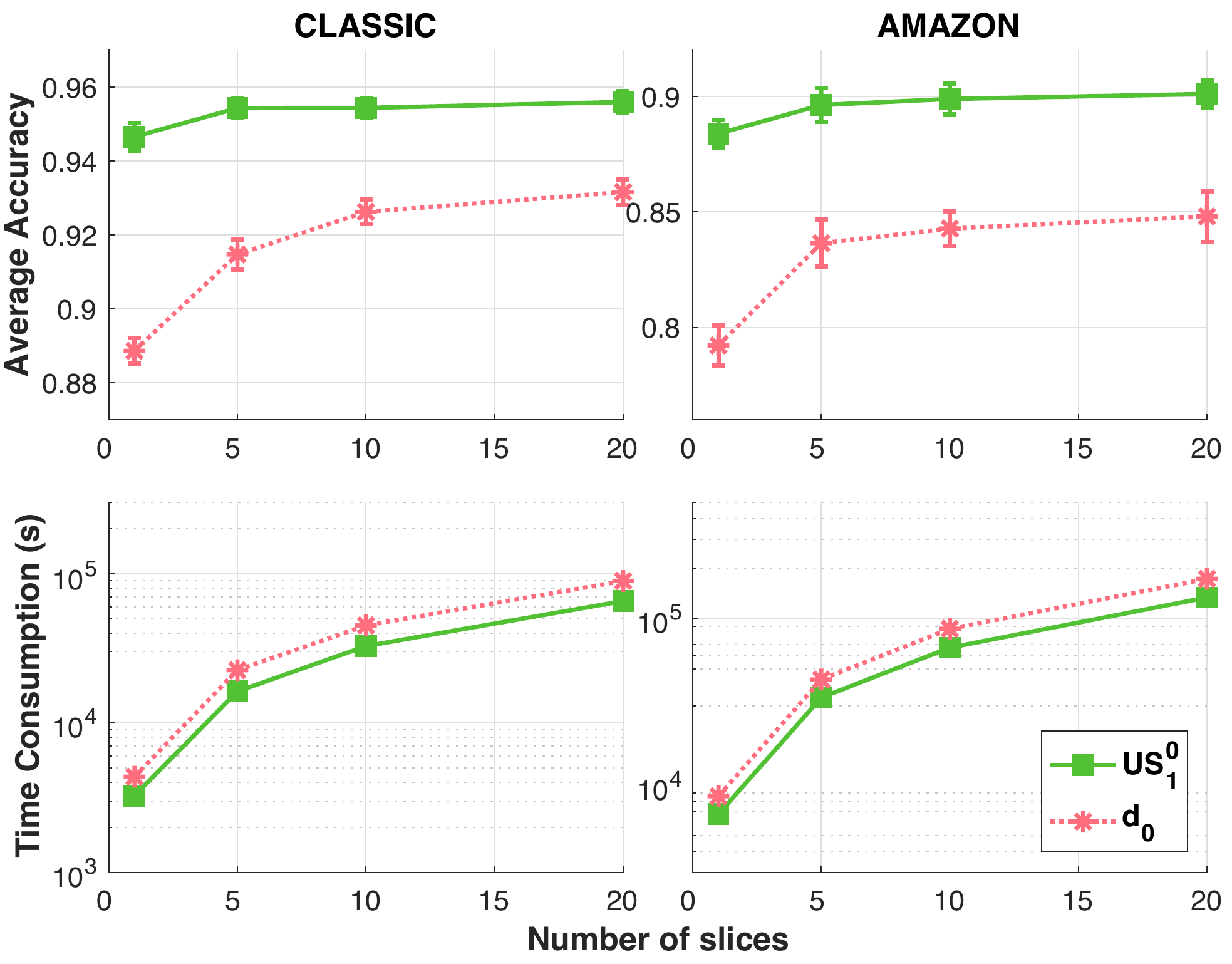}
  \end{center}
  \vspace{-14pt}
  \caption{SVM results and time consumption for kernel matrices of slice variants for UST and EPT on a tree in document classification with graph $\G_{\text{Log}}$ with $M=4 \times 10^4$.}
  \label{fg:DOC_40KLog_SLICE_app}
 \vspace{-6pt}
\end{figure}

\end{itemize}

\paragraph{For TDA.}
\begin{itemize}
    \item For $M=10^2$, we illustrate the SVM results and time consumption for kernels matrices and the effect of the number of slices for graph $\G_{\text{Sqrt}}$ in Figure~\ref{fg:TDA_100Sqrt_app} and Figure~\ref{fg:TDA_100Sqrt_SLICE_app} respectively. The corresponding results for graph $\G_{\text{Log}}$ are in Figure~\ref{fg:TDA_100Log_app} and Figure~\ref{fg:TDA_100Log_SLICE_app}.
 
 \begin{figure}[h]
  \vspace{-2pt}
  \begin{center}
    \includegraphics[width=0.3\textwidth]{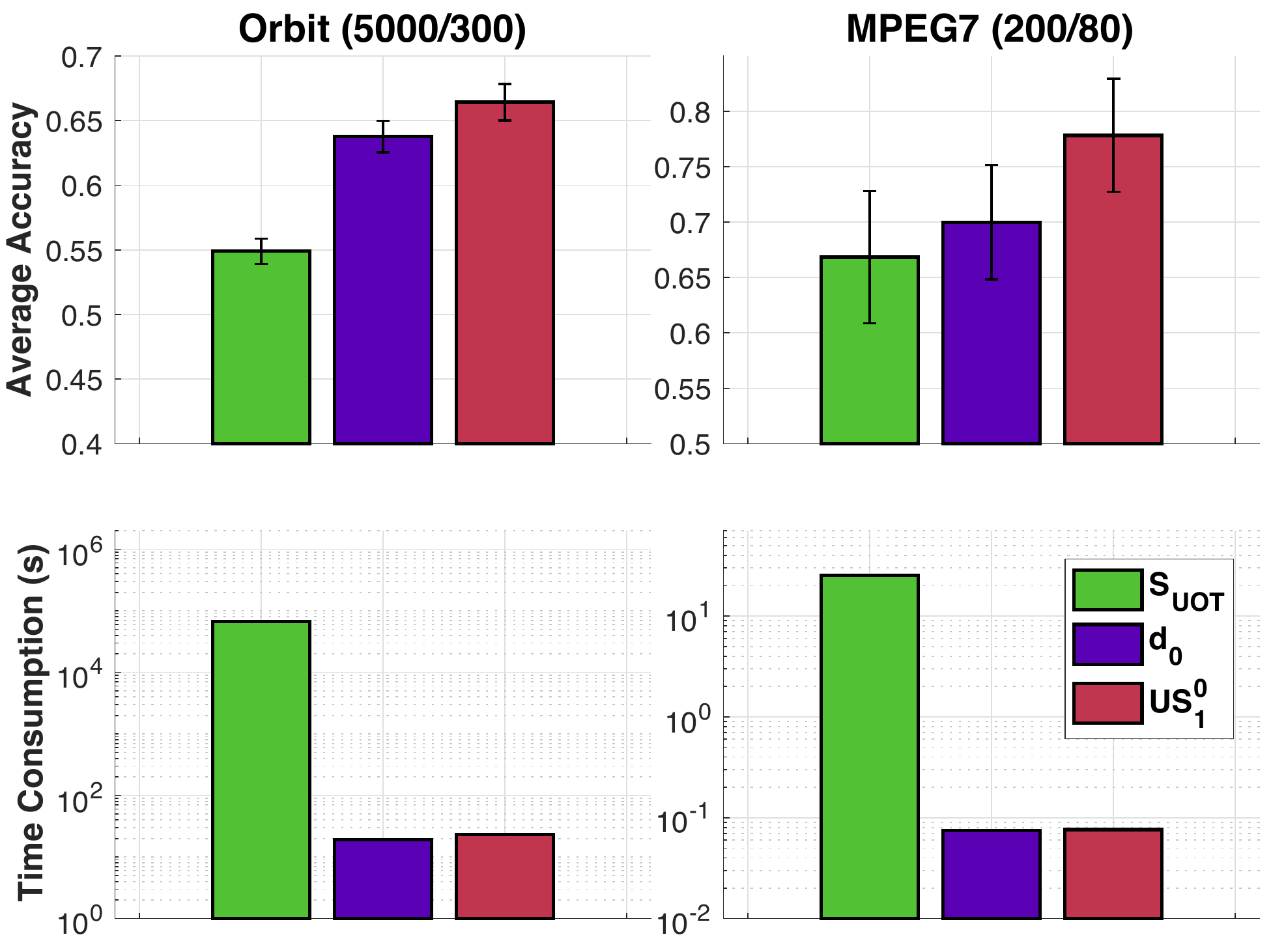}
  \end{center}
  \vspace{-14pt}
  \caption{SVM results and time consumption for kernel matrices in TDA with graph $\G_{\text{Sqrt}}$ with $M=10^2$.}
  \label{fg:TDA_100Sqrt_app}
 \vspace{-6pt}
\end{figure}

\begin{figure}[h]
 \vspace{-2pt}
  \begin{center}
    \includegraphics[width=0.3\textwidth]{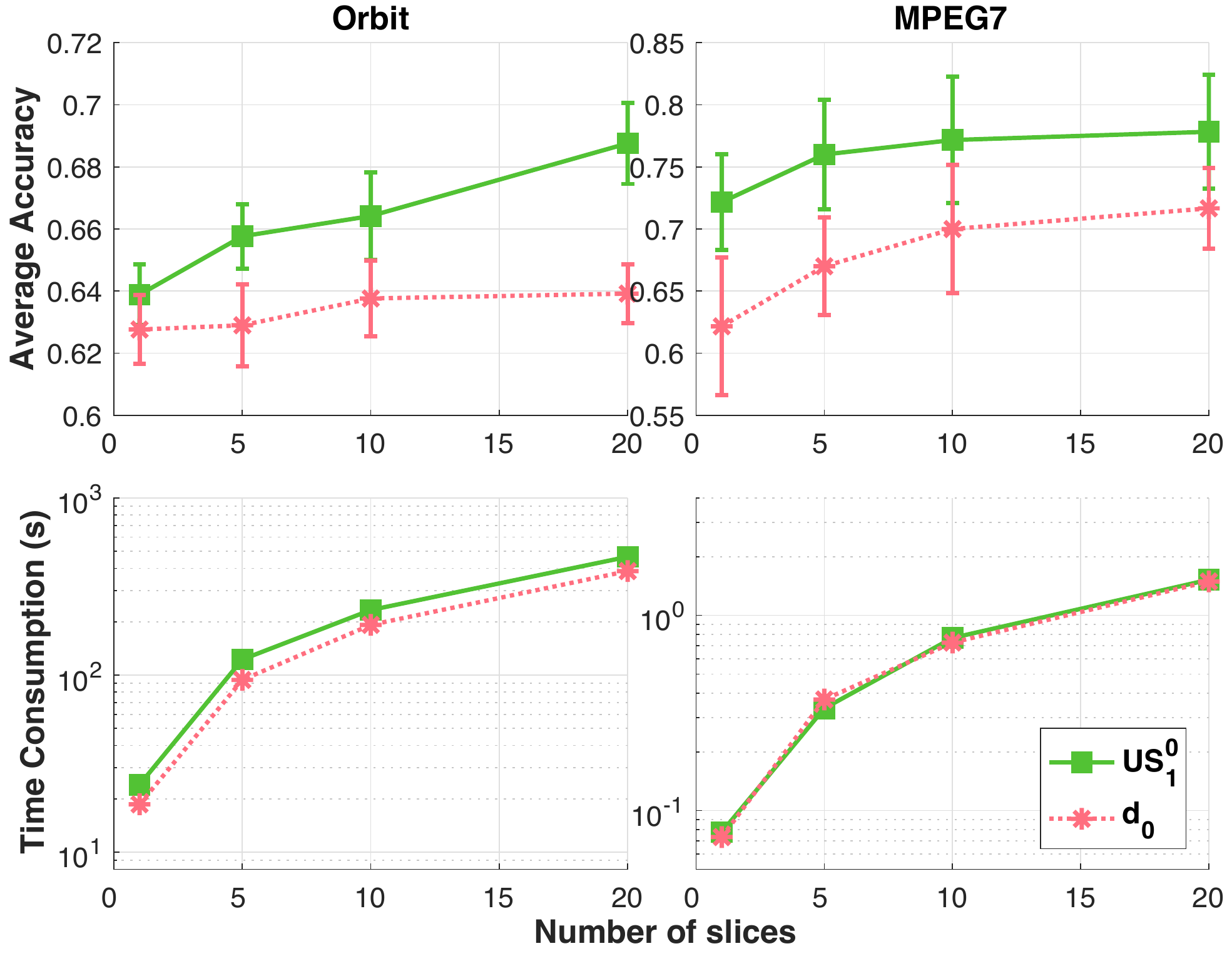}
  \end{center}
  \vspace{-14pt}
  \caption{SVM results and time consumption for kernel matrices of slice variants for UST and EPT on a tree in TDA with graph $\G_{\text{Sqrt}}$ with $M=10^2$.}
  \label{fg:TDA_100Sqrt_SLICE_app}
 \vspace{-6pt}
\end{figure}

 \begin{figure}[h]
  \vspace{-2pt}
  \begin{center}
    \includegraphics[width=0.3\textwidth]{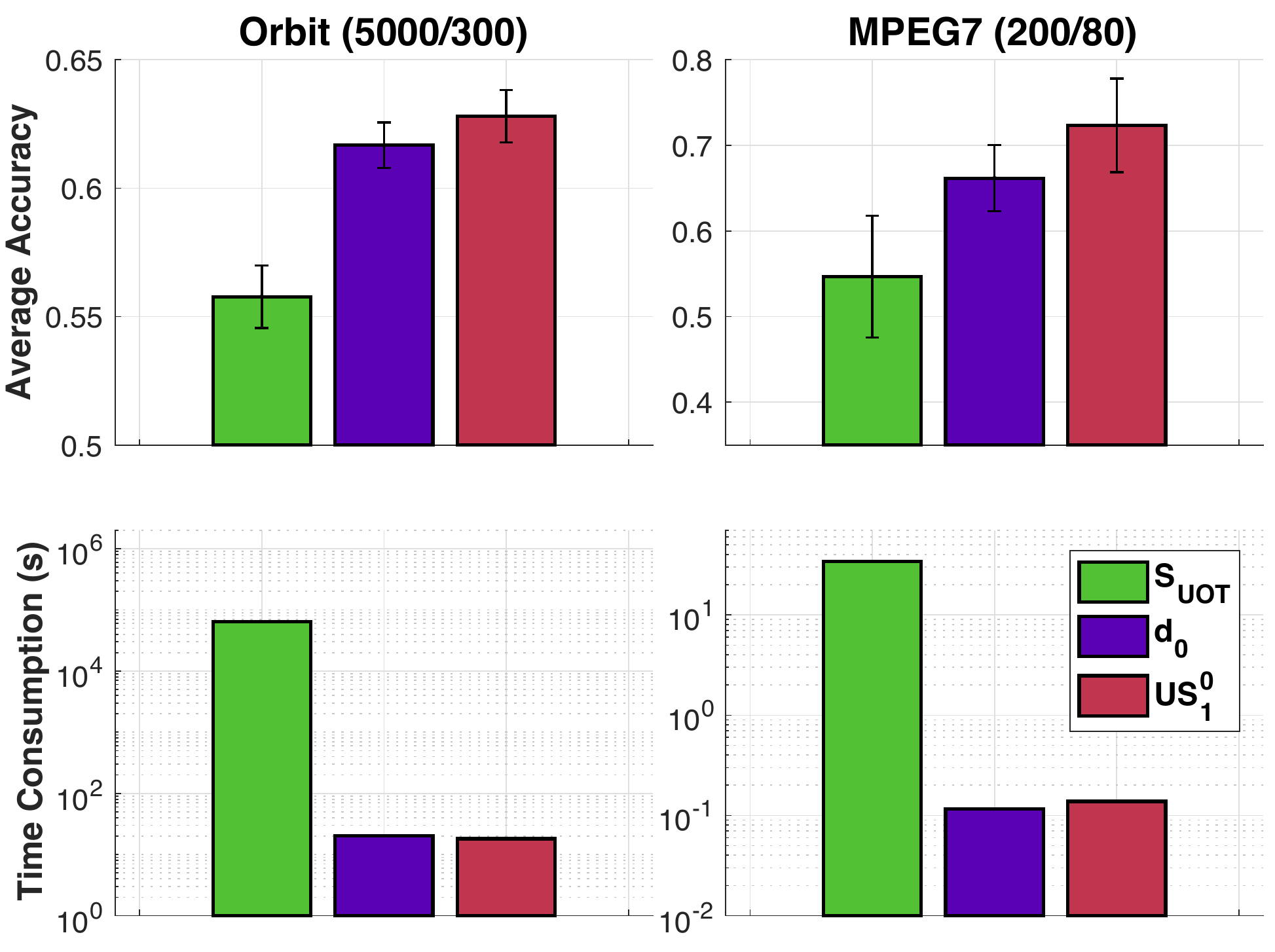}
  \end{center}
  \vspace{-14pt}
  \caption{SVM results and time consumption for kernel matrices in TDA with graph $\G_{\text{Log}}$ with $M=10^2$.}
  \label{fg:TDA_100Log_app}
 \vspace{-6pt}
\end{figure}

\begin{figure}[h]
  \vspace{-2pt}
  \begin{center}
    \includegraphics[width=0.3\textwidth]{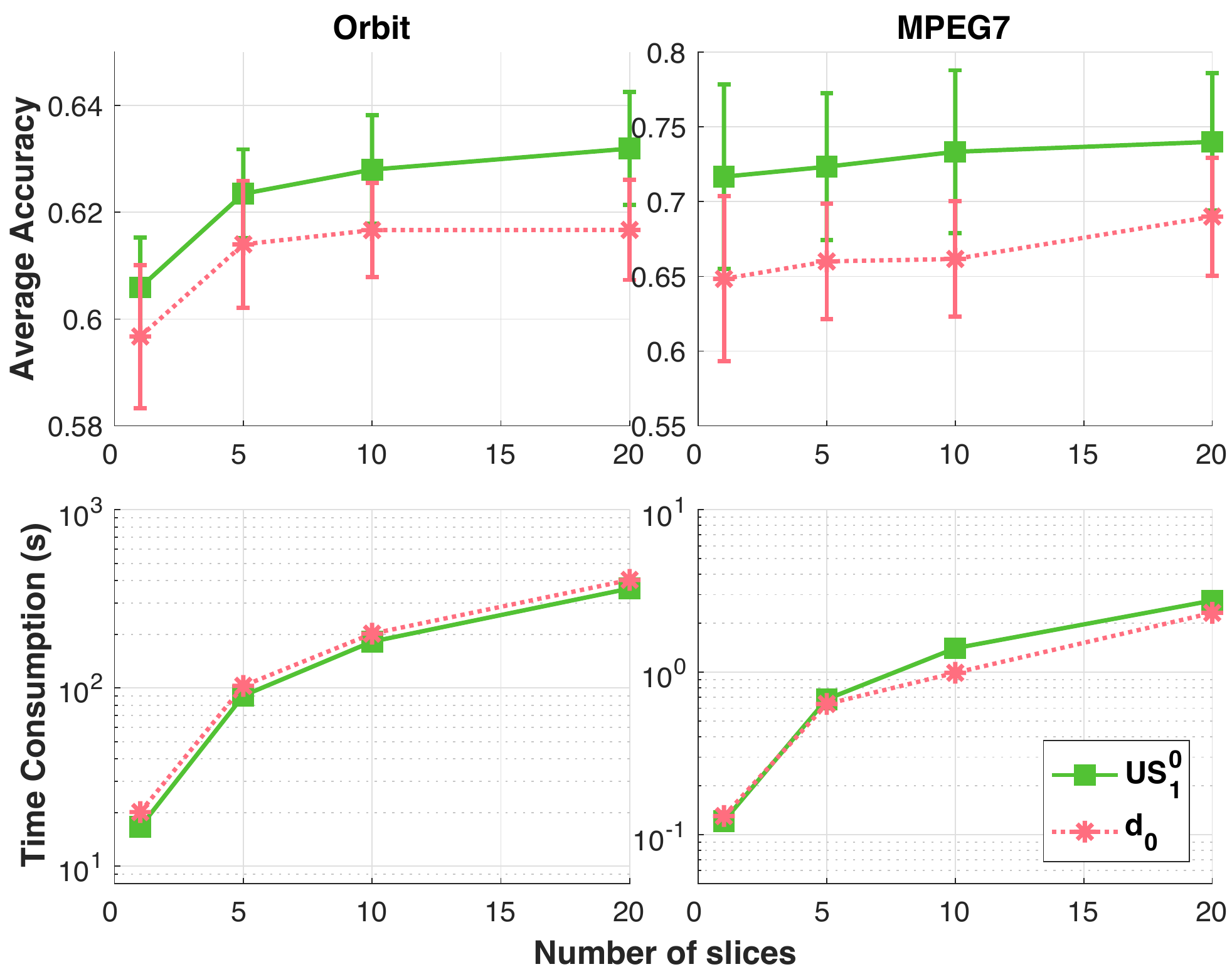}
  \end{center}
  \vspace{-14pt}
  \caption{SVM results and time consumption for kernel matrices of slice variants for UST and EPT on a tree in TDA with graph $\G_{\text{Log}}$ with $M=10^2$.}
  \label{fg:TDA_100Log_SLICE_app}
 \vspace{-6pt}
\end{figure}


    \item For $M=10^3$, we illustrate the SVM results and time consumption for kernels matrices and the effect of the number of slices for graph $\G_{\text{Sqrt}}$ in Figure~\ref{fg:TDA_1KSqrt_app} and Figure~\ref{fg:TDA_1KSqrt_SLICE_app} respectively. The corresponding results for graph $\G_{\text{Log}}$ are in Figure~\ref{fg:TDA_1KLog_app} and Figure~\ref{fg:TDA_1KLog_SLICE_app}.
 
 \begin{figure}[h]
  \vspace{-2pt}
  \begin{center}
    \includegraphics[width=0.27\textwidth]{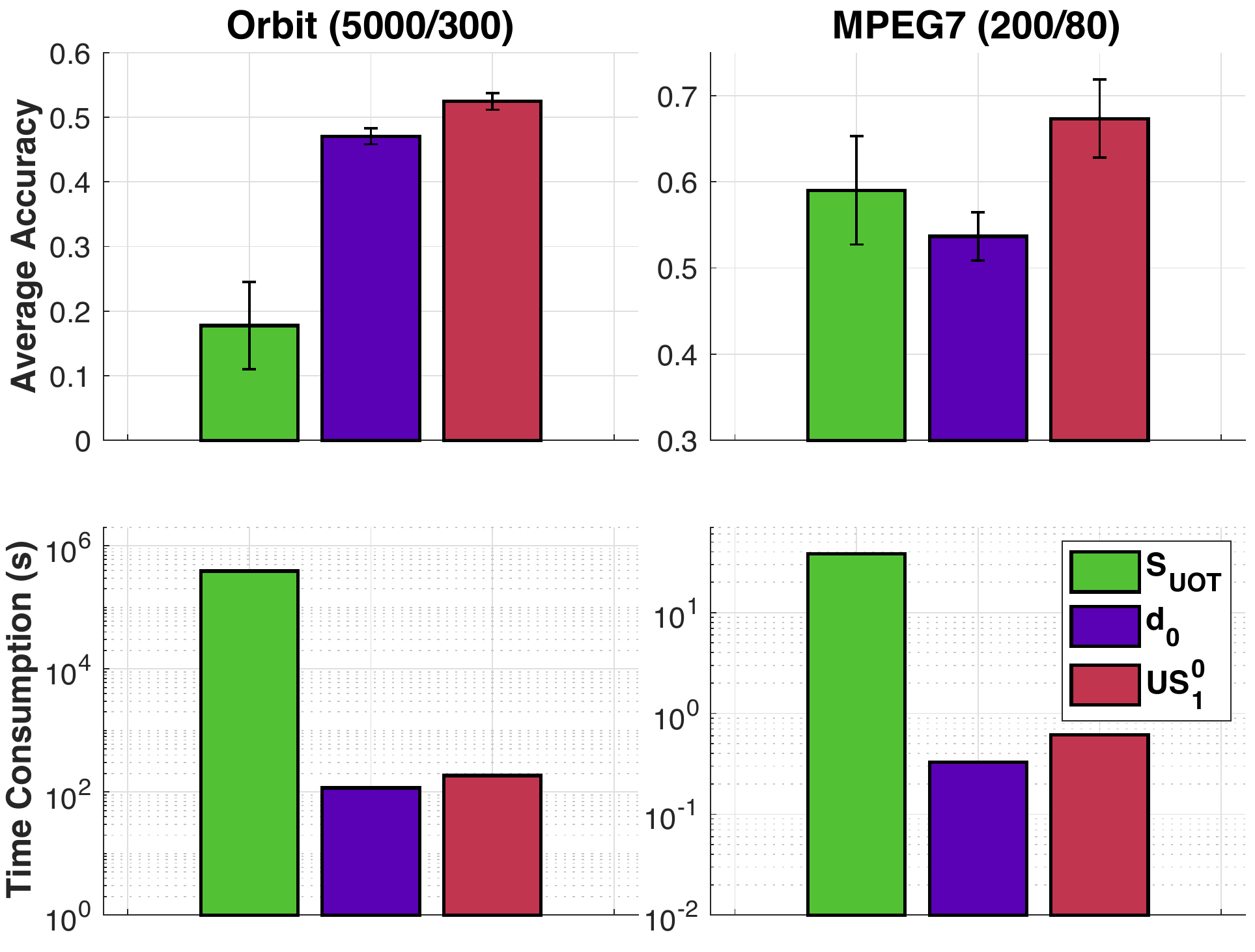}
  \end{center}
  \vspace{-14pt}
  \caption{SVM results and time consumption for kernel matrices in TDA with graph $\G_{\text{Sqrt}}$ with $M=10^3$.}
  \label{fg:TDA_1KSqrt_app}
 \vspace{-6pt}
\end{figure}

\begin{figure}[h]
  \vspace{-2pt}
  \begin{center}
    \includegraphics[width=0.3\textwidth]{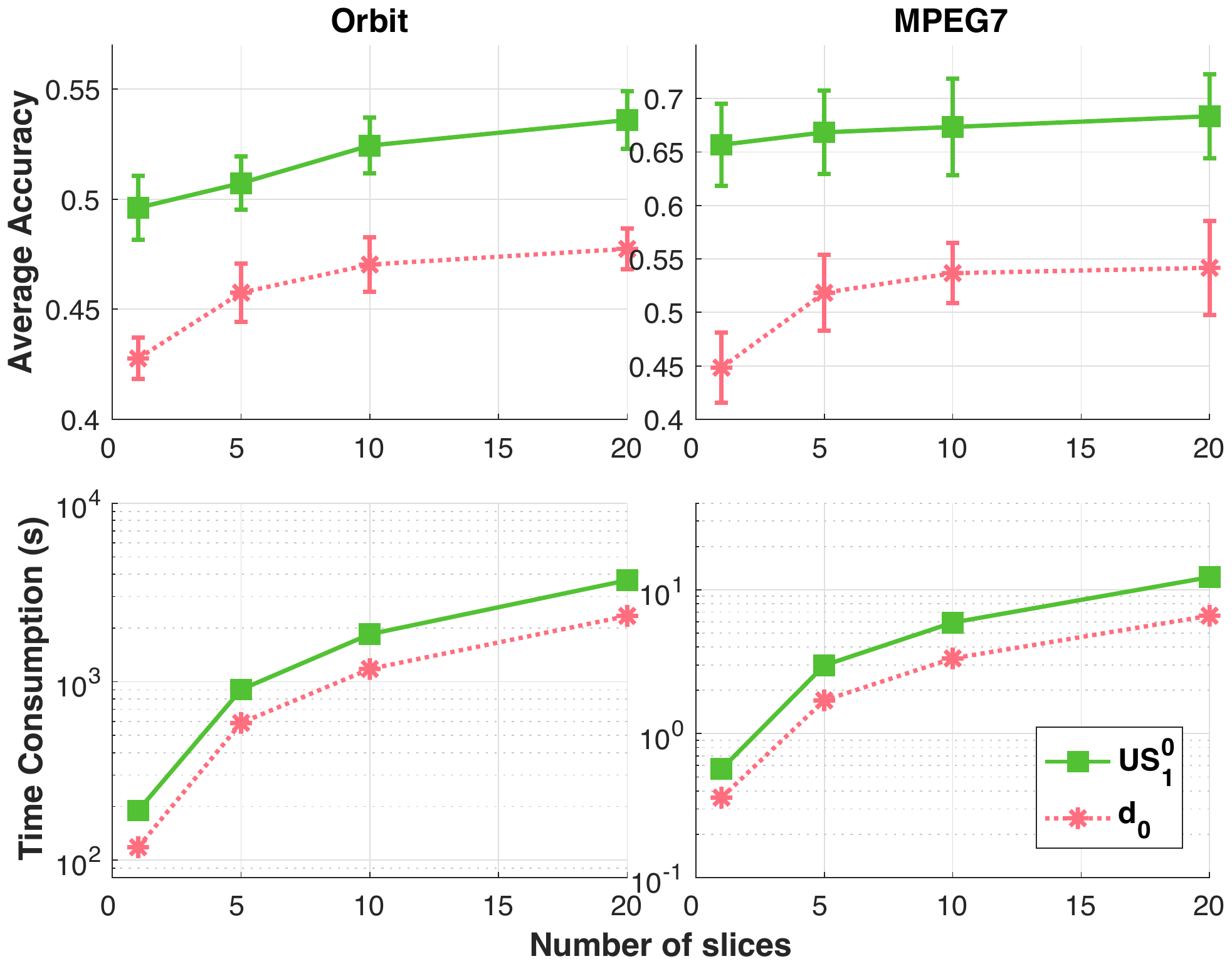}
  \end{center}
  \vspace{-14pt}
  \caption{SVM results and time consumption for kernel matrices of slice variants for UST and EPT on a tree in TDA with graph $\G_{\text{Sqrt}}$ with $M=10^3$.}
  \label{fg:TDA_1KSqrt_SLICE_app}
 \vspace{-6pt}
\end{figure}

 \begin{figure}[h]
  \vspace{-2pt}
  \begin{center}
    \includegraphics[width=0.27\textwidth]{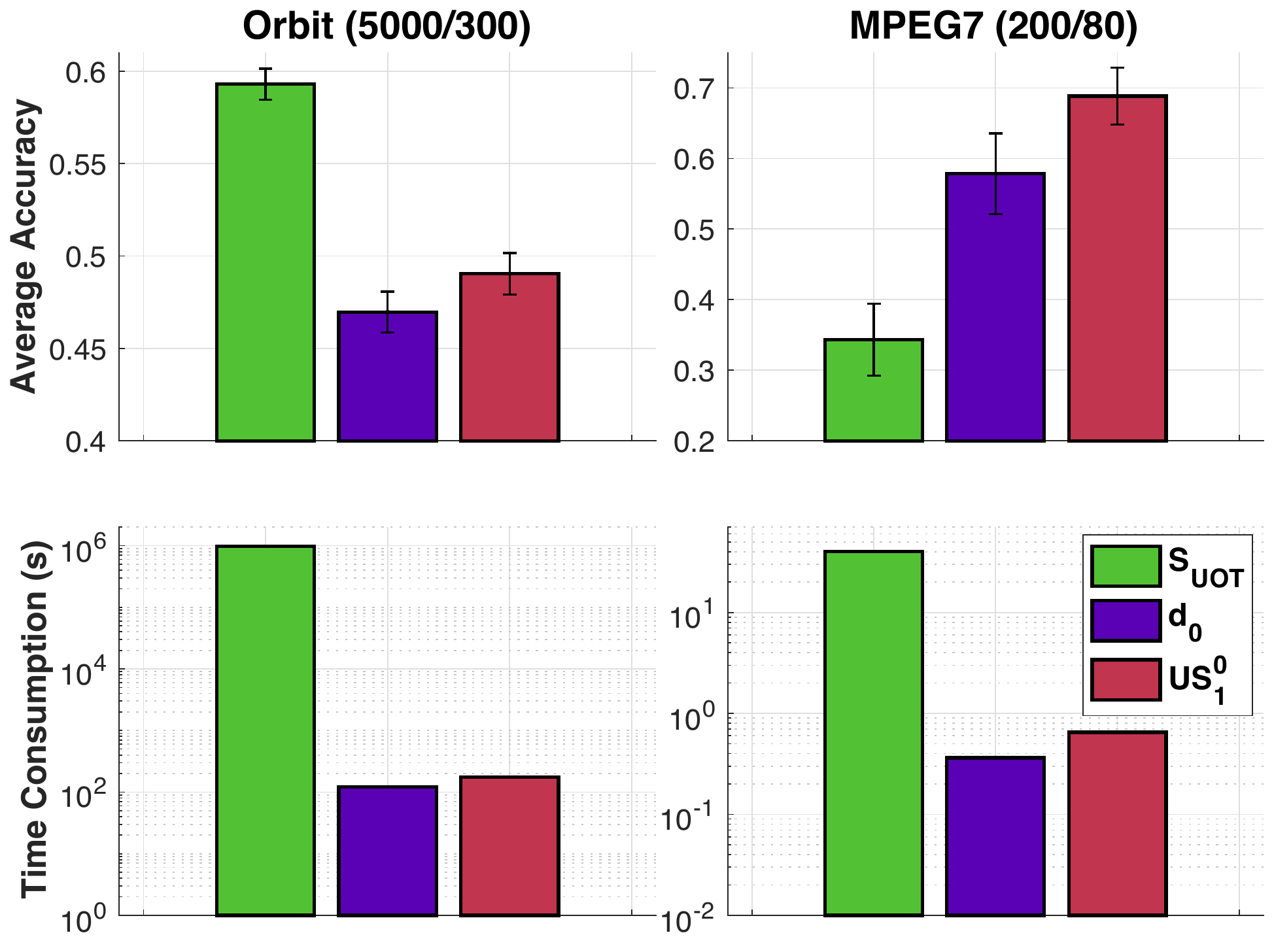}
  \end{center}
  \vspace{-14pt}
  \caption{SVM results and time consumption for kernel matrices in TDA with graph $\G_{\text{Log}}$ with $M=10^3$.}
  \label{fg:TDA_1KLog_app}
 \vspace{-6pt}
\end{figure}

\begin{figure}[h]
  \vspace{-2pt}
  \begin{center}
    \includegraphics[width=0.3\textwidth]{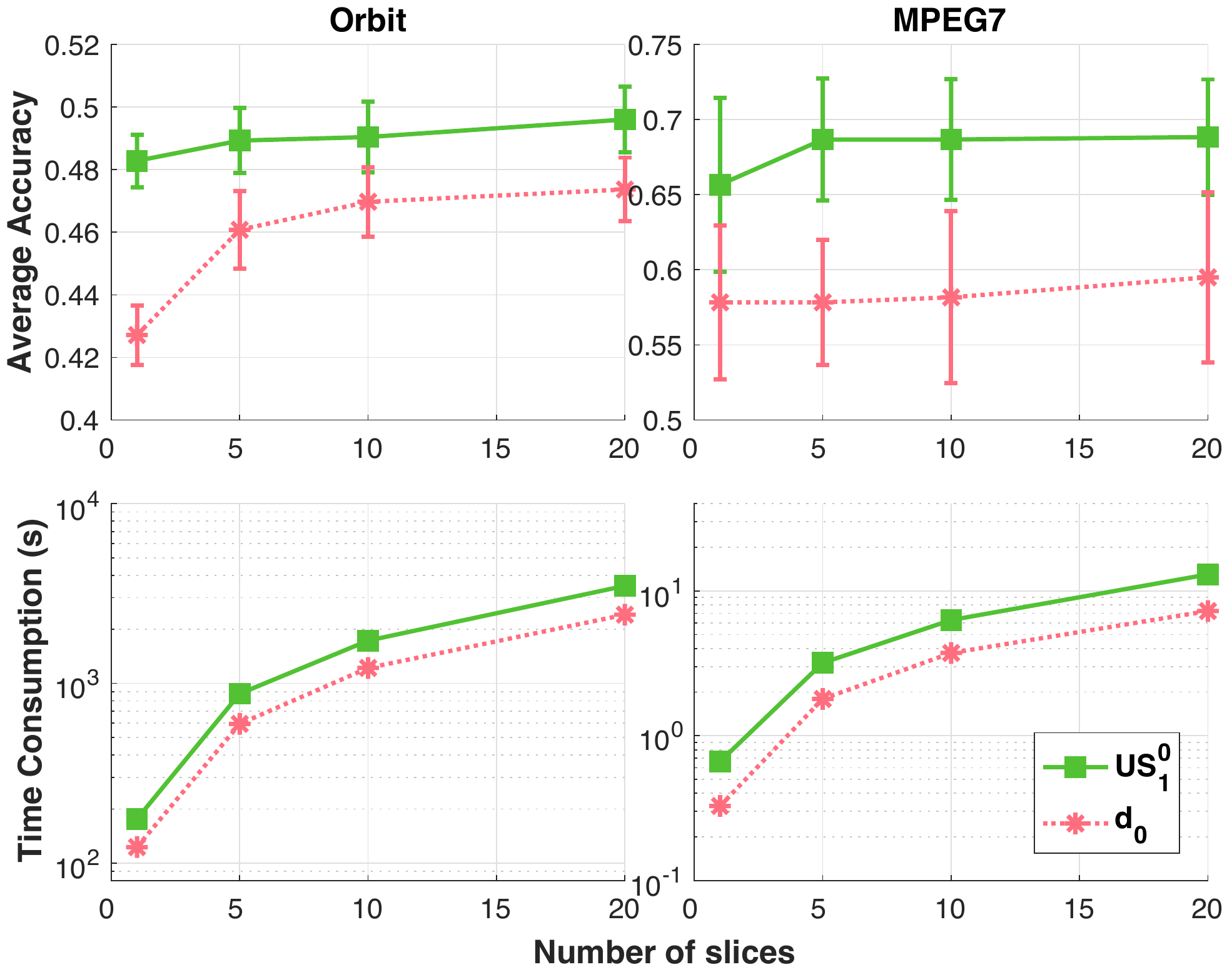}
  \end{center}
  \vspace{-14pt}
  \caption{SVM results and time consumption for kernel matrices of slice variants for UST and EPT on a tree in TDA with graph $\G_{\text{Log}}$ with $M=10^3$.}
  \label{fg:TDA_1KLog_SLICE_app}
 \vspace{-6pt}
\end{figure}

    \item For $M=10^4$ on \texttt{Orbit} dataset and $M=10^3$ on \texttt{MPEG7} dataset (due to the same size of \texttt{MPEG7} dataset), we illustrate the SVM results and time consumption for kernels matrices and the effect of the number of slices for graph $\G_{\text{Sqrt}}$ in Figure~\ref{fg:TDA_Mix1K10KSqrt_app} and Figure~\ref{fg:TDA_Mix1K10KSqrt_SLICE_app} respectively. The corresponding results for graph $\G_{\text{Log}}$ are in Figure~\ref{fg:TDA_Mix1K10KLog_app} and Figure~\ref{fg:TDA_Mix1K10KLog_SLICE_app}.
 
 \begin{figure}[h]
  \vspace{-2pt}
  \begin{center}
    \includegraphics[width=0.3\textwidth]{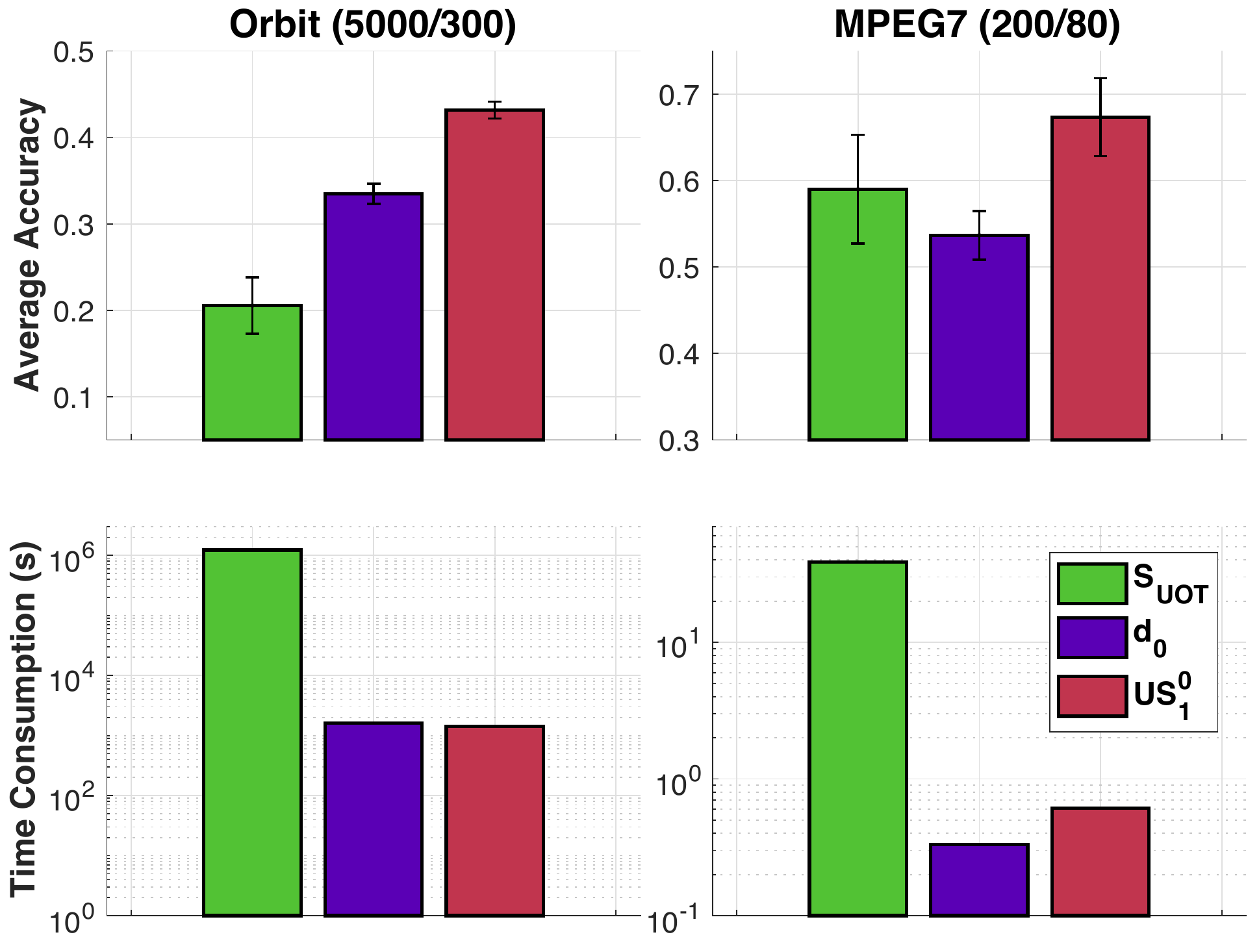}
  \end{center}
  \vspace{-14pt}
  \caption{SVM results and time consumption for kernel matrices in TDA with graph $\G_{\text{Sqrt}}$ with $M=10^4$ for \texttt{Orbit} and with $M=10^3$ for \texttt{MPEG7}.}
  \label{fg:TDA_Mix1K10KSqrt_app}
 \vspace{-6pt}
\end{figure}

\begin{figure}[h]
  \vspace{-2pt}
  \begin{center}
    \includegraphics[width=0.3\textwidth]{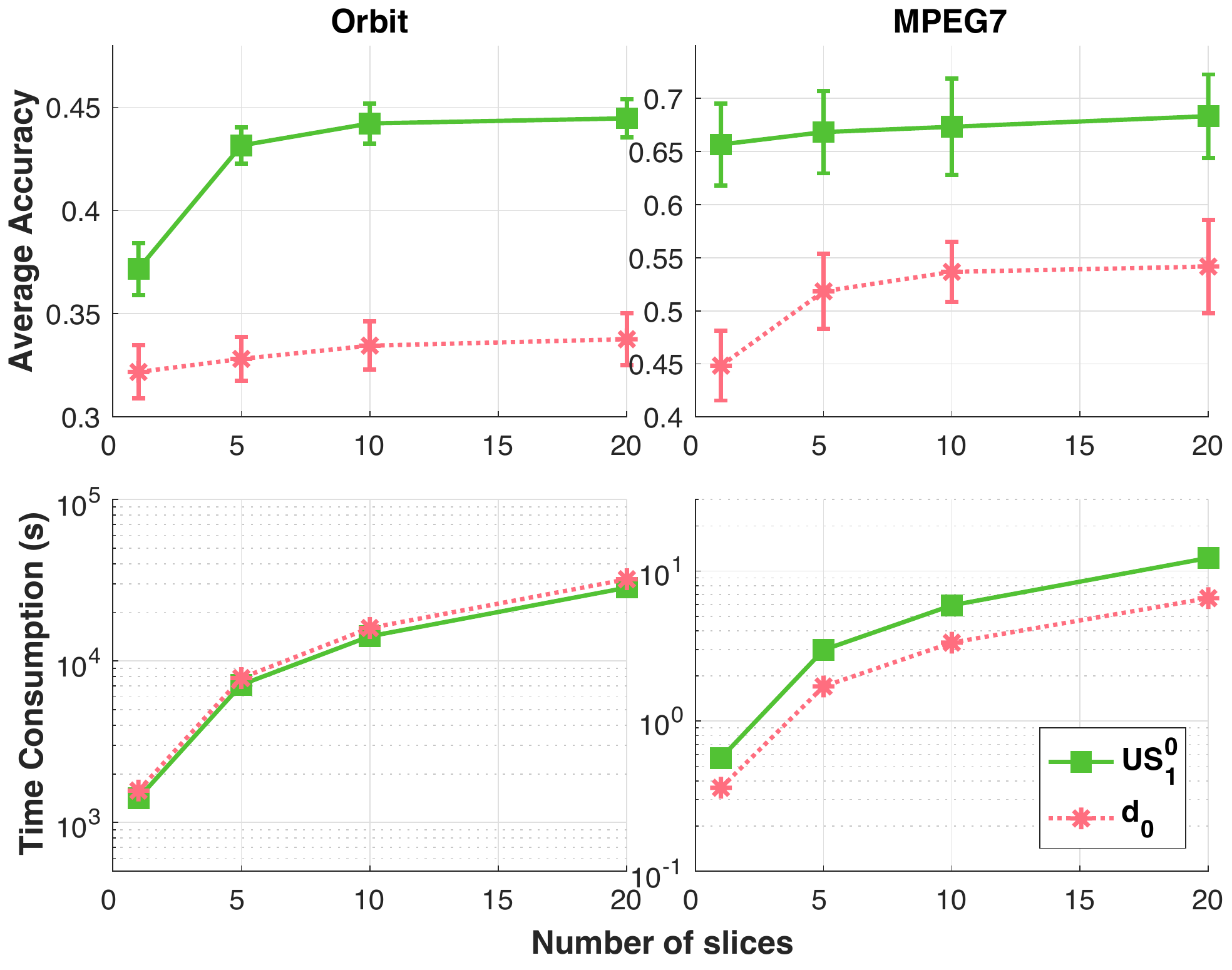}
  \end{center}
  \vspace{-14pt}
  \caption{SVM results and time consumption for kernel matrices of slice variants for UST and EPT on a tree in TDA with graph $\G_{\text{Sqrt}}$ with $M=10^4$ for \texttt{Orbit} and with $M=10^3$ for \texttt{MPEG7}.}
  \label{fg:TDA_Mix1K10KSqrt_SLICE_app}
 \vspace{-6pt}
\end{figure}

 \begin{figure}[h]
  \vspace{-2pt}
  \begin{center}
    \includegraphics[width=0.3\textwidth]{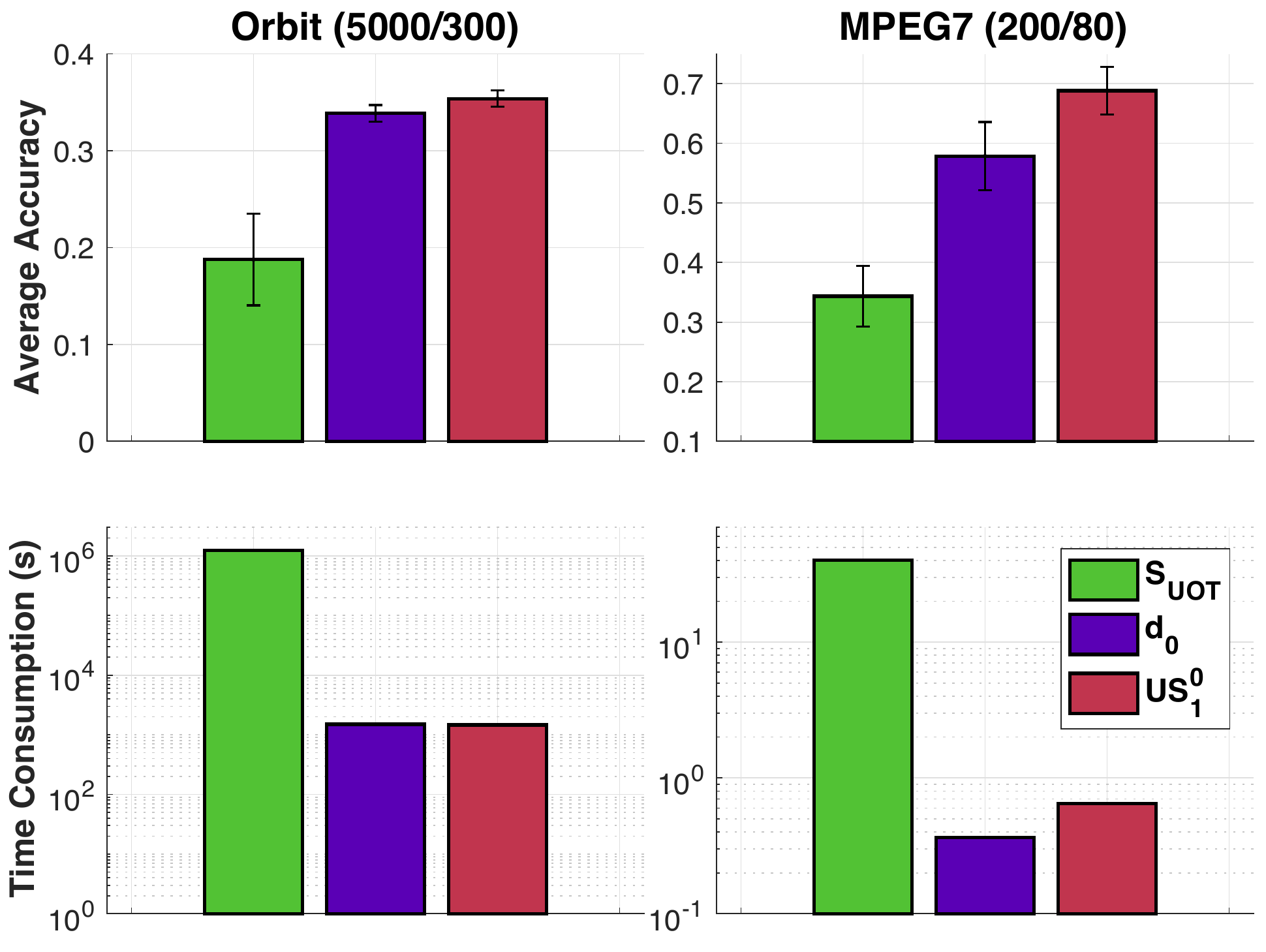}
  \end{center}
  \vspace{-14pt}
  \caption{SVM results and time consumption for kernel matrices in TDA with graph $\G_{\text{Log}}$ with $M=10^4$ for \texttt{Orbit} and with $M=10^3$ for \texttt{MPEG7}.}
  \label{fg:TDA_Mix1K10KLog_app}
 \vspace{-6pt}
\end{figure}

\begin{figure}[h]
  \vspace{-2pt}
  \begin{center}
    \includegraphics[width=0.3\textwidth]{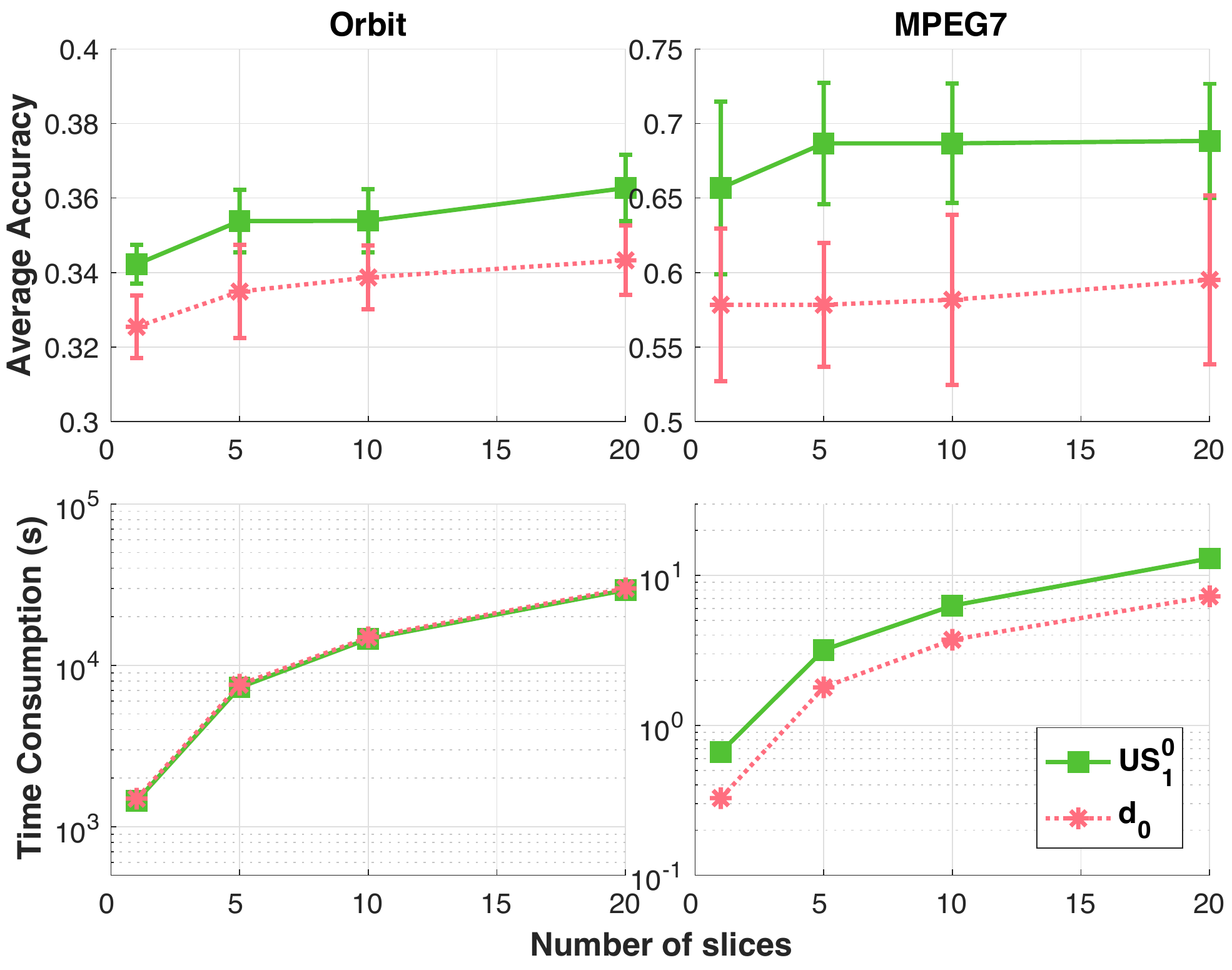}
  \end{center}
  \vspace{-14pt}
  \caption{SVM results and time consumption for kernel matrices of slice variants for UST and EPT on a tree in TDA with graph $\G_{\text{Log}}$ with $M=10^4$ for \texttt{Orbit} and with $M=10^3$ for \texttt{MPEG7}.}
  \label{fg:TDA_Mix1K10KLog_SLICE_app}
 \vspace{-6pt}
\end{figure}

\end{itemize}

\begin{figure}[h]
  \vspace{-2pt}
  \begin{center}
    \includegraphics[width=0.42\textwidth]{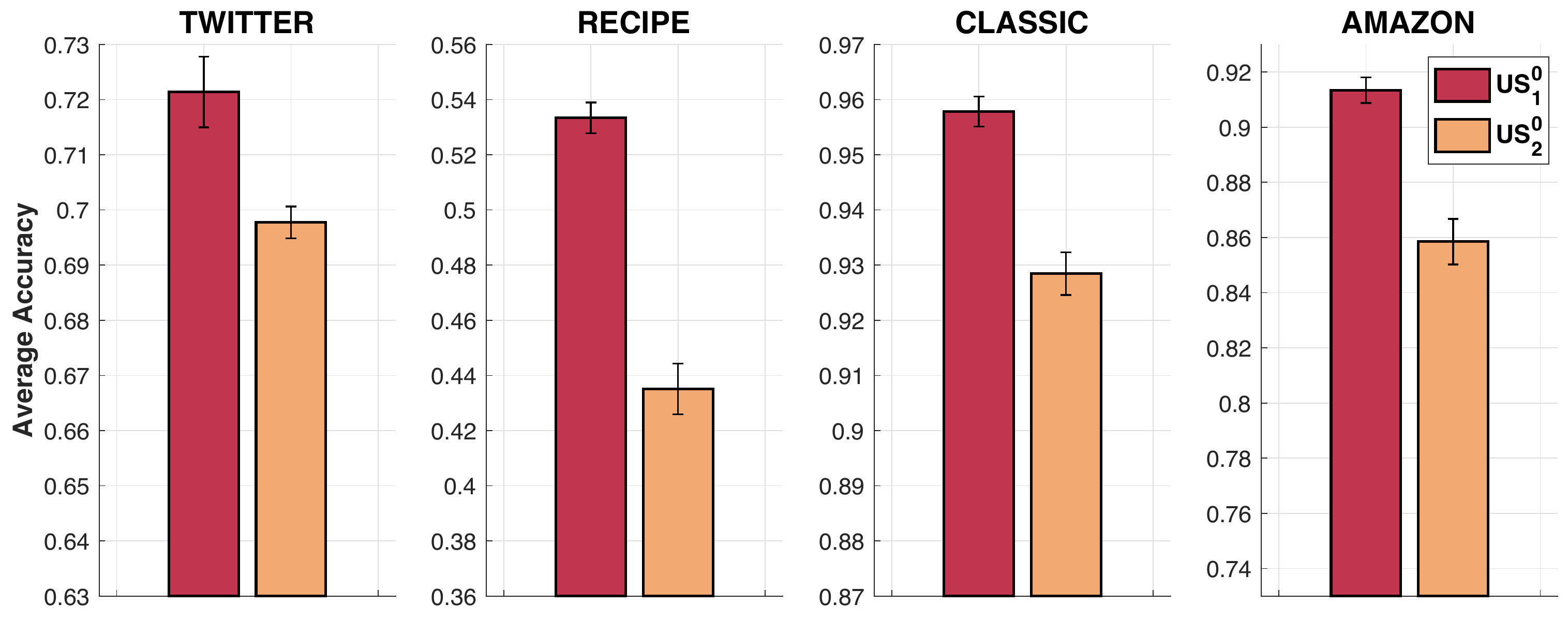}
  \end{center}
  \vspace{-14pt}
  \caption{SVM results and time consumption for kernel matrices in document classification with graph $\G_{\text{Sqrt}}$ with $M=10^4$.}
  \label{fg:DOC_10KSqrt_p1p2_app}
 \vspace{-6pt}
\end{figure}

\begin{figure}[h]
  \vspace{-2pt}
  \begin{center}
    \includegraphics[width=0.27\textwidth]{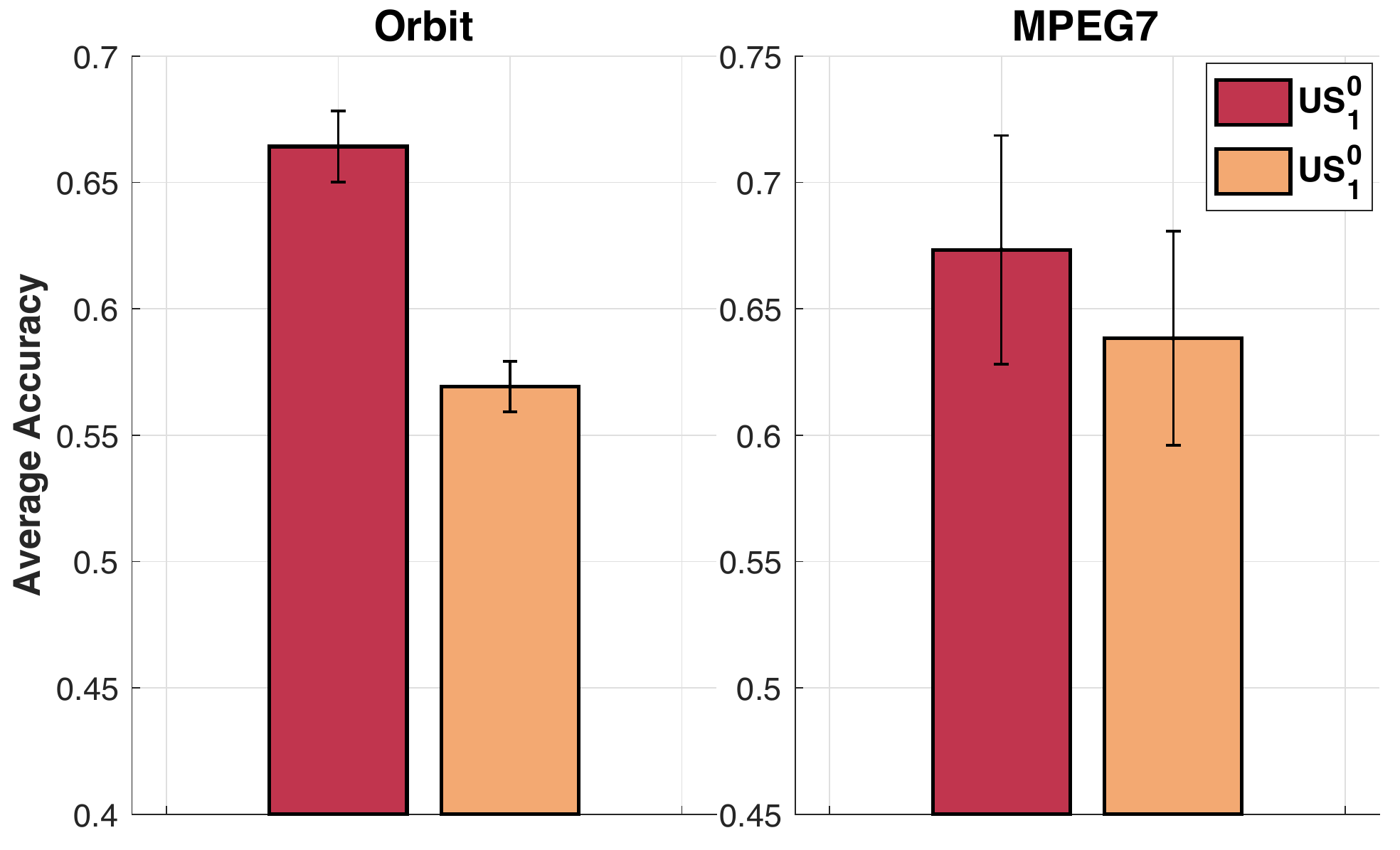}
  \end{center}
  \vspace{-14pt}
  \caption{SVM results and time consumption for kernel matrices in TDA with graph $\G_{\text{Sqrt}}$ with $M=10^2$ for \texttt{Orbit} and with $M=10^3$ for \texttt{MPEG7}.}
  \label{fg:TDA_Mix1K10KSqrt_p1p2_app}
 \vspace{-6pt}
\end{figure}

\begin{figure}[h]
  \vspace{-2pt}
  \begin{center}
    \includegraphics[width=0.42\textwidth]{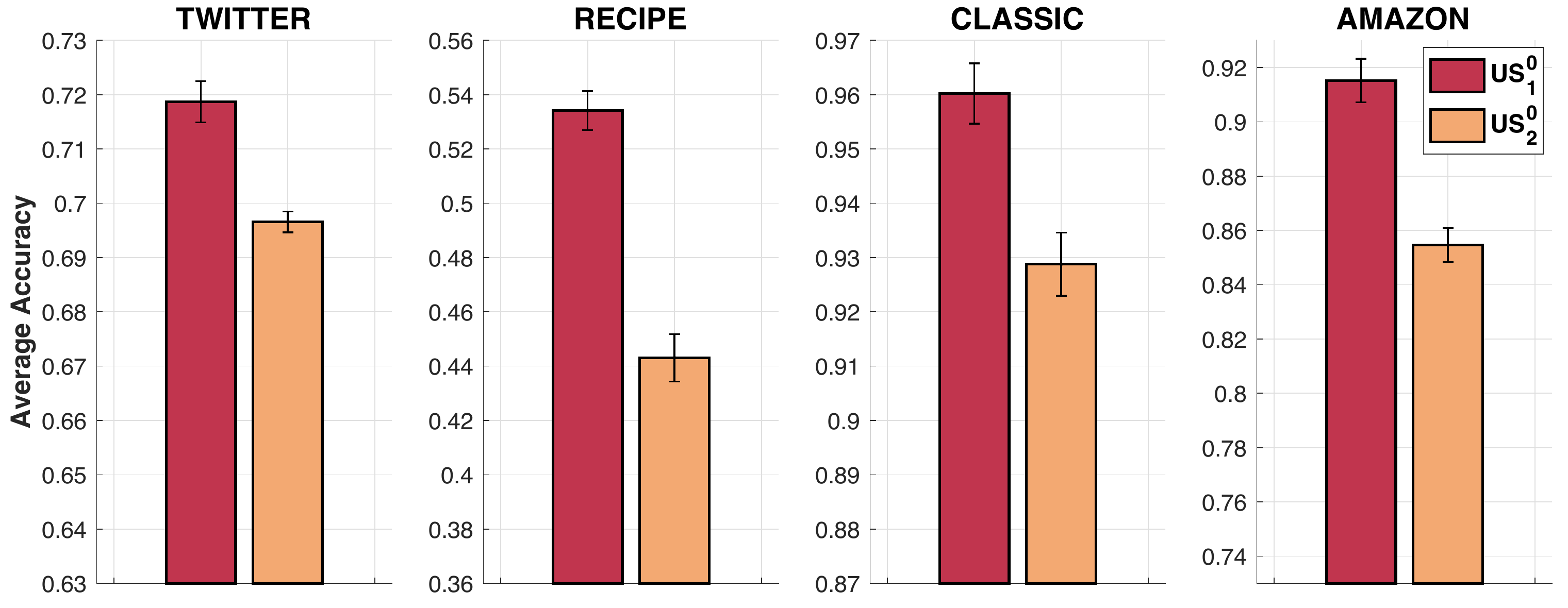}
  \end{center}
  \vspace{-14pt}
  \caption{SVM results and time consumption for kernel matrices in document classification with graph $\G_{\text{Log}}$ with $M=10^4$.}
  \label{fg:DOC_10KLog_p1p2_app}
 \vspace{-6pt}
\end{figure}

\begin{figure}[h]
  \vspace{-2pt}
  \begin{center}
    \includegraphics[width=0.27\textwidth]{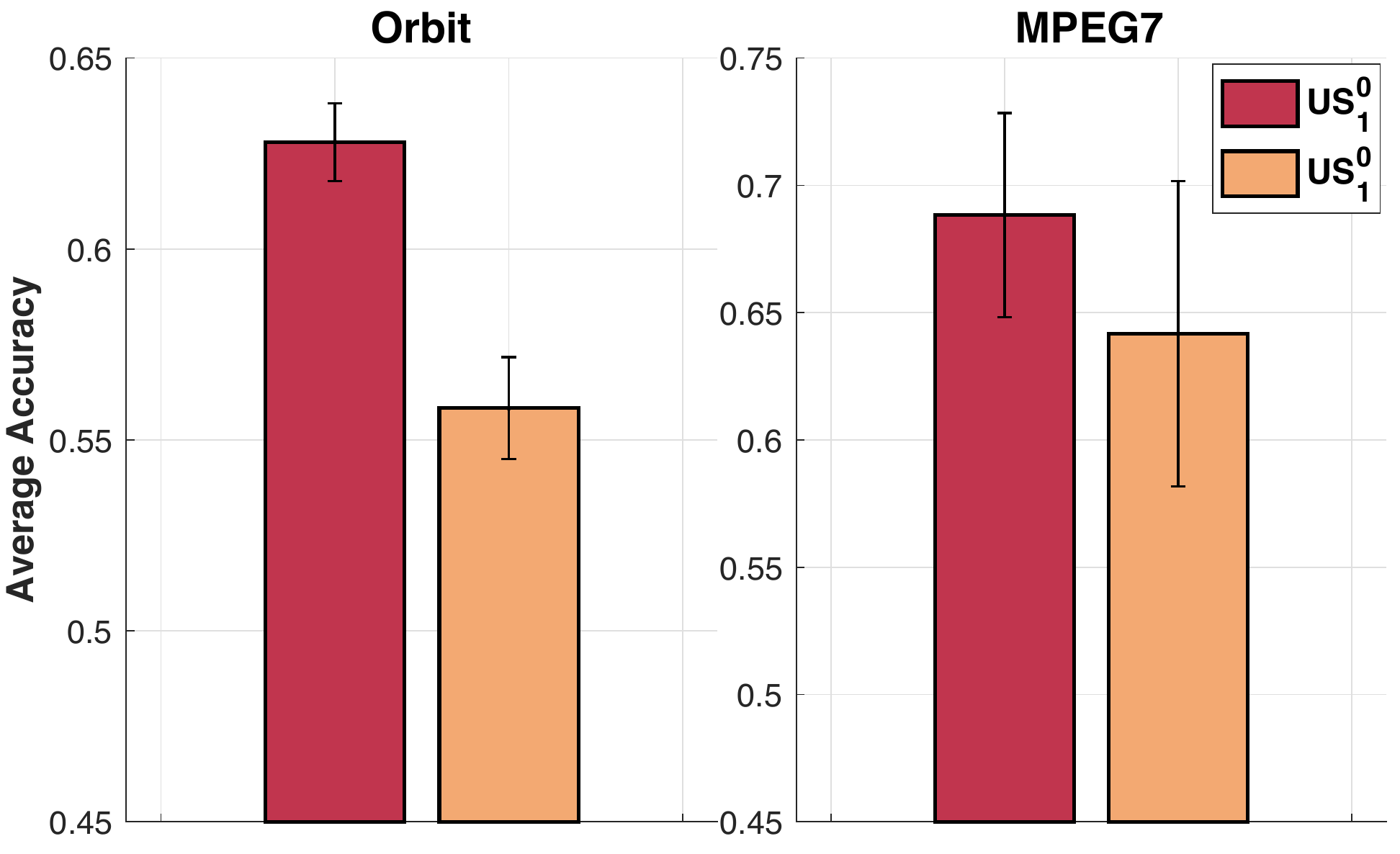}
  \end{center}
  \vspace{-14pt}
  \caption{SVM results and time consumption for kernel matrices in TDA with graph $\G_{\text{Log}}$ with $M=10^2$ for \texttt{Orbit} and with $M=10^3$ for \texttt{MPEG7}.}
  \label{fg:TDA_Mix1K10KLog_p1p2_app}
 \vspace{-6pt}
\end{figure}

\paragraph{With different exponent $p$ for UST.}
We also carry out experiments for different $p$ in unbalanced Sobolev transport using the same setting for $M$ in the main text (i.e., $M=10^4$ for document datasets, $M=10^3$ for \texttt{MPEG7} dataset and $M=10^2$ for \texttt{Orbit} dataset) on graph $\G_{\text{Sqrt}}$ and graph $\G_{\text{Log}}$.  Figure~\ref{fg:DOC_10KSqrt_p1p2_app} and Figure~\ref{fg:TDA_Mix1K10KSqrt_p1p2_app} illustrate performances on document classification and TDA respectively with graph $\G_{\text{Sqrt}}$. For graph $\G_{\text{Log}}$, the corresponding results are shown in Figure~\ref{fg:DOC_10KLog_p1p2_app} and Figure~\ref{fg:TDA_Mix1K10KLog_p1p2_app}.\footnote{We skip plots about time consumption since the time consumption of UST for $p=1$ and $p=2$ are almost identical. Please refer to other Figures where we illustrate the time consumption of UST for $p=1$.}

\paragraph{With Sinkhorn divergence-based approach for UOT~\citep{sejourne2019sinkhorn} as an extra baseline.} Furthermore, we also consider Sinkhorn divergence-based approach for UOT ($\text{SD}_{\text{UOT}}$)~\citep{sejourne2019sinkhorn} as an extra baseline. As we noted in the main manuscript, $\text{SD}_{\text{UOT}}$ is the debiased version of Sinkhorn-based approach for UOT ($\text{S}_{\text{UOT}}$) which may be helpful for applications. Both $\text{SD}_{\text{UOT}}$ and $\text{S}_{\text{UOT}}$ are empirically indefinite and they have the same computational complexity.

\begin{figure}[h]
  \vspace{-2pt}
  \begin{center}
    \includegraphics[width=0.42\textwidth]{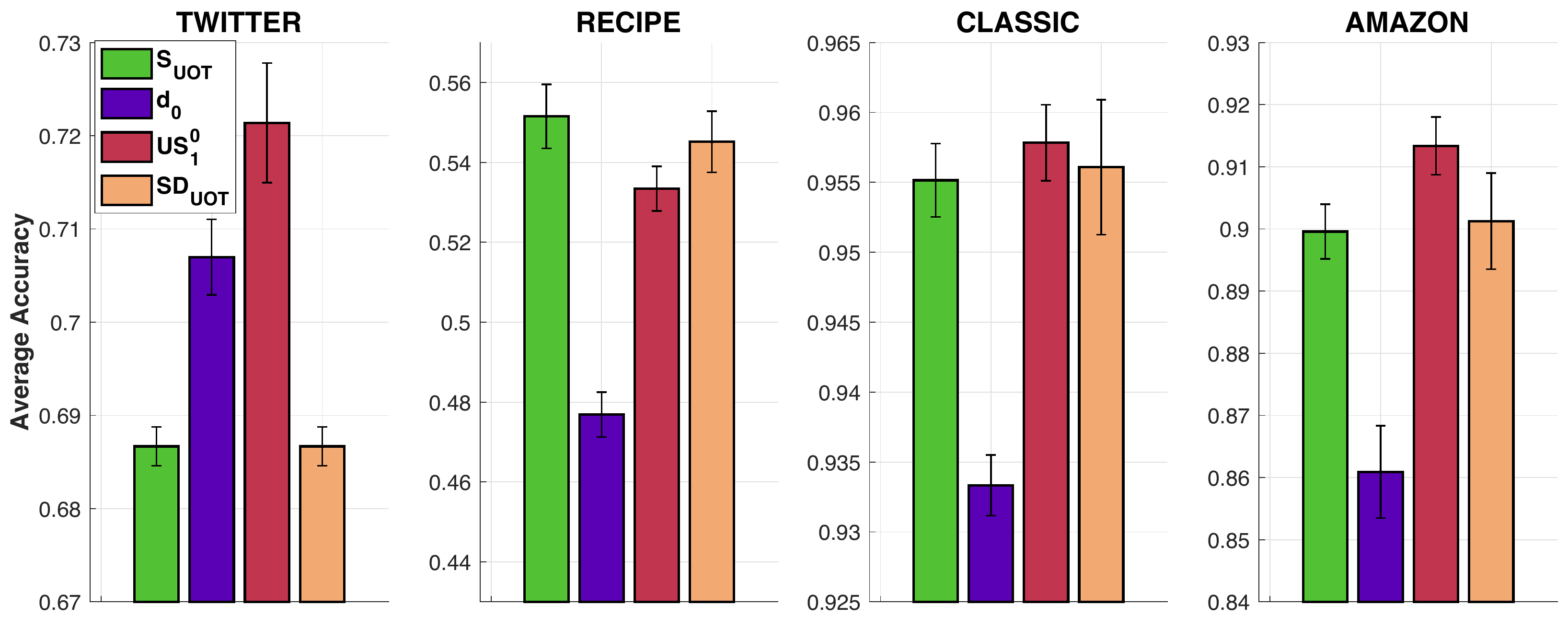}
  \end{center}
  \vspace{-14pt}
  \caption{SVM results for document classification with graph $\G_{\text{Sqrt}}$ with an extra baseline ($\text{SD}_{\text{UOT}}$).}
  \label{fg:SD_DOC_10KSqrt_main_app}
 \vspace{-6pt}
\end{figure}

\begin{figure}[h]
  \vspace{-2pt}
  \begin{center}
    \includegraphics[width=0.27\textwidth]{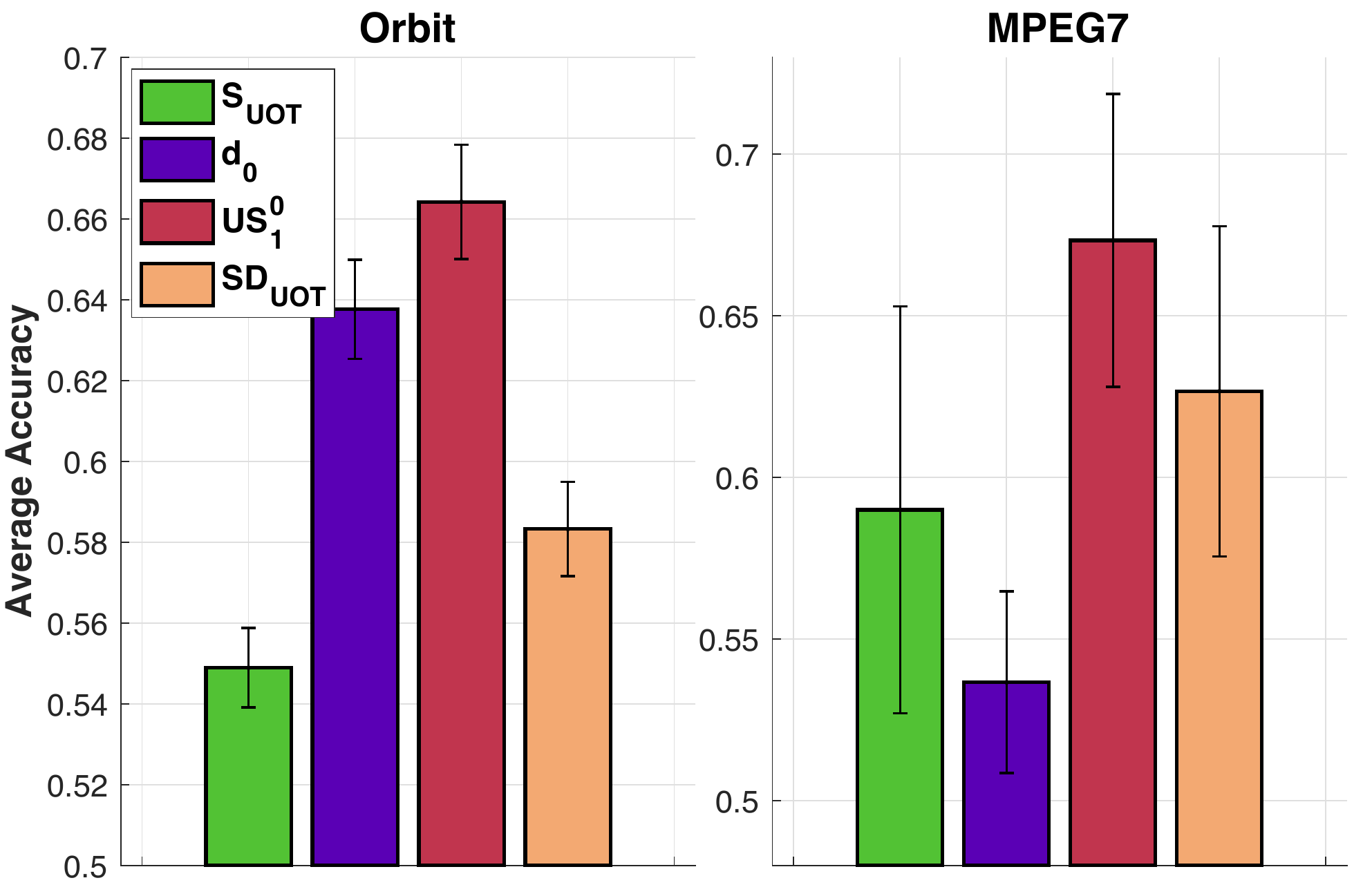}
  \end{center}
  \vspace{-14pt}
  \caption{SVM results for TDA with graph $\G_{\text{Sqrt}}$ with an extra baseline ($\text{SD}_{\text{UOT}}$).}
  \label{fg:SD_TDA_mix1K100Sqrt_main_app}
 \vspace{-6pt}
\end{figure}

\begin{figure}[h]
  \vspace{-2pt}
  \begin{center}
    \includegraphics[width=0.42\textwidth]{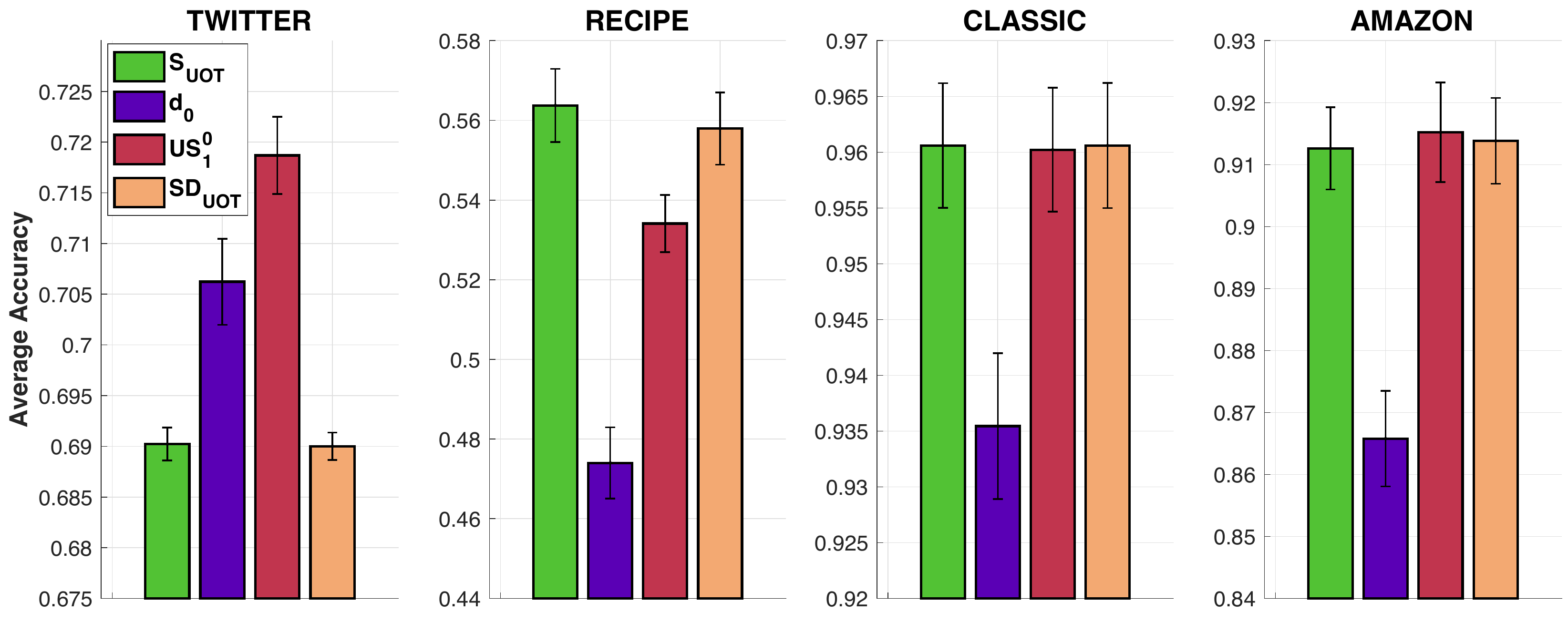}
  \end{center}
  \vspace{-14pt}
  \caption{SVM results for document classification with graph $\G_{\text{Log}}$ with an extra baseline ($\text{SD}_{\text{UOT}}$).}
  \label{fg:SD_DOC_10KLog_main_app}
 \vspace{-6pt}
\end{figure}

\begin{figure}[h]
  \vspace{-2pt}
  \begin{center}
    \includegraphics[width=0.27\textwidth]{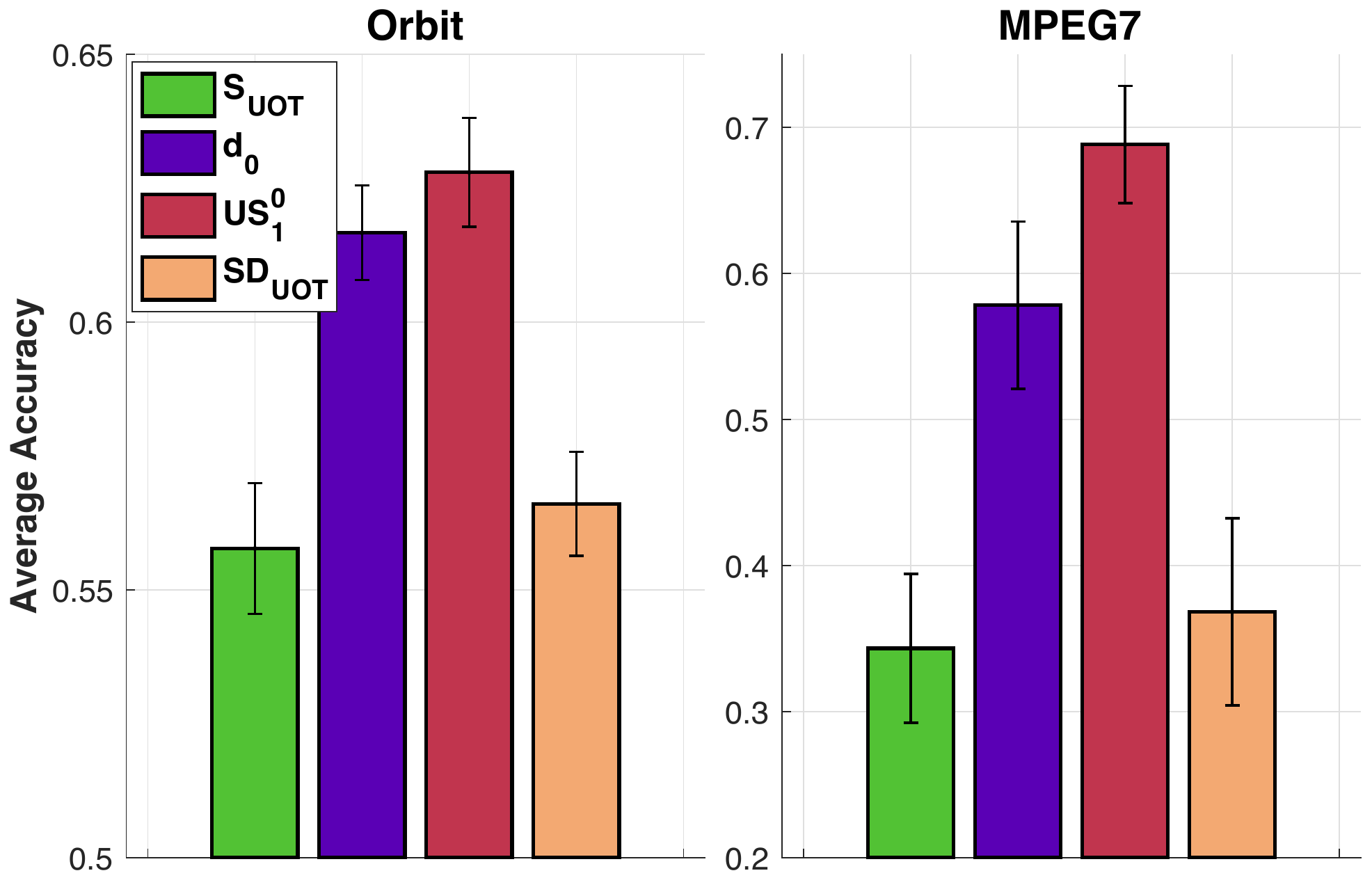}
  \end{center}
  \vspace{-14pt}
  \caption{SVM results for TDA with graph $\G_{\text{Log}}$ with an extra baseline ($\text{SD}_{\text{UOT}}$).}
  \label{fg:SD_TDA_mix1K100Log_main_app}
 \vspace{-6pt}
\end{figure}

We illustrate SVM results for document classification and TDA with the extra baseline $\text{SD}_{\text{UOT}}$ for both graph $\G_{\text{Sqrt}}$ and $\G_{\text{Log}}$ corresponding to Figure~\ref{fg:DOC_10KSqrt_main} (in the main text), Figure~\ref{fg:TDA_mix1K100Sqrt_main} (in the main text), Figure~\ref{fg:DOC_10KLog_main_app}, and Figure~\ref{fg:TDA_mix1K100Log_main_app} in Figure~\ref{fg:SD_DOC_10KSqrt_main_app}, Figure~\ref{fg:SD_TDA_mix1K100Sqrt_main_app}, Figure~\ref{fg:SD_DOC_10KLog_main_app}, Figure~\ref{fg:SD_TDA_mix1K100Log_main_app} respectively.


\subsubsection{Further Discussions on Empirical Results}

\paragraph{The unbalanced Sobolev transport (UST) $\text{US}_{p}^{\alpha}$ versus $d_{\alpha}$ of entropy partial transport (EPT) on a tree.} Overall, performances of the UST compare favorably with those of $d_{\alpha}$ of EPT on a tree. Moreover, time consumption of UST is comparable to that of $d_{\alpha}$ of EPT on trees. So, by exploiting the full graph structure, UST improves performances of $d_{\alpha}$ of EPT on a tree and still keeps the advantage about the computational complexity.

\paragraph{The unbalanced Sobolev transport (UST) versus Sinkhorn-based unbalanced optimal transport (UOT).} The performances of UST is comparable to those of Sinkhorn-based UOT. Recall that kernels for UST are positive definite while kernels for Sinkhorn-based UOT are empirically indefinite. This indefiniteness may affect performances of Sinkhorn-UOT in some settings (e.g., datasets or graph structure). It is worth noting that the UST is several order faster than Sinkhorn-based UOT. Therefore, it is prohibited to apply Sinkhorn-based UOT for large-scale settings while our proposed approach (UST) is scalable to such settings.

\paragraph{The effects of the number of slices (i.e., the number of root nodes used for averaging).} In general, when one increases the number of slices for the UST (and $d_{\alpha}$ of EPT on a tree), their corresponding performances are also increased but it comes with a trade-off about time consumption (i.e., linear to the number of slices). We observe that 10 slices seems a good trade-off between performances and time consumption, similar to observations in~\citep{le2021ept}.

\paragraph{Unbalanced Sobolev transport with different $p$.} In our experiments on document classification and TDA, we observe that $p=1$ for UST consistently gives better performances than $p=2$ for UST.\footnote{Recall that UST with $p=1$ has a stronger connection to EPT on graphs thatn UST with $p=2$ as illustrated in Lemma~\ref{lm:lipschitz-vs-sobolev}.} Generally, one may turn parameter $p$ to improve performances of UST in applications.

\paragraph{The extra baseline: Sinkhorn divergence-based approach for UOT.} In our experiments, the performances of the extra baseline $\text{SD}_{\text{UOT}}$ are relative with those of $\text{S}_{\text{UOT}}$ when comparing with performances of $d_{\alpha}$ (EPT on a tree) and our proposed UST. The debias property of $\text{SD}_{\text{UOT}}$ improves performances of $\text{S}_{\text{UOT}}$ in some datasets, especially for datasets in TDA tasks (\texttt{Orbit} and $\texttt{MPEG7}$). For document datasets, performances of $\text{SD}_{\text{UOT}}$ and $\text{S}_{\text{UOT}}$ are comparative (the role of debias property is not clear).

\end{document}